\documentclass{article}

     \PassOptionsToPackage{numbers, compress}{natbib}

 \usepackage[main, final]{neurips_2026}

\usepackage[utf8]{inputenc} %
\usepackage[T1]{fontenc}    %
\usepackage{url}            %
\usepackage{booktabs}       %
\usepackage{amsfonts}       %
\usepackage{nicefrac}       %
\usepackage{microtype}      %
\usepackage{xcolor}         %

\usepackage{xcolor}
\definecolor{mydarkblue}{rgb}{0,0.08,0.45} %

\usepackage{amsmath,amsfonts,bm}

\def\eqref#1{equation~\ref{#1}}

\def\1{\bm{1}}

\def\vb{{\bm{b}}}

\def\vm{{\bm{m}}}

\def\vu{{\bm{u}}}
\def\vv{{\bm{v}}}
\def\vw{{\bm{w}}}
\def\vx{{\bm{x}}}

\def\vz{{\bm{z}}}

\def\mI{{\bm{I}}}

\DeclareMathAlphabet{\mathsfit}{\encodingdefault}{\sfdefault}{m}{sl}
\SetMathAlphabet{\mathsfit}{bold}{\encodingdefault}{\sfdefault}{bx}{n}

 \usepackage[colorlinks,allcolors=mydarkblue,breaklinks]{hyperref}
\usepackage{booktabs}
\usepackage{colortbl}
\usepackage{subcaption}
\usepackage{textcomp}
\usepackage{wrapfig}

\definecolor{gold}{RGB}{255,215,0}
\definecolor{silver}{RGB}{192,192,192}
\definecolor{bronze}{RGB}{205,127,50}

\usepackage{xspace}
\newcommand{\eg}{\textit{e.g.}\xspace}
\newcommand{\ie}{\textit{i.e.}\xspace}

\usepackage{amsthm}
\newtheorem*{remark}{Remark}

\renewcommand{\mid}{\,|\,}

\usepackage{algorithm}
\usepackage{algorithmic}
\usepackage{amsmath, amssymb}

\usepackage{tikz,pgfplots}
\usetikzlibrary{arrows.meta,calc,fit,positioning,backgrounds,shadows.blur}
\usetikzlibrary{calc, positioning, arrows.meta, shapes.geometric, fit, backgrounds}
\newlength{\figurewidth}
\newlength{\figureheight}

\pgfplotsset{/pgf/number format/.cd,1000 sep={}}
  
\usepackage[capitalize,sort,compress,nameinlink]{cleveref}
\crefname{section}{Sec.}{Secs.}
\crefname{appendix}{App.}{Apps.}
\crefname{figure}{Fig.}{Figs.}
\crefname{equation}{Eq.}{Eqs.}

\renewcommand{\paragraph}[1]{\smallskip\noindent\textbf{#1}~~}

\title{Sparsely Supervised Diffusion}

\author{%
  Wenshuai Zhao\textsuperscript{1,4}
  \quad Zhiyuan Li\textsuperscript{1}
  \quad Yi Zhao\textsuperscript{2}
  \quad Mohammad Hassan Vali\textsuperscript{1,4} \\
  \bfseries Martin Trapp\textsuperscript{3}
  \quad Joni Pajarinen\textsuperscript{1}
  \quad Juho Kannala\textsuperscript{1,5}
  \quad Arno Solin\textsuperscript{1,4} \\[0.5em]
  \normalfont\textsuperscript{1}Aalto University, Finland
  \quad\textsuperscript{2}Imperial College London, UK \\
  \normalfont\textsuperscript{3}KTH Royal Institute of Technology, Sweden
  \quad\textsuperscript{4}ELLIS Institute Finland \\
  \normalfont\textsuperscript{5}University of Oulu, Finland \\[6pt]
  \normalfont Project website: \url{https://aaltoml.github.io/ssd/}
}

\begin{document}

\maketitle

\begin{abstract}
  Diffusion models have shown remarkable success across a wide range of generative tasks. However, they often suffer from spatially inconsistent generation, arguably due to excessive correlations learned by the model. This can produce samples that are locally plausible but globally inconsistent. We propose a principled method to mitigate this issue. Our method, sparsely supervised diffusion (SSD), is a simple yet effective masking framework that can be implemented with only a few lines of code. We analytically show that SSD fundamentally alters diffusion model training by modifying the spectrum of the data covariance and that it suppresses correlations in the data covariance matrix. Experiments show that our method more accurately approximates the underlying population score function, reduces memorization on small datasets, and promotes the use of essential contextual information during generation. Moreover, even when up to 98\% of pixels are masked, it achieves competitive FID scores across a range of datasets and, importantly, avoids the training instability commonly observed on small datasets.

\end{abstract}

\begin{wrapfigure}{r}{.43\linewidth}
\raggedleft\scriptsize
  \definecolor{crimson2143940}{RGB}{214,39,40}
  \definecolor{darkgray176}{RGB}{176,176,176}
  \definecolor{darkorange25512714}{RGB}{255,127,14}
  \definecolor{forestgreen4416044}{RGB}{44,160,44}
  
  \colorlet{first}{lightgray}
  \colorlet{second}{darkorange25512714!30!lightgray}
  \colorlet{third}{darkorange25512714}
  \colorlet{fourth}{darkorange25512714!50!yellow}

  \vspace*{-2em}
  \resizebox{.43\textwidth}{!}{%
  \begin{minipage}{.48\textwidth}
  \setlength{\figurewidth}{.86\columnwidth}
  \setlength{\figureheight}{.5\columnwidth}
  \scriptsize
  \pgfplotsset{
    scale only axis,
    grid=major, grid style={dotted, gray!40},
    xticklabel style={font=\tiny},
    yticklabel style={font=\tiny},
    ylabel near ticks,
    xlabel={Epochs},
    ylabel={$\leftarrow$ FID},
    xtick={500,1000,1500,2000,2500,3000,3500},
    xticklabels={0.5k,1k,1.5k,2k,2.5k,3k,3.5k},
    ticklabel style={color=black!40},
    axis line style={draw=none},
  }
\begin{tikzpicture}

\definecolor{crimson2143940}{RGB}{214,39,40}
\definecolor{darkgray176}{RGB}{176,176,176}
\definecolor{darkorange25512714}{RGB}{255,127,14}
\definecolor{forestgreen4416044}{RGB}{44,160,44}

\begin{axis}[
height=\figureheight,
tick align=outside,
tick pos=left,
width=\figurewidth,
x grid style={darkgray176},
xmajorgrids,
xmin=40, xmax=3560,
y grid style={darkgray176},
ymajorgrids,
ymin=4.24204020500183, ymax=14.5204348087311]
\addplot [draw=first, mark=none, opacity=0.6]
table[row sep=\\]{
x  y \\
200 8.96892929077148 \\
400 6.72095775604248 \\
600 5.76910829544067 \\
800 5.5677490234375 \\
1000 5.33182764053345 \\
1200 5.91435432434082 \\
1400 6.20600461959839 \\
1600 7.18269872665405 \\
1800 7.82406425476074 \\
2000 8.92866516113281 \\
2200 9.9609432220459 \\
2400 10.653489112854 \\
2600 11.3387908935547 \\
2800 11.864447593689 \\
3000 12.081072807312 \\
3200 13.0366659164429 \\
3400 13.724889755249 \\
};
\addplot [draw=second, mark=none, opacity=0.6]
table[row sep=\\]{
x  y \\
200 8.83951473236084 \\
400 6.74816179275513 \\
600 5.74889039993286 \\
800 5.40960502624512 \\
1000 5.54468631744385 \\
1200 5.91447448730469 \\
1400 6.27147388458252 \\
1600 7.07302856445312 \\
1800 7.58705520629883 \\
2000 8.61972713470459 \\
2200 9.42508125305176 \\
2400 10.2973337173462 \\
2600 11.0853090286255 \\
2800 11.3902139663696 \\
3000 12.1984491348267 \\
3200 12.6077003479004 \\
3400 13.2908735275269 \\
};
\addplot [draw=third, mark=none, opacity=0.6]
table[row sep=\\]{
x  y \\
200 10.449728012085 \\
400 7.53783893585205 \\
600 6.33512115478516 \\
800 5.62308931350708 \\
1000 5.35447406768799 \\
1200 5.2280101776123 \\
1400 5.02900362014771 \\
1600 4.98228216171265 \\
1800 4.84622478485107 \\
2000 4.91031980514526 \\
2200 4.90564107894897 \\
2400 5.13733673095703 \\
2600 5.37219715118408 \\
2800 5.31976366043091 \\
3000 5.53416633605957 \\
3200 5.90942811965942 \\
3400 6.36428022384644 \\
};
\addplot [draw=fourth, mark=none, opacity=0.6]
table[row sep=\\]{
x  y \\
200 14.0532350540161 \\
400 9.42171096801758 \\
600 7.58261489868164 \\
800 6.53447866439819 \\
1000 6.08382272720337 \\
1200 5.71555519104004 \\
1400 5.59054088592529 \\
1600 5.33140707015991 \\
1800 5.19654083251953 \\
2000 5.08614683151245 \\
2200 5.03406047821045 \\
2400 5.09185123443604 \\
2600 5.13203954696655 \\
2800 4.7092399597168 \\
3000 4.94003772735596 \\
3200 5.12761402130127 \\
3400 5.1040506362915 \\
};
\addplot [ultra thick, first, opacity=0.8, mark=none]
table[row sep=\\]{
200 8.96892929077148 \\
400 6.72095775604248 \\
600 5.76910829544067 \\
800 5.5677490234375 \\
1000 5.33182764053345 \\
1200 5.91435432434082 \\
1400 6.20600461959839 \\
1600 7.18269872665405 \\
1800 7.82406425476074 \\
2000 8.92866516113281 \\
2200 9.9609432220459 \\
2400 10.653489112854 \\
2600 11.3387908935547 \\
2800 11.864447593689 \\
3000 12.081072807312 \\
3200 13.0366659164429 \\
3400 13.724889755249 \\
};
\addplot [ultra thick, second, opacity=0.8, mark=none]
table[row sep=\\]{
200 8.83951473236084 \\
400 6.74816179275513 \\
600 5.74889039993286 \\
800 5.40960502624512 \\
1000 5.54468631744385 \\
1200 5.91447448730469 \\
1400 6.27147388458252 \\
1600 7.07302856445312 \\
1800 7.58705520629883 \\
2000 8.61972713470459 \\
2200 9.42508125305176 \\
2400 10.2973337173462 \\
2600 11.0853090286255 \\
2800 11.3902139663696 \\
3000 12.1984491348267 \\
3200 12.6077003479004 \\
3400 13.2908735275269 \\
};
\addplot [ultra thick, third, opacity=0.8, mark=none]
table[row sep=\\]{
200 10.449728012085 \\
400 7.53783893585205 \\
600 6.33512115478516 \\
800 5.62308931350708 \\
1000 5.35447406768799 \\
1200 5.2280101776123 \\
1400 5.02900362014771 \\
1600 4.98228216171265 \\
1800 4.84622478485107 \\
2000 4.91031980514526 \\
2200 4.90564107894897 \\
2400 5.13733673095703 \\
2600 5.37219715118408 \\
2800 5.31976366043091 \\
3000 5.53416633605957 \\
3200 5.90942811965942 \\
3400 6.36428022384644 \\
};
\addplot [ultra thick, fourth, opacity=0.8, mark=none]
table[row sep=\\]{
200 14.0532350540161 \\
400 9.42171096801758 \\
600 7.58261489868164 \\
800 6.53447866439819 \\
1000 6.08382272720337 \\
1200 5.71555519104004 \\
1400 5.59054088592529 \\
1600 5.33140707015991 \\
1800 5.19654083251953 \\
2000 5.08614683151245 \\
2200 5.03406047821045 \\
2400 5.09185123443604 \\
2600 5.13203954696655 \\
2800 4.7092399597168 \\
3000 4.94003772735596 \\
3200 5.12761402130127 \\
3400 5.1040506362915 \\
};

\node[font=\scriptsize] at (axis cs: 2000,11.5) {Overfitting};
\node[font=\scriptsize] at (axis cs: 2500,6.5) {Generalizing};

\end{axis}

\end{tikzpicture}
\\
  \begin{subfigure}{.24\textwidth}
  \centering
  \strut \color{first} No masking
  \tikz\node[inner sep=1.5pt, rounded corners=2pt, fill=first] 
    {\includegraphics[width=.97\textwidth]{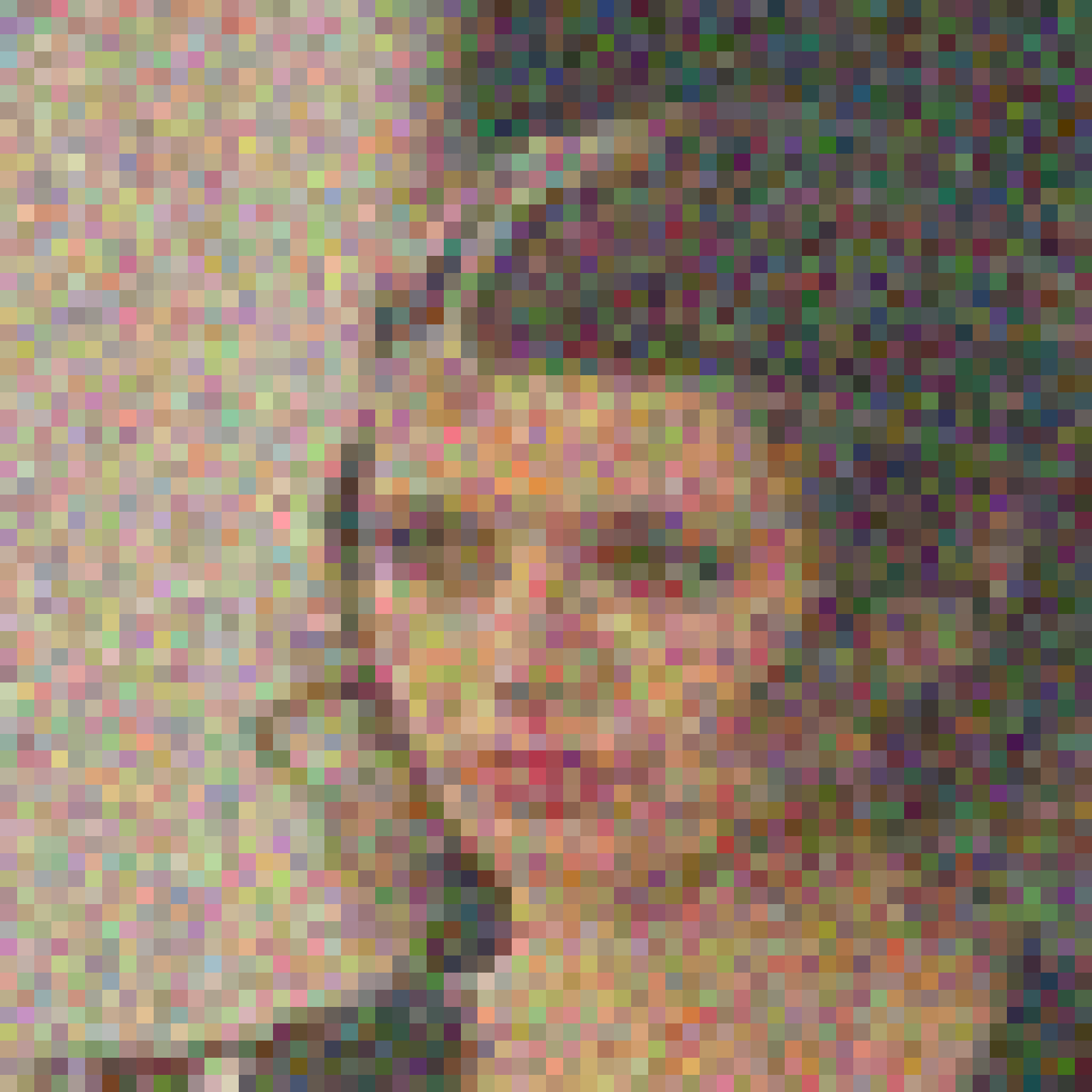}};
  \end{subfigure}
  \hfill  
  \begin{subfigure}{.24\textwidth}
    \centering
    \strut \color{second} $\eta=0.5$
    \tikz\node[inner sep=1.5pt, rounded corners=2pt, fill=second] 
    {\includegraphics[width=.97\textwidth]{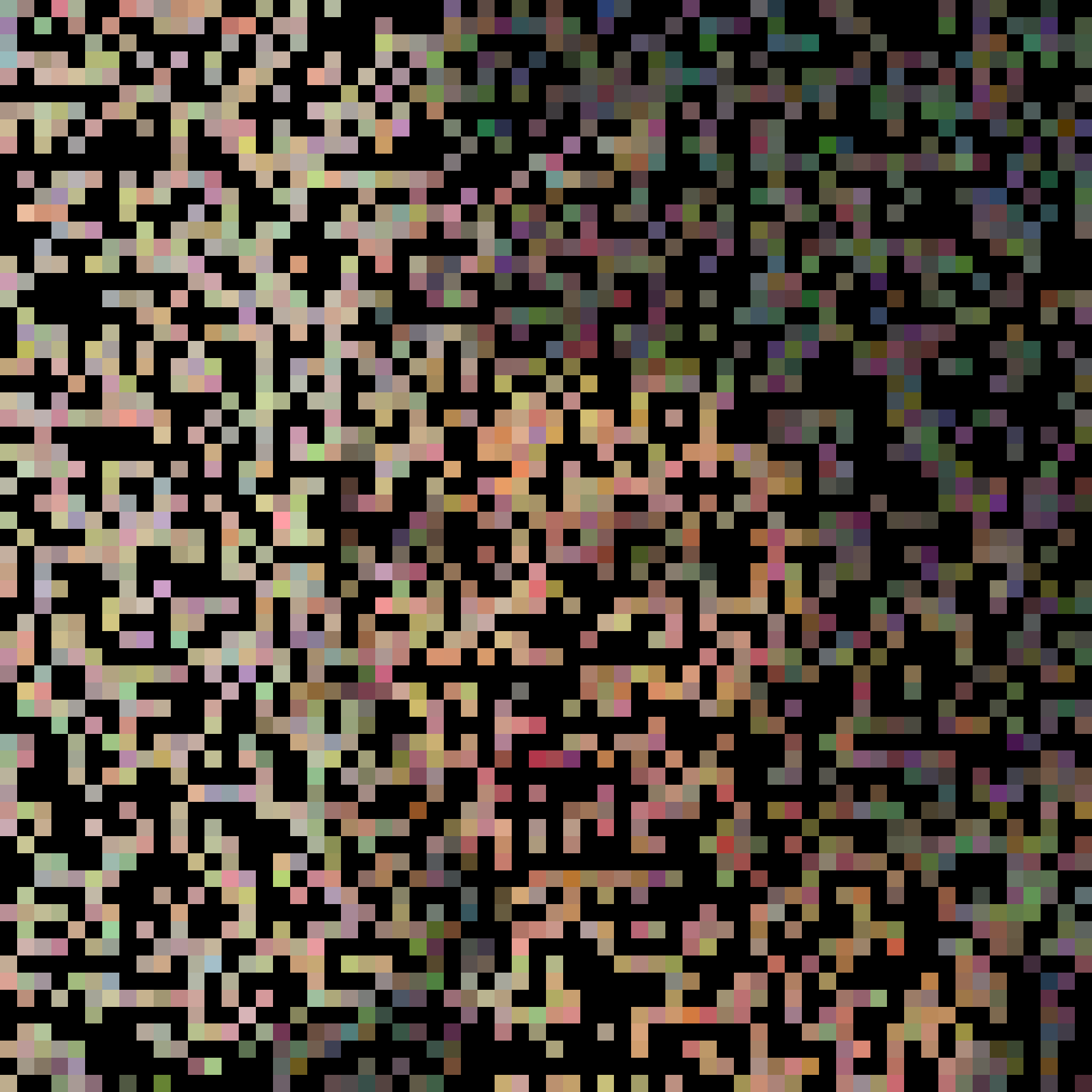}};
  \end{subfigure}
  \hfill
  \begin{subfigure}{.24\textwidth}
    \centering
    \strut \color{third} $\eta=0.95$
    \tikz\node[inner sep=1.5pt, rounded corners=2pt, fill=third] 
    {\includegraphics[width=.97\textwidth]{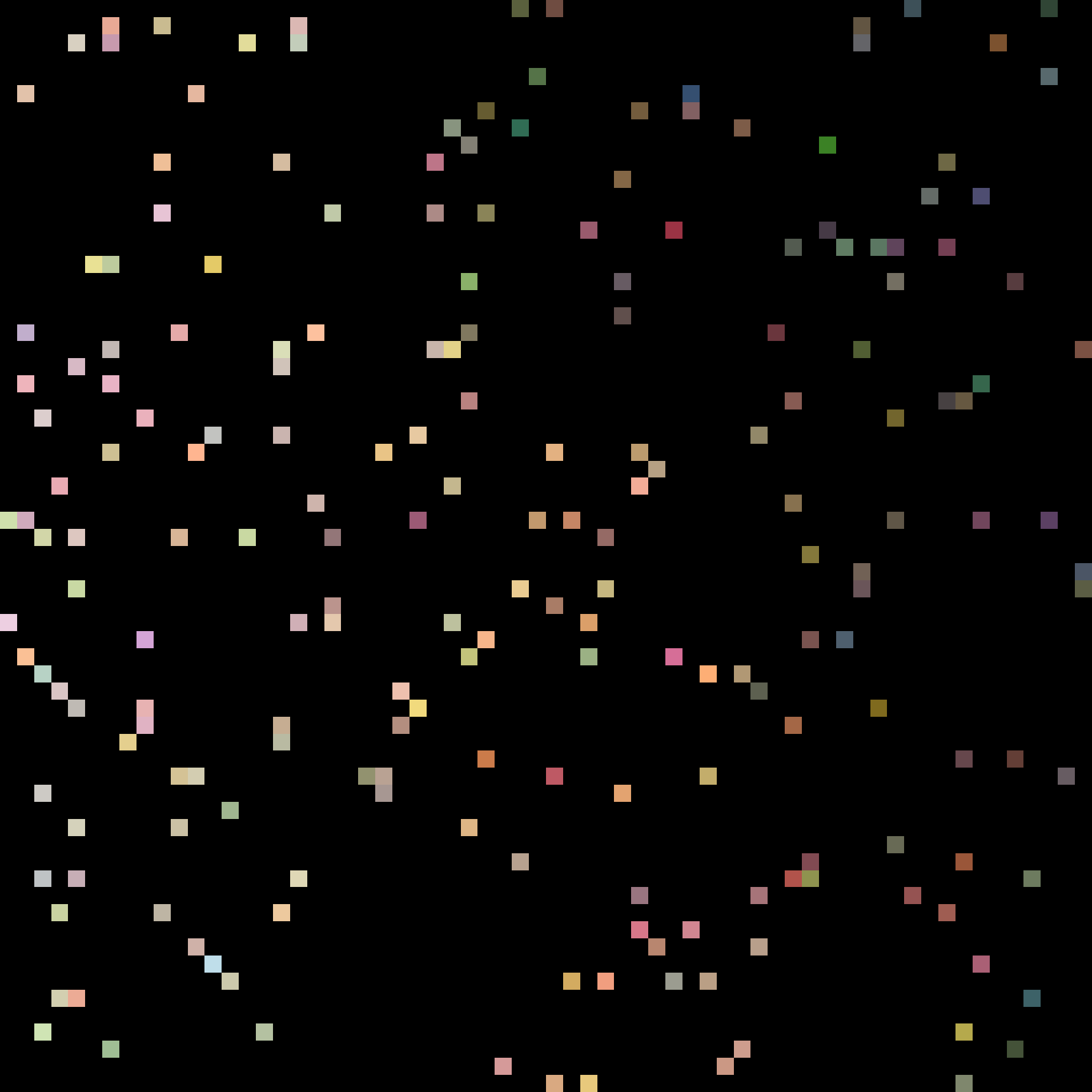}};
  \end{subfigure}
  \hfill
  \begin{subfigure}{.24\textwidth}
    \centering
    \strut \color{fourth} $\eta=0.98$
    \tikz\node[inner sep=1.5pt, rounded corners=2pt, fill=fourth] 
    {\includegraphics[width=.97\textwidth]{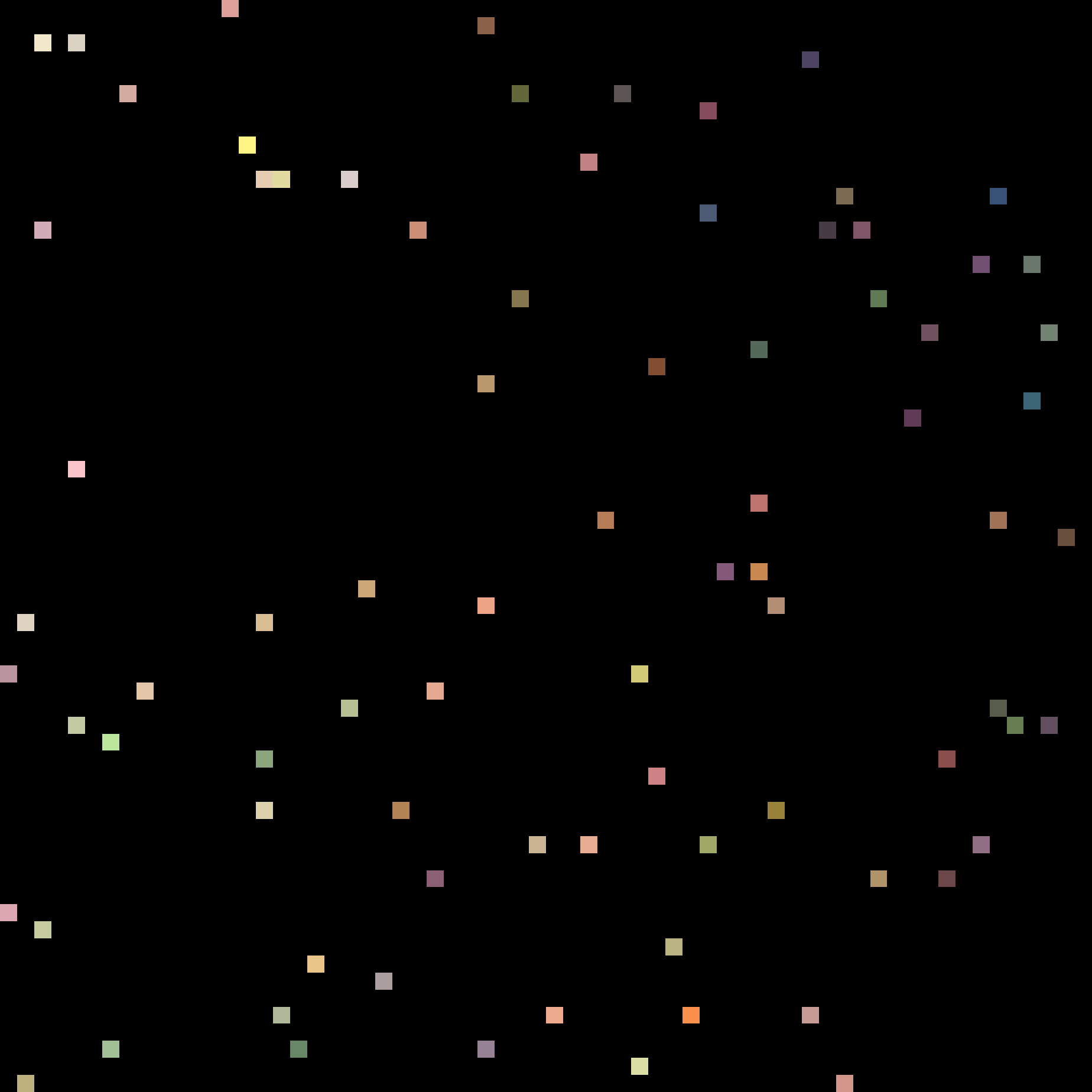}};
  \end{subfigure}\\[2pt]
  \end{minipage}}\\[-5pt]
  \caption{\textbf{Masking up to 98\% of pixels improves training}. SSD learns exclusively from  unmasked pixels while being encouraged to generalize over the masked regions.}%
  \label{fig:teaser}
  \vspace*{-1em}
\end{wrapfigure}
\section{Introduction}
Generative modeling aims to approximate and sample from typically unknown data distributions~\citep{albergo2023stochastic}. Among the various frameworks proposed~\citep{goodfellow2014generative, kingma2013auto, van2016pixel, papamakarios2021normalizing}, diffusion models~\citep{ho2020denoising, peebles2023scalable} including flow matching~\citep{lipman2022flow, liu2022flow, albergo2022building, albergo2023stochastic} have achieved remarkable success across diverse domains~\citep{ma2024sit, rombach2022high, saharia2022photorealistic, blattmann2023stable, kong2020diffwave}, largely attributable to their simple regression-based training objective, such as to regress the injected noise or score function~\citep{sohl2015deep, song2019generative, ho2020denoising, song2020score}. However, diffusion models suffer from generating unrealistic samples even when only trained on natural data~\citep{shen2024rethinking}, such as the well-known six-finger hand generation problem~\citep{narasimhaswamy2024handiffuser}. This might be problematic in critical domains such as health~\citep{chenthamarakshan2020cogmol} and robotics~\citep{chi2025diffusion}. 

This spatially inconsistent generation can arguably be attributed to spurious correlations learned from finite training sets~\citep{bonnaire2025diffusion}. Specifically, the score-matching objective used to train diffusion models relies on the empirical score function derived from limited data, whereas the ideal objective would target the underlying continuous population score function~\citep{kadkhodaie2024generalization, niedoba2025towards, kamb2025an}. This objective mismatch leads the model prone to memorization~\citep{gu2023memorization, bonnaire2025diffusion}, causing it to generate samples that merely recombine training samples rather than faithfully sampling from the true target distribution. While prior studies have investigated this phenomenon~\citep{aithal2024understanding} through the lenses of locality~\citep{lukoianovlocality, niedoba2025towards}, spectral bias~\citep{rissanen2023generative, wang2025analytical}, and implicit dynamical regularization~\citep{bonnaire2025diffusion, baptista2025memorization}, existing solutions remain largely heuristic. Although domain-specific workarounds have been proposed~\citep{narasimhaswamy2024handiffuser}, to the best of our knowledge, a principled methodology for mitigating spatial inconsistencies in generation has yet to be established.

We introduce \emph{Sparsely Supervised Diffusion} (SSD), a principled framework designed to fundamentally alter the standard training dynamics of diffusion models. SSD mitigates spatial inconsistency by suppressing spurious correlations in the data covariance matrix. During training, we apply random spatial masks when computing the regression loss, which forces the model to learn solely from unmasked pixels and inherently generalize to the masked regions. Notably, SSD is entirely model-agnostic and can be integrated with just a few lines of code.

Our primary contributions are summarized as follows: {\em (i)}~We analytically show that SSD modifies the spectrum of the data covariance matrix and functions as an \textit{Occam's razor} over data correlations. {\em (ii)}~Because the widely used Fréchet Inception Distance (FID) score cannot adequately reflect the spatial consistency of the generated samples, we design a targeted set of experiments. These evaluations show that SSD generates samples with better spatial consistency, more accurately approximates the underlying population score function, and better captures essential contextual information. {\em (iii)}~Large-scale experiments also show that SSD achieves competitive FID scores even when up to 98\% pixels are masked. Furthermore, it mitigates memorization and training instability on small datasets.

\section{Related work}

We first discuss the underexplored spatially inconsistent generation problem in diffusion models, then survey masking strategies employed in diffusion models for different targets, and finally introduce work on analyzing the training dynamics of diffusion models.

\paragraph{Spatial inconsistency and memorization in diffusion models}
Despite the remarkable success of diffusion~\citep{ho2020denoising, song2020score} and flow matching models~\citep{lipman2022flow, albergo2023stochastic} in image and video generation~\citep{rombach2022high, ma2024sit}, they suffer from spatially inconsistent generation problems~\citep{aithal2024understanding}. Moreover, memorization can emerge when the training dataset is small~\citep{kadkhodaie2024generalization}. These problems come from the fundamental mechanism inherent in diffusion models~\citep{kadkhodaie2024generalization, kamb2025an, niedoba2025towards}. When denoising a given pixel, the model relies on certain neighborhood of the input image, due either to network inductive biases~\cite{kamb2025an} or excessive correlations learned by the model from finite samples~\cite{lukoianovlocality}. Although there are some heuristic works targeting for realistic generation in specific domains~\citep{narasimhaswamy2024handiffuser, lu2026phys4d}, no work addresses this problem in a principled way.

\paragraph{Masking in diffusion models}
We distinguish our method from existing masking techniques in diffusion models as we target for the underexplored spatial inconsistency problem. Discrete diffusion models incorporate masking as a replacement for Gaussian noise in continuous spaces, primarily to adapt diffusion to discrete domains such as text and code~\citep{austin2021structured, shi2024simplified, gat2024discrete}. Masked Diffusion Transformer (MDT, \cite{gao2023masked}) exposes the model only to unmasked patches and trains it to predict the missing ones. However, MDT relies on a heavy asymmetric encoder–decoder design, akin to Masked Autoencoders (MAE, \cite{he2022masked}), limiting its applicability to general diffusion frameworks. In contrast, our method introduces masking directly into the regression loss, making it architecture-agnostic and straightforward to implement. Ambient Diffusion~\citep{daras2023ambient, daras2025ambient} passively trains diffusion models using corrupted data, similar as ours, but does not investigate the spatial inconsistency problem. Moreover, we provide an theoretical analysis to show how the proposed SSD changes the training dynamics of diffusion models.

\paragraph{Training dynamics of diffusion models} Recent studies have sought to characterize the training dynamics of diffusion models to understand how they generalize to unseen data~\cite{bonnaire2025diffusion, kadkhodaie2024generalization}. Prior work has revealed a spectral bias in the training dynamics of diffusion models~\cite{rissanen2023generative}, wherein training progresses from a generalization regime to a memorization regime, corresponding to the sequential learning of generalizable low-frequency patterns followed by overfitting to individual data samples~\cite{bonnaire2025diffusion}. Within this line of work, the spectrum of the data covariance has emerged as a crucial factor governing diffusion training dynamics~\cite{wang2025analytical}. In this work, we analytically show that the proposed SSD fundamentally modifies the spectrum of the data covariance, thereby altering the training dynamics of diffusion models and mitigating both excessive correlation-induced inconsistencies and memorization.

\section{Background}
We provide an overview of diffusion and flow matching models~\citep{albergo2023stochastic}. Particularly, we discuss how the empirical denoiser tends to memorize training samples.

\subsection{Diffusion models}
Given an unknown data distribution, instead of directly estimating the probability density $p(\vx), \vx \in \mathbb{R}^d$, diffusion models learn the score function~\citep{song2020score} or denoiser~\citep{ho2020denoising} from noise-corrupted data to iteratively transform noises from a prior distribution to the target data distribution.

\paragraph{Forward process}
The forward process of diffusion models can be described via the following stochastic differential equation (SDE, see, \eg, \cite{sarkka2019applied}):
\begin{equation}
    \mathrm{d}\vz =f(\vz,t)\,\mathrm{d}t+g(t)\,\mathrm{d}\vw,
    \label{equ:sde}
\end{equation}
where $\vz \in \mathbb{R}^d$ represents intermediate corrupted samples during the diffusion process, and $f(\vz,t)$ and $g(t)$ are known as the drift and the diffusion functions. Typically, $\vw(t)$ is assumed to be the standard Wiener process. For each timestep $t \in (0, T]$, we obtain the marginal distribution $p_t(\vz)=\int p_t(\vz\mid \vx)p(\vx)\,\mathrm{d}\vx$ from \cref{equ:sde}. Generally, by setting proper $f(\vz, t)$ and $g(t)$, we would like to have $p_t(\vz\mid \vx)$ as a Gaussian distribution with closed-form mean and variance.\looseness-1

\paragraph{Backward process}
The core objective of diffusion models is to learn the time-reversal of \cref{equ:sde}. This reverse process is governed by the corresponding reverse-time SDE~\citep{song2020score}:
\begin{equation}
\mathrm{d}\vz = \left[f(\vz, t) - g(t)^2\nabla_\vz \log p_t(\vz)\right]\mathrm{d}t + g(t)\,\mathrm{d}\tilde{\vw}.
\label{equ:dz}
\end{equation}

The reverse SDE in \cref{equ:dz} requires estimation of the score function $\nabla_\vz \log p_t(\vz)$. For the chosen diffusion process, the score function takes the explicit form~\citep{niedoba2025towards}:
\begin{equation}
\nabla_\vz \log p_t(\vz) = \frac{\mathbb{E}[\vx \mid \vz, t] - \vz}{t^2}.
\label{equ:score}
\end{equation}
Note that the score estimation in \cref{equ:score} is mathematically equivalent to estimating the posterior mean $\mathbb{E}[\vx \mid \vz, t]$, which is also the objective of denoising. Since the true data distribution $p(\vx)$ is generally unknown, exact computation of the posterior $p_t(\vx \mid \vz)$ and hence $\mathbb{E}[\vx \mid \vz, t]$ is intractable. Instead, diffusion models employ neural networks as denoisers to approximate $\mathbb{E}[\vx \mid \vz, t]$. These networks are trained using an empirical data distribution $p_{\mathcal{D}}(\vx) = \frac{1}{N}\sum_{\vx^{(i)} \in \mathcal{D}} \delta(\vx - \vx^{i})$, where the dataset is $\mathcal{D} = \{\vx^{1}, \ldots, \vx^{N} \mid \vx^{i} \sim p(\vx)\}$. The training objective is:\looseness-1
\begin{equation}
  \mathbb{E}_{\vx^{i}\sim p_{\mathcal{D}}(\vx), \vz\sim p_t(\vz \mid \vx^{i}), t\sim p(t)}\left[\lambda(t)\left\|\vx^{i} - D_\theta(\vz, t)\right\|^2\right],
\label{equ:obj_denoiser}
\end{equation}
where $\lambda(t)$ is a weighting function and $D_\theta(\vz, t)$ represents the neural network denoiser.

\paragraph{Optimal denoiser}
The theoretical minimizer of \cref{equ:obj_denoiser} and the optimal denoiser for any $(\vz, t)$ pair is the empirical posterior mean \citep{vincent2010stacked}:
\begin{equation}
\mathbb{E}_{\vx\sim p_{\mathcal{D}}}[\vx \mid \vz, t] = \sum_{\vx^{i}\in {\mathcal{D}}} p_t(\vx^{i} \mid \vz)\vx^{i},
\label{equ:optimal_denoiser}
\end{equation}
which is the average over the images of the data set $\mathcal{D}$, weighted by their posterior probability. Note that the empirical optimal denoiser in \cref{equ:optimal_denoiser} can only generate samples in the training dataset $\mathcal{D}$, \ie, \emph{the optimal denoiser memorizes}~\citep{kamb2025an}. How \cref{equ:dz,equ:score} result in \emph{memorization} using the optimal denoiser in \cref{equ:optimal_denoiser} has been well established in~\citet{biroli2024dynamical}.

\subsection{Flow matching}
Flow matching provides a unified perspective on diffusion models by directly learning a time-dependent vector field that transforms noisy data $\vz$ into the target data distribution. This transformation admits multiple probability paths, such as diffusion paths and linear conditional paths, with diffusion models arising as a special case. In this section, we focus on the objective of learning a linear conditional probability path:
\begin{equation}
    \mathcal{L}_{\textrm{FM}}=\mathbb{E}_{t, \pi(\vx_0), p(\vx_1)}\left\| \vv_{\theta}(t, \vz \mid \vx_1) -(\vx_1-\vx_0) \right\|^2,
\end{equation}
where $\vx_0 \in \pi(\vx_0)$ denotes noise samples from a prior distribution, $\vx_1 \in p(\vx_1)$ are target data samples, and $\vz$ is the intermediate state. By learning the vector field $\vv_{\theta}(t, \vz)$, flow matching iteratively transforms noise samples into data samples through an ordinary differential equation
\begin{equation}
    \vz_{t+h}=\vz_t + h \vv_{\theta}(\vz_t, t),
\end{equation}
where $h>0$ is a user-defined time step. 

\section{Methods}

We first introduce the proposed \emph{Sparsely Supervised Diffusion} (SSD). We then show that the proposed masking strategy alters the spectrum of the data covariance, thereby changing the learning dynamics of diffusion models. Finally, we analyze how the changed spectrum changes the locality mechanism in diffusion models.

\subsection{Sparsely supervised diffusion (SSD)}

To mitigate the spatially inconsistent generation problem in diffusion models, which arguably arises from the excessive exploitation of data correlations that may be biased by limited samples, we propose SSD, a simple yet effective masking strategy for training diffusion models. The masking strategy introduced in SSD is general and can be applied to both diffusion and flow-matching models. For clarity, we present it under the flow matching formulation:
\begin{equation}
\mathcal{L}_{\textrm{SSD}}(\vv_{\theta})=\mathbb{E}_{\vx \sim p(\vx)} \mathbb{E}_{\vm \sim p(\vm)} \Big[\sum_{j \:\forall \: \vm_j = 1}
\| \vv_{\theta}(t, \vz \vert \vx)[j] - \vv^\ast(t, \vz \vert \vx)[j] \|^2 \Big],
\label{equ:smd}
\end{equation}
\begin{wrapfigure}{r}{0.5\columnwidth}
    \centering\footnotesize
    \vspace{-\baselineskip} %
    \resizebox{1\linewidth}{!}{%

\newcommand{\ImgSize}{1.35cm}   %
\newcommand{\ColSep}{2.8cm}     %
\newcommand{\RowSep}{2.4cm}     %
\newcommand{\HeaderGap}{0.8cm}  %

\newcommand{\UnetW}{1.6cm}       %
\newcommand{\UnetGap}{0.25cm}    %
\newcommand{\UnetBridgeY}{0.35cm}%

\tikzset{
  font=\footnotesize,
  imgnode/.style={
    draw=black!30,
    line width=0.5pt,
    fill=white,
    minimum width=\ImgSize,
    minimum height=\ImgSize,
    inner sep=0pt,
    path picture={
      \draw[black!10, step=0.2] (path picture bounding box.south west) grid (path picture bounding box.north east);
      \draw[black!20] (path picture bounding box.south west) -- (path picture bounding box.north east);
    }
  },
  masknode/.style={
    imgnode,
    fill=black!80,
    draw=white,
    line width=1pt
  },
  unetblock/.style={
    fill=black!5,
    draw=black!40,
    line width=0.8pt,
    rounded corners=2pt
  },
  headerbox/.style={
    font=\bfseries\small,
    anchor=south,
    inner ysep=4pt
  },
  badge/.style={
    draw=none,
    rounded corners=1pt,
    font=\tiny\bfseries,
    align=center,
    inner sep=2pt,
    minimum height=\ImgSize
  },
  fwd/.style={-{Latex[length=5pt, width=4pt]}, draw=black!70, line width=0.8pt},
  bwd/.style={-{Latex[length=5pt, width=4pt]}, draw=red!70!black, line width=0.8pt, dashed, dash pattern=on 3pt off 2pt},
  vlabel/.style={
    rotate=90, 
    anchor=south, 
    font=\bfseries\scriptsize, 
    text=black!60
  }
}

\begin{tikzpicture}

\coordinate (L1) at (0,0);
\coordinate (M1) at (\ColSep, 0);
\coordinate (R1) at (2*\ColSep, 0);

\coordinate (L2) at (0, -\RowSep);
\coordinate (M2) at (\ColSep, -\RowSep);
\coordinate (R2) at (2*\ColSep, -\RowSep);

\node[imgnode] (noise1) at (L1) {}; 
\node[inner sep=0pt] at (L1) {\includegraphics[width=\ImgSize]{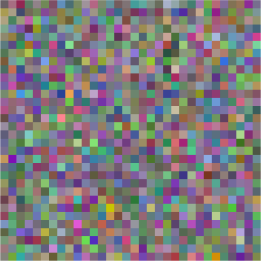}};

\node[imgnode] (vel) at (R1) {};
\node[inner sep=0pt] at (R1) {\includegraphics[width=\ImgSize]{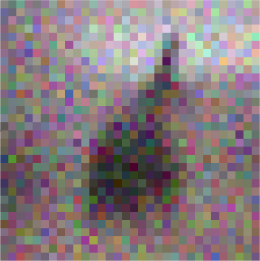}};

\node[imgnode] (noise2) at (L2) {};
\node[inner sep=0pt] at (L2) {\includegraphics[width=\ImgSize]{figs/noise}};

\begin{scope}[shift={(R2)}]
  \foreach \i in {2,1,0}{
    \coordinate (offset) at (-\i*0.15cm, \i*0.12cm);
    \node[masknode] (mask\i) at (offset) {}; 
    \node[inner sep=1pt, fill=white, shift={(offset)}] {\includegraphics[width=\ImgSize]{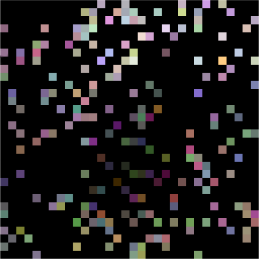}};
  }
\end{scope}
\node[fit=(mask2)(mask0), inner sep=0pt] (maskStack) {};

\foreach \pos/\name in {M1/u1, M2/u2} {
    \draw[unetblock] 
        ($(\pos) + (-0.5*\UnetW,  0.5*\ImgSize)$) --   %
        ($(\pos) + (-0.5*\UnetGap, \UnetBridgeY)$) --  %
        ($(\pos) + (-0.5*\UnetGap, -\UnetBridgeY)$) -- %
        ($(\pos) + (-0.5*\UnetW, -0.5*\ImgSize)$) --   %
        cycle;

    \draw[unetblock] 
        ($(\pos) + ( 0.5*\UnetW,  0.5*\ImgSize)$) --   %
        ($(\pos) + ( 0.5*\UnetGap, \UnetBridgeY)$) --  %
        ($(\pos) + ( 0.5*\UnetGap, -\UnetBridgeY)$) -- %
        ($(\pos) + ( 0.5*\UnetW, -0.5*\ImgSize)$) --   %
        cycle;
    
    \node[font=\sffamily\tiny, text=black!60, inner sep=1pt] at ($(\pos) + (0, -0.5*\ImgSize - 0.25cm)$) {U-Net};
    
    \coordinate (\name-in) at ($(\pos) + (-0.5*\UnetW, 0)$);
    \coordinate (\name-out) at ($(\pos) + ( 0.5*\UnetW, 0)$);
}

\coordinate (headerY) at ($(L1)+(0, \HeaderGap)$);
\node[headerbox] at (L1 |- headerY) {Input};
\node[headerbox] at (M1 |- headerY) {Model};
\node[headerbox] at (R1 |- headerY) {Supervision};

\node[below=3pt of noise1] {$x_0$ (Noise)};
\node[below=3pt of noise2] {$x_0$ (Noise)};
\node[below=3pt of vel] {$x_1 - x_0$};
\node[below=3pt of maskStack] {Masked Diff.};

\node[vlabel] at ($(noise1.west) + (-0.4cm, 0)$) {Flow Matching};
\node[vlabel] at ($(noise2.west) + (-0.4cm, 0)$) {SSD (Ours)};

\draw[fwd] (noise1) -- node[below,font=\tiny\strut]{fwd} (u1-in);
\draw[bwd] (vel) -- node[below,font=\tiny\strut]{bwd} (u1-out);
\draw[fwd] (noise2) -- (u2-in);

\foreach \yoffset in {0.2, 0, -0.2} {
    \draw[bwd] 
        ($(maskStack.west) + (0, \yoffset)$) coordinate (aux) -- (u2-out |- aux);
}

\node[badge, fill=orange!10, text=orange!60!black, anchor=west] 
    at ($(vel.east) + (0.2cm, 0)$) 
    {\rotatebox{90}{Full-pixel}};

\node[badge, fill=green!10, text=green!40!black, anchor=west] 
    at ($(maskStack.east) + (0.2cm, 0.1cm)$) 
    {\rotatebox{90}{Sparse}};

\end{tikzpicture} %
}
    \caption{\textbf{Overview of the proposed method} in comparison with standard flow matching (FM).}
    \label{fig:overview}
\end{wrapfigure}
where $\vm \in \{0,1\}^d$ denotes a randomly sampled binary mask with each entry $\vm_j \sim \mathrm{Bernoulli}(1-\eta)$, and $d$ is the dimensionality of the image $\vx$. $\vv^\ast$ is the target velocity field. The mask ratio $\eta\in(0,1)$ is used to control the proportion of visible elements. Through masking, the objective $\mathcal{L}_{\textrm{SSD}}$ is optimized only over the unmasked pixel positions.

\begin{remark}
    By masking pixels in the target signal, SSD forces the model to rely on essential contextual information to generate complete pixels, as illustrated in \cref{fig:overview}. At the same time, it implicitly generates multiple variants of each sample, thereby mitigating memorization.
\end{remark}

\subsection{Masking changes covariance spectrum}
The spectrum of data covariance, \ie, the set of eigenvalues, has been shown as the critical factor that affects the diffusion model training, such as convergence speed, separating generalization and memorization regimes and locality~\cite{wang2025analytical, bonnaire2025diffusion, lukoianovlocality}. Here we analytically show that the proposed masking strategy alters the spectrum and thus changes the learning dynamics as evidenced in~\cref{sec:exp}. 

To simplify the analysis, we compare the covariance matrices $\boldsymbol\Sigma$ and $\tilde{\boldsymbol\Sigma}$, computed from original data $\vx  \sim \mathcal{N}(\mathbf{0}, \boldsymbol{\Sigma})$ and masked data $\tilde{\vx } = \vm \odot \vx $, where $\odot$ denotes element-wise multiplication. Denote the eigendecomposition of $\bm\Sigma$ as
\begin{equation}
    \boldsymbol{\Sigma} = \mathbf U \boldsymbol{\Lambda} \mathbf U^\top,
\qquad
\boldsymbol{\Lambda} = \textrm{diag}(\lambda_1,\ldots,\lambda_d),
\label{equ:gaussian_covariance}
\end{equation}

where $\mathbf U$ is the matrix of eigenvectors $\vu_i$ and $\textrm{diag}(\lambda_1,\ldots,\lambda_d)$ represents the diagonal matrix formed by eigenvalues. We thus can compute the new covariance of masked data as
\begin{equation}
\tilde{\bm\Sigma} = (1-\eta)^2\boldsymbol{\Sigma} + \eta(1-\eta)\,\mathbf{D},
\label{equ:tilde_sigma}
\end{equation}
where $\mathbf{D}=\textrm{diag}(\bm\Sigma)$ represents the diagonal matrix formed from the diagonal entries of $\bm{\Sigma}$. The formulation of $\tilde{\bm\Sigma}$ reveals that the masking operation attenuates the off-diagonal elements of the covariance more strongly than the diagonal ones. In other words, \textbf{the resulting covariance matrix regularizes the inter-dimensional correlations present in the data}.\looseness-1

We project $\tilde{\bm\Sigma}$ onto one eigenvector $\vu_i$ of $\boldsymbol{\Sigma}$,
\begin{equation}
\tilde{\lambda}_i
:= \vu_i^\top \tilde{\boldsymbol{\Sigma}} \vu_i
= (1-\eta)^2 \lambda_i
+ \eta(1-\eta)\sum_{k} \vu_{ik}^2 \boldsymbol{\Sigma}_{kk}.
\label{equ:new_lambda}
\end{equation}
The spectrum change can be measured by the ratio:
\begin{equation}
\beta_i :=\frac{\tilde{\lambda}_i}{\lambda_i}
= (1-\eta)^2
+ \eta(1-\eta)\frac{\sum_k \vu_{ik}^2 \boldsymbol{\Sigma}_{kk}}{\lambda_i}.
\end{equation}
Note that $\sum_{k} \vu_{ik}^2 \boldsymbol{\Sigma}_{kk}=\vu_i^\top \mathbf D \vu_i,$ and $\lambda_i=\vu_i^\top \mathbf \Sigma \vu_i$. Thus, 
\begin{equation}
\beta_i \propto \frac{\vu_i^\top \mathbf D \vu_i}{\vu_i^\top \mathbf \Sigma \vu_i}.
\end{equation}
This equation indicates that the masking operation promotes the direction $\vu_i$ more than the method without masking when $\beta_i$ is large, and attenuates it when the ratio is small. A full derivation can be found in~\cref{sec:app_spectrum}.
\begin{remark}
    Intuitively, a large value of $\beta_i$ implies that the coordinate-wise variations along direction $\vu_i$ are strong (\ie, $\vu_i^\top \mathbf D \vu_i$ is large), while the overall data correlations captured by the full covariance matrix lead to a relatively small variance $\lambda_i$ along this direction.
\end{remark}

\subsection{Masking changes locality}
As shown by \citet{lukoianovlocality}, the \textit{locality} observed in diffusion models arises primarily from data statistics—specifically, the \textit{correlations} present in the training dataset—rather than from the inductive bias of the neural network, contrary to earlier claims~\cite{kamb2025an}. Building on this analysis,~\cref{equ:tilde_sigma} shows that the proposed masking strategy suppresses excessive correlations (non-diagonal elements in $\tilde{\bm\Sigma}$) induced by limited data samples, thereby modifying the underlying mechanism that gives rise to locality in diffusion models.

Locality can be empirically quantified using the gradient sensitivity field during pixel generation, which characterizes the extent to which other pixel locations influence a given pixel in the generative process. Formally, assuming a locally linear denoiser $f(\vx_t, t)=\bm{A}_t\vx_t+\vb_t$ and Gaussian data with covariance defined in~\cref{equ:gaussian_covariance}, the gradient sensitivity at output pixel $q$ is given by:
\begin{equation}
S_f^q(\vx_t, t)= \left[ \frac{\partial f(\vx_t, t)}{\partial \vx_t} \right]_q \notag= \frac{1}{\sqrt{\alpha_t}}
\left[
\mathbf{U}\,
\mathrm{diag}\!\left( \frac{\mathrm{SNR}_i}{\mathrm{SNR}_i + 1} \right)
\mathbf{U}^\top
\right]_q ,
\label{equ:sensitivity}
\end{equation}
where $\alpha_t$ is the noise level at timestep $t$. $\mathrm{SNR}_i$ represents the signal-to-noise ratio defined by $\mathrm{SNR}_i = \frac{\lambda_i^2}{\sigma_t^2}, \;\text{for } i = \{1, 2, \ldots, d\}$,
where $\sigma_t^2$ is the noise variance. 

From~\cref{equ:sensitivity} we can see that the sensitivity field is determined by the principal components of the covariance matrix and can be far from being local. As shown in~\cref{equ:new_lambda}, our method changes the eigenvalues of the data covariance and thereby changes locality. Intuitively, our method suppresses excessive correlations while promoting critical ones, effectively acting as an \textit{Occam’s razor} over data correlations~\cite{rasmussen2000occam}. As a result, the model exhibits stronger yet selective correlation structures, as evidenced by our experiments in~\cref{sec:gradient_sensitivity}.

\section{Experiments}\label{sec:exp}
\paragraph{Overview} We design a series of experiments to answer the following questions:
Does the proposed SSD masking strategy
\textbf{(Q1)}~affect FID scores?
\textbf{(Q2)}~help to stabilize training on small datasets?
\textbf{(Q3)}~mitigate memorization in diffusion-based generation?
\textbf{(Q4)}~enable the model to capture more essential contextual information during generation?
\textbf{(Q5)}~improve the spatial consistency of generated samples?
\textbf{(Q6)}~facilitate more accurate estimation of the population score function?

\paragraph{Data \& experimental setup} We evaluate the proposed method on four large-scale datasets: CIFAR10 (32$\times$32)~\citep{Krizhevsky2009LearningML}, CelebA-50K (64$\times$64)~\citep{liu2015faceattributes}, LSUN Bedroom (32$\times$32)~\citep{yu2015lsun}, and ImageNet (256$\times$256)~\citep{ILSVRC15}. For ImageNet, SSD is implemented on DiT-XL/2~\citep{peebles2023scalable} at 256$\times$256 resolution with the Stable Diffusion VAE (ft-MSE). We evaluate at a classifier-free guidance (CFG) scale of 1.5. Detailed training and evaluation settings can be found in~\cref{sec:supp_implementation}. For the remaining datasets, we adopt the flow matching framework~(FM, \cite{lipman2024flowmatchingguidecode}) with a convolutional U-Net architecture~\citep{ronneberger2015u}, using a fixed number of function evaluations (NFE) of 50 and Heun’s second-order method~\citep{ascher1998computer} as the ODE solver, combined with the sampling strategy of~\citet{karras2022elucidating}. Fréchet Inception Distance (FID)~\citep{heusel2017gans} is computed over 50k generated samples. In all experiments, we do not tune the masking ratio; instead, we evaluate a range of masking ratios when computational resources allow. Determining an optimal masking ratio is left for future work.

\begin{figure*}[t]
  \centering\footnotesize
  \pgfplotsset{
    grid=major, grid style={dotted, gray!40},
    xticklabel style={font=\tiny},
    yticklabel style={font=\tiny},
    ylabel near ticks,
    ylabel style={font=\scriptsize},
  }    
  \pgfplotsset{
  every axis legend/.append style={
    font=\scriptsize,
    inner sep=1pt, %
    nodes={
      scale=0.8,
      transform shape,
      inner xsep=1pt,
      inner ysep=1pt,
      text depth=0.1ex,
      text height=1.4ex,
    },
    row sep=-1pt,
    column sep=1pt,
  },
  }
  \setlength{\figurewidth}{.35\textwidth}
  \setlength{\figureheight}{.75\figurewidth}
    \begin{subfigure}{0.33\linewidth}
        \raggedleft
        \pgfplotsset{ylabel={$\leftarrow$ FID}}
\begin{tikzpicture}

\definecolor{coral}{RGB}{255,127,80}
\definecolor{darkgray176}{RGB}{176,176,176}
\definecolor{lightgray204}{RGB}{204,204,204}
\definecolor{steelblue}{RGB}{70,130,180}

\begin{axis}[
height=\figureheight,
legend cell align={left},
legend style={fill opacity=0.8, draw opacity=1, text opacity=1, draw=lightgray204},
tick align=outside,
tick pos=left,
width=\figurewidth,
x grid style={darkgray176},
xmajorgrids,
xmin=-46, xmax=3144,
xtick style={color=black},
y grid style={darkgray176},
ymajorgrids,
ymin=2, ymax=5,
ytick style={color=black}
]
\path [draw=steelblue, fill=steelblue, opacity=0.2]
(axis cs:99,6.75921147549172)
--(axis cs:99,6.00601012027244)
--(axis cs:199,3.676920859843)
--(axis cs:299,3.02701439994378)
--(axis cs:399,2.67729377193877)
--(axis cs:499,2.54035705592495)
--(axis cs:599,2.39659758514007)
--(axis cs:699,2.29836593451322)
--(axis cs:799,2.19116191423404)
--(axis cs:899,2.35851531195334)
--(axis cs:999,2.29049275987924)
--(axis cs:1099,2.20633692795614)
--(axis cs:1199,2.12373989920394)
--(axis cs:1299,2.32654386116812)
--(axis cs:1399,2.32156424331047)
--(axis cs:1499,2.2351512563749)
--(axis cs:1599,2.23548169718966)
--(axis cs:1699,2.19080098346013)
--(axis cs:1799,2.16205093246665)
--(axis cs:1899,2.31337538071935)
--(axis cs:1999,2.23355276893943)
--(axis cs:2099,2.2966592877746)
--(axis cs:2199,2.22535373267119)
--(axis cs:2299,2.13649788678173)
--(axis cs:2399,2.23454616924948)
--(axis cs:2499,2.16303064807127)
--(axis cs:2599,2.26385629326659)
--(axis cs:2699,2.1796576407776)
--(axis cs:2799,2.36145744586208)
--(axis cs:2899,2.22243337679715)
--(axis cs:2999,2.21686964758104)
--(axis cs:2999,2.33758527509505)
--(axis cs:2999,2.33758527509505)
--(axis cs:2899,2.34462614010959)
--(axis cs:2799,2.38870765900394)
--(axis cs:2699,2.28858524290465)
--(axis cs:2599,2.38342845290345)
--(axis cs:2499,2.26428327576448)
--(axis cs:2399,2.31233562568002)
--(axis cs:2299,2.22656926333424)
--(axis cs:2199,2.39644299450929)
--(axis cs:2099,2.37984890047308)
--(axis cs:1999,2.32645337272317)
--(axis cs:1899,2.49269709280665)
--(axis cs:1799,2.30476608890329)
--(axis cs:1699,2.3338603191064)
--(axis cs:1599,2.29247538460508)
--(axis cs:1499,2.54876319045585)
--(axis cs:1399,2.35679729176186)
--(axis cs:1299,2.45753545211007)
--(axis cs:1199,2.32767075200303)
--(axis cs:1099,2.38806991522928)
--(axis cs:999,2.48083914643942)
--(axis cs:899,2.44174177480051)
--(axis cs:799,2.46964939081204)
--(axis cs:699,2.45832480607211)
--(axis cs:599,2.75675360256593)
--(axis cs:499,2.76390808556217)
--(axis cs:399,2.95577967719605)
--(axis cs:299,3.5160207496305)
--(axis cs:199,4.14563825842787)
--(axis cs:99,6.75921147549172)
--cycle;

\path [draw=coral, fill=coral, opacity=0.2]
(axis cs:99,7.94512333251475)
--(axis cs:99,7.47365890167715)
--(axis cs:199,4.45568052951039)
--(axis cs:299,3.60010138675685)
--(axis cs:399,3.10815102155729)
--(axis cs:499,2.89254634068331)
--(axis cs:599,2.65795987511248)
--(axis cs:699,2.47971138783482)
--(axis cs:799,2.41104098302879)
--(axis cs:899,2.36762053623377)
--(axis cs:999,2.29531876548242)
--(axis cs:1099,2.29023987252422)
--(axis cs:1199,2.14191363748423)
--(axis cs:1299,2.32658946357825)
--(axis cs:1399,2.31437383367935)
--(axis cs:1499,2.34510653669689)
--(axis cs:1599,2.25034775888077)
--(axis cs:1699,2.16153904431391)
--(axis cs:1799,2.13971587227675)
--(axis cs:1899,2.28995636552057)
--(axis cs:1999,2.22817614366309)
--(axis cs:2099,2.23962127899547)
--(axis cs:2199,2.21829021442505)
--(axis cs:2299,2.10709666383696)
--(axis cs:2399,2.19502603411883)
--(axis cs:2499,2.18144467534355)
--(axis cs:2599,2.23775420142739)
--(axis cs:2699,2.09496998293825)
--(axis cs:2799,2.20503692176423)
--(axis cs:2899,2.14878882244146)
--(axis cs:2999,2.1634222764759)
--(axis cs:2999,2.31708080579949)
--(axis cs:2999,2.31708080579949)
--(axis cs:2899,2.36830102130854)
--(axis cs:2799,2.42303140613952)
--(axis cs:2699,2.24509871499113)
--(axis cs:2599,2.30017879055412)
--(axis cs:2499,2.31197091875741)
--(axis cs:2399,2.31848729252607)
--(axis cs:2299,2.27277673112916)
--(axis cs:2199,2.25461840164093)
--(axis cs:2099,2.4238357570229)
--(axis cs:1999,2.50055942247613)
--(axis cs:1899,2.49241908938208)
--(axis cs:1799,2.33432606650498)
--(axis cs:1699,2.48442069060278)
--(axis cs:1599,2.37070283735642)
--(axis cs:1499,2.44307083432819)
--(axis cs:1399,2.4537284545668)
--(axis cs:1299,2.60764110244653)
--(axis cs:1199,2.40608574453481)
--(axis cs:1099,2.52521604101948)
--(axis cs:999,2.6635707189231)
--(axis cs:899,2.5791639600736)
--(axis cs:799,2.57454077260933)
--(axis cs:699,2.61967280081722)
--(axis cs:599,2.90896815394788)
--(axis cs:499,3.01061578108945)
--(axis cs:399,3.27004056875185)
--(axis cs:299,3.96600898578649)
--(axis cs:199,4.88164957341014)
--(axis cs:99,7.94512333251475)
--cycle;

\addplot [line width=2pt, steelblue, opacity=0.9]
table[row sep=\\]{
99 6.38261079788208 \\
199 3.91127955913544 \\
299 3.27151757478714 \\
399 2.81653672456741 \\
499 2.65213257074356 \\
599 2.576675593853 \\
699 2.37834537029266 \\
799 2.33040565252304 \\
899 2.40012854337692 \\
999 2.38566595315933 \\
1099 2.29720342159271 \\
1199 2.22570532560349 \\
1299 2.3920396566391 \\
1399 2.33918076753616 \\
1499 2.39195722341537 \\
1599 2.26397854089737 \\
1699 2.26233065128326 \\
1799 2.23340851068497 \\
1899 2.403036236763 \\
1999 2.2800030708313 \\
2099 2.33825409412384 \\
2199 2.31089836359024 \\
2299 2.18153357505798 \\
2399 2.27344089746475 \\
2499 2.21365696191788 \\
2599 2.32364237308502 \\
2699 2.23412144184113 \\
2799 2.37508255243301 \\
2899 2.28352975845337 \\
2999 2.27722746133804 \\
};
\addlegendentry{$\eta=0.0$ (baseline)}
\addplot [line width=2pt, coral, opacity=0.9]
table[row sep=\\]{
99 7.70939111709595 \\
199 4.66866505146027 \\
299 3.78305518627167 \\
399 3.18909579515457 \\
499 2.95158106088638 \\
599 2.78346401453018 \\
699 2.54969209432602 \\
799 2.49279087781906 \\
899 2.47339224815369 \\
999 2.47944474220276 \\
1099 2.40772795677185 \\
1199 2.27399969100952 \\
1299 2.46711528301239 \\
1399 2.38405114412308 \\
1499 2.39408868551254 \\
1599 2.31052529811859 \\
1699 2.32297986745834 \\
1799 2.23702096939087 \\
1899 2.39118772745132 \\
1999 2.36436778306961 \\
2099 2.33172851800919 \\
2199 2.23645430803299 \\
2299 2.18993669748306 \\
2399 2.25675666332245 \\
2499 2.24670779705048 \\
2599 2.26896649599075 \\
2699 2.17003434896469 \\
2799 2.31403416395187 \\
2899 2.258544921875 \\
2999 2.2402515411377 \\
};
\addlegendentry{$\eta=0.8$ (ours)}
\end{axis}

\end{tikzpicture}
         \caption{FIDs on CIFAR10}
        \label{fig:cifar10_few}
    \end{subfigure}\hspace{-1em}%
    \begin{subfigure}{0.33\linewidth}
        \raggedleft
\begin{tikzpicture}

\definecolor{coral}{RGB}{255,127,80}
\definecolor{darkgray176}{RGB}{176,176,176}
\definecolor{lightgray204}{RGB}{204,204,204}
\definecolor{steelblue}{RGB}{70,130,180}

\begin{axis}[
height=\figureheight,
legend cell align={left},
legend style={fill opacity=0.8, draw opacity=1, text opacity=1, draw=lightgray204},
tick align=outside,
tick pos=left,
width=\figurewidth,
x grid style={darkgray176},
xmajorgrids,
xmin=5, xmax=2095,
xtick style={color=black},
y grid style={darkgray176},
ymajorgrids,
ymin=1.6, ymax=3,
ytick style={color=black}
]
\path [draw=steelblue, fill=steelblue, opacity=0.2]
(axis cs:100,3.34018010164215)
--(axis cs:100,3.20618813966797)
--(axis cs:200,2.50771454451762)
--(axis cs:300,2.31707110835455)
--(axis cs:400,2.16879416483647)
--(axis cs:500,2.07506177085003)
--(axis cs:600,1.82076300274404)
--(axis cs:700,1.84886501275994)
--(axis cs:800,1.8619719782864)
--(axis cs:900,1.75408535564636)
--(axis cs:1000,1.73017796107389)
--(axis cs:1100,1.92559717421066)
--(axis cs:1200,1.72006499685212)
--(axis cs:1300,2.01216670081448)
--(axis cs:1400,1.87958519762545)
--(axis cs:1500,1.69663324809609)
--(axis cs:1600,1.8920483765574)
--(axis cs:1700,2.02727104633224)
--(axis cs:1800,1.95430925523181)
--(axis cs:1900,1.89705630019533)
--(axis cs:2000,1.86178258662925)
--(axis cs:2000,2.58837171787515)
--(axis cs:2000,2.58837171787515)
--(axis cs:1900,2.34563092514647)
--(axis cs:1800,2.3898205455647)
--(axis cs:1700,2.27780233413804)
--(axis cs:1600,2.11140958690924)
--(axis cs:1500,2.09029176973762)
--(axis cs:1400,2.04800976211042)
--(axis cs:1300,2.07794074013401)
--(axis cs:1200,2.25261056505279)
--(axis cs:1100,2.14262457366455)
--(axis cs:1000,2.05319605044268)
--(axis cs:900,1.84477109348084)
--(axis cs:800,1.96400880422245)
--(axis cs:700,2.13955148017906)
--(axis cs:600,2.17122351039377)
--(axis cs:500,2.42267921311298)
--(axis cs:400,2.54877864342921)
--(axis cs:300,2.59998962448694)
--(axis cs:200,2.76029070736684)
--(axis cs:100,3.34018010164215)
--cycle;

\path [draw=coral, fill=coral, opacity=0.2]
(axis cs:100,3.53293083724506)
--(axis cs:100,3.22205270710461)
--(axis cs:200,2.62881220930972)
--(axis cs:300,2.38554901058626)
--(axis cs:400,2.2144455851823)
--(axis cs:500,2.28656205316193)
--(axis cs:600,1.95984448092598)
--(axis cs:700,1.91193836747508)
--(axis cs:800,1.87003036958006)
--(axis cs:900,1.73912753670357)
--(axis cs:1000,1.78755342353133)
--(axis cs:1100,1.80109322480204)
--(axis cs:1200,1.86526111200137)
--(axis cs:1300,1.89294354434912)
--(axis cs:1400,1.67940299025478)
--(axis cs:1500,1.73047513326331)
--(axis cs:1600,1.78695009797649)
--(axis cs:1700,1.78340527475778)
--(axis cs:1800,1.83068925724794)
--(axis cs:1900,1.72165659223595)
--(axis cs:2000,1.62334800775287)
--(axis cs:2000,2.07274531309369)
--(axis cs:2000,2.07274531309369)
--(axis cs:1900,1.82142117465934)
--(axis cs:1800,1.91544252051542)
--(axis cs:1700,1.89380225478705)
--(axis cs:1600,1.89222712665958)
--(axis cs:1500,1.83532493749932)
--(axis cs:1400,1.96115495451985)
--(axis cs:1300,2.00593151334817)
--(axis cs:1200,2.10769596264081)
--(axis cs:1100,2.1739874441242)
--(axis cs:1000,2.05644763123247)
--(axis cs:900,1.89854941756584)
--(axis cs:800,1.95602539557192)
--(axis cs:700,2.10447132767338)
--(axis cs:600,2.19869731766563)
--(axis cs:500,2.36793127874724)
--(axis cs:400,2.90371239764394)
--(axis cs:300,2.68709569995451)
--(axis cs:200,2.81154285317502)
--(axis cs:100,3.53293083724506)
--cycle;

\addplot [line width=2pt, steelblue, opacity=0.9]
table[row sep=\\]{
100 3.27318412065506 \\
200 2.63400262594223 \\
300 2.45853036642075 \\
400 2.35878640413284 \\
500 2.24887049198151 \\
600 1.99599325656891 \\
700 1.9942082464695 \\
800 1.91299039125443 \\
900 1.7994282245636 \\
1000 1.89168700575829 \\
1100 2.03411087393761 \\
1200 1.98633778095245 \\
1300 2.04505372047424 \\
1400 1.96379747986794 \\
1500 1.89346250891685 \\
1600 2.00172898173332 \\
1700 2.15253669023514 \\
1800 2.17206490039825 \\
1900 2.1213436126709 \\
2000 2.2250771522522 \\
};
\addlegendentry{$\eta=0.0$ (baseline)}
\addplot [line width=2pt, coral, opacity=0.9]
table[row sep=\\]{
100 3.37749177217484 \\
200 2.72017753124237 \\
300 2.53632235527039 \\
400 2.55907899141312 \\
500 2.32724666595459 \\
600 2.07927089929581 \\
700 2.00820484757423 \\
800 1.91302788257599 \\
900 1.8188384771347 \\
1000 1.9220005273819 \\
1100 1.98754033446312 \\
1200 1.98647853732109 \\
1300 1.94943752884865 \\
1400 1.82027897238731 \\
1500 1.78290003538132 \\
1600 1.83958861231804 \\
1700 1.83860376477242 \\
1800 1.87306588888168 \\
1900 1.77153888344765 \\
2000 1.84804666042328 \\
};
\addlegendentry{$\eta=0.8$ (ours)}
\end{axis}

\end{tikzpicture}
         \caption{FIDs on CelebA-50K}
        \label{fig:celeba_50k}
    \end{subfigure}\hspace{-1em}%
    \begin{subfigure}{0.33\linewidth}
        \raggedleft
\begin{tikzpicture}

\definecolor{coral}{RGB}{255,127,80}
\definecolor{darkgray176}{RGB}{176,176,176}
\definecolor{lightgray204}{RGB}{204,204,204}
\definecolor{steelblue}{RGB}{70,130,180}

\begin{axis}[
height=\figureheight,
legend cell align={left},
legend style={fill opacity=0.8, draw opacity=1, text opacity=1, draw=lightgray204},
tick align=outside,
tick pos=left,
width=\figurewidth,
x grid style={darkgray176},
xmajorgrids,
xmin=109, xmax=2089,
xtick style={color=black},
y grid style={darkgray176},
ymajorgrids,
ymin=1.29309926182381, ymax=1.67574696031278,
ytick style={color=black},
ytick={1.25,1.3,1.35,1.4,1.45,1.5,1.55,1.6,1.65,1.7},
yticklabels={1.2,1.3,1.4,1.4,1.4,1.5,1.6,1.6,1.6,1.7}
]
\path [draw=steelblue, fill=steelblue, opacity=0.2]
(axis cs:199,1.65835388310873)
--(axis cs:199,1.50354782930887)
--(axis cs:399,1.40148264651084)
--(axis cs:599,1.36849796673671)
--(axis cs:799,1.38771086674447)
--(axis cs:999,1.3264936820416)
--(axis cs:1199,1.40595657919314)
--(axis cs:1399,1.33444516222254)
--(axis cs:1599,1.31273160363909)
--(axis cs:1799,1.32255032363009)
--(axis cs:1999,1.34023773542787)
--(axis cs:1999,1.45053708680724)
--(axis cs:1999,1.45053708680724)
--(axis cs:1799,1.42923972306181)
--(axis cs:1599,1.45810074423078)
--(axis cs:1399,1.44343846280798)
--(axis cs:1199,1.47635700608824)
--(axis cs:999,1.4870476826282)
--(axis cs:799,1.54688424128776)
--(axis cs:599,1.53623607256993)
--(axis cs:399,1.58912990803933)
--(axis cs:199,1.65835388310873)
--cycle;

\path [draw=coral, fill=coral, opacity=0.2]
(axis cs:199,1.57636335689052)
--(axis cs:199,1.4217077771999)
--(axis cs:399,1.39952754110066)
--(axis cs:599,1.36976666266064)
--(axis cs:799,1.31153983296418)
--(axis cs:999,1.35390194906376)
--(axis cs:1199,1.38954268048133)
--(axis cs:1399,1.35815161454414)
--(axis cs:1599,1.31049233902786)
--(axis cs:1799,1.32218252746312)
--(axis cs:1999,1.37432921556878)
--(axis cs:1999,1.41365855308128)
--(axis cs:1999,1.41365855308128)
--(axis cs:1799,1.44204200180324)
--(axis cs:1599,1.46430307398942)
--(axis cs:1399,1.45294511330392)
--(axis cs:1199,1.49470623185311)
--(axis cs:999,1.48506197677471)
--(axis cs:799,1.5485056596849)
--(axis cs:599,1.50794850772282)
--(axis cs:399,1.53475124508174)
--(axis cs:199,1.57636335689052)
--cycle;

\addplot [line width=2pt, steelblue, opacity=0.9]
table[row sep=\\]{
199 1.5809508562088 \\
399 1.49530627727509 \\
599 1.45236701965332 \\
799 1.46729755401611 \\
999 1.4067706823349 \\
1199 1.44115679264069 \\
1399 1.38894181251526 \\
1599 1.38541617393494 \\
1799 1.37589502334595 \\
1999 1.39538741111755 \\
};
\addlegendentry{$\eta=0.0$ (baseline)}
\addplot [line width=2pt, coral, opacity=0.9]
table[row sep=\\]{
199 1.49903556704521 \\
399 1.4671393930912 \\
599 1.43885758519173 \\
799 1.43002274632454 \\
999 1.41948196291924 \\
1199 1.44212445616722 \\
1399 1.40554836392403 \\
1599 1.38739770650864 \\
1799 1.38211226463318 \\
1999 1.39399388432503 \\
};
\addlegendentry{$\eta=0.8$ (ours)}
\end{axis}

\end{tikzpicture}
         \caption{FIDs on LSUN Bedroom}
        \label{fig:lsun_plot}
    \end{subfigure}
    \caption{\textbf{Effect of masking during training.} When $\eta=0$, the model reduces to the baseline flow matching~\citep{lipman2024flowmatchingguidecode}. Across three datasets, SSD with up to 80\% masked pixels can still achieve comparable performance as the baseline. Notably, the baseline in~\cref{fig:celeba_50k} eventually explodes while SSD remains stable. The shaded regions indicate the 95\% confidence interval over four runs.\looseness-1}
    \label{fig:full_plot}
\end{figure*}
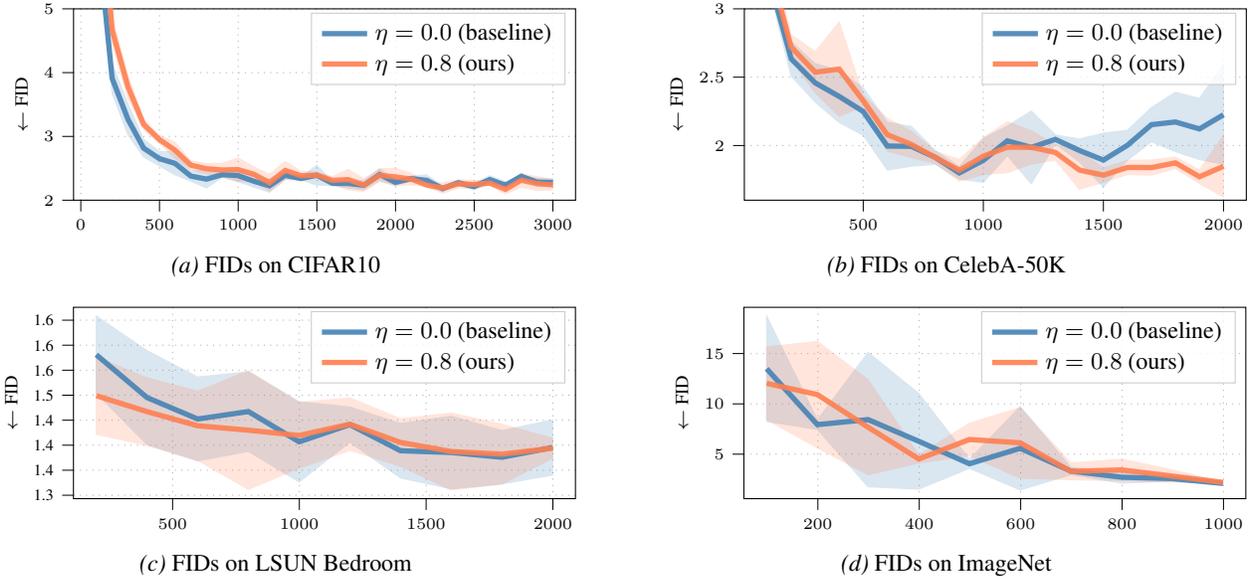

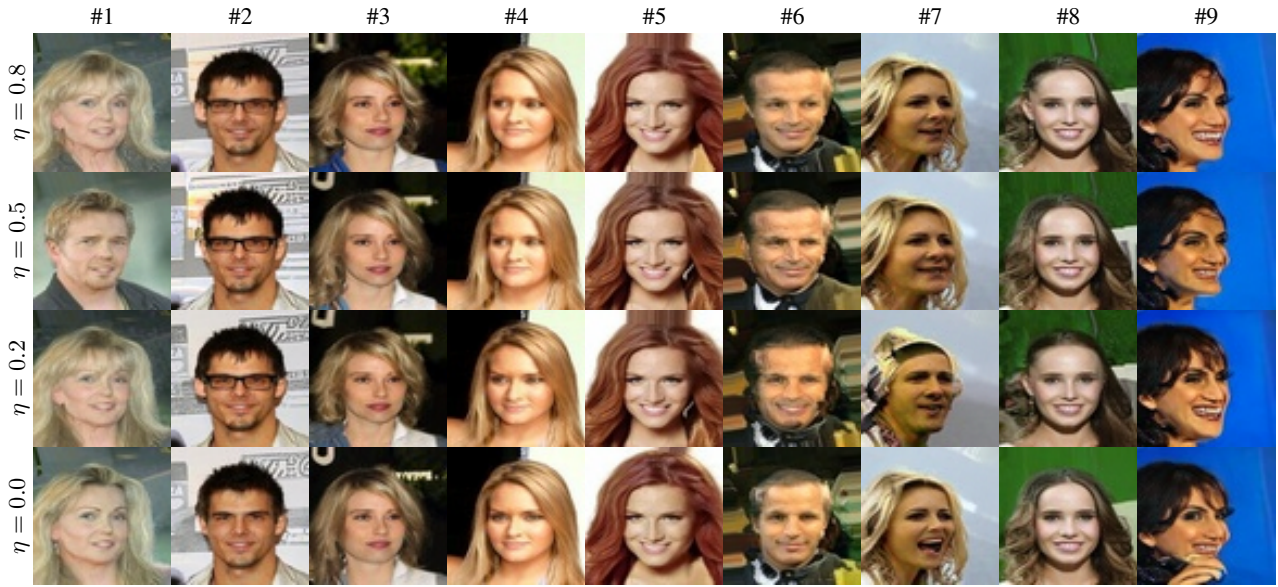
\begin{figure*}[t]
  \centering\footnotesize
  \begin{tikzpicture}[inner sep=0] 
    
   \setlength{\figurewidth}{.107\textwidth}
   
   \node (1-1) at (1*\figurewidth,1*\figurewidth) {\includegraphics[width=\figurewidth]{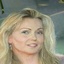}};
\node[rotate=90,font=\strut,minimum width=\figurewidth,align=center,anchor=south] at (1-1.west) {$\eta=0.0$};
\node (1-2) at (2*\figurewidth,1*\figurewidth) {\includegraphics[width=\figurewidth]{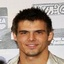}};
\node (1-3) at (3*\figurewidth,1*\figurewidth) {\includegraphics[width=\figurewidth]{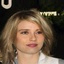}};
\node (1-4) at (4*\figurewidth,1*\figurewidth) {\includegraphics[width=\figurewidth]{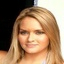}};
\node (1-5) at (5*\figurewidth,1*\figurewidth) {\includegraphics[width=\figurewidth]{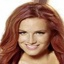}};
\node (1-6) at (6*\figurewidth,1*\figurewidth) {\includegraphics[width=\figurewidth]{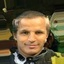}};
\node (1-7) at (7*\figurewidth,1*\figurewidth) {\includegraphics[width=\figurewidth]{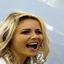}};
\node (1-8) at (8*\figurewidth,1*\figurewidth) {\includegraphics[width=\figurewidth]{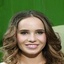}};
\node (1-9) at (9*\figurewidth,1*\figurewidth) {\includegraphics[width=\figurewidth]{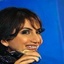}};
\node (2-1) at (1*\figurewidth,2*\figurewidth) {\includegraphics[width=\figurewidth]{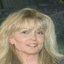}};
\node[rotate=90,font=\strut,minimum width=\figurewidth,align=center,anchor=south] at (2-1.west) {$\eta=0.2$};
\node (2-2) at (2*\figurewidth,2*\figurewidth) {\includegraphics[width=\figurewidth]{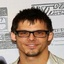}};
\node (2-3) at (3*\figurewidth,2*\figurewidth) {\includegraphics[width=\figurewidth]{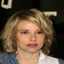}};
\node (2-4) at (4*\figurewidth,2*\figurewidth) {\includegraphics[width=\figurewidth]{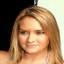}};
\node (2-5) at (5*\figurewidth,2*\figurewidth) {\includegraphics[width=\figurewidth]{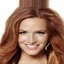}};
\node (2-6) at (6*\figurewidth,2*\figurewidth) {\includegraphics[width=\figurewidth]{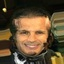}};
\node (2-7) at (7*\figurewidth,2*\figurewidth) {\includegraphics[width=\figurewidth]{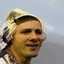}};
\node (2-8) at (8*\figurewidth,2*\figurewidth) {\includegraphics[width=\figurewidth]{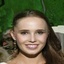}};
\node (2-9) at (9*\figurewidth,2*\figurewidth) {\includegraphics[width=\figurewidth]{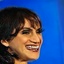}};
\node (3-1) at (1*\figurewidth,3*\figurewidth) {\includegraphics[width=\figurewidth]{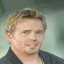}};
\node[rotate=90,font=\strut,minimum width=\figurewidth,align=center,anchor=south] at (3-1.west) {$\eta=0.5$};
\node (3-2) at (2*\figurewidth,3*\figurewidth) {\includegraphics[width=\figurewidth]{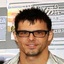}};
\node (3-3) at (3*\figurewidth,3*\figurewidth) {\includegraphics[width=\figurewidth]{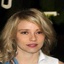}};
\node (3-4) at (4*\figurewidth,3*\figurewidth) {\includegraphics[width=\figurewidth]{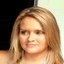}};
\node (3-5) at (5*\figurewidth,3*\figurewidth) {\includegraphics[width=\figurewidth]{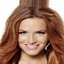}};
\node (3-6) at (6*\figurewidth,3*\figurewidth) {\includegraphics[width=\figurewidth]{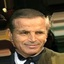}};
\node (3-7) at (7*\figurewidth,3*\figurewidth) {\includegraphics[width=\figurewidth]{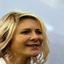}};
\node (3-8) at (8*\figurewidth,3*\figurewidth) {\includegraphics[width=\figurewidth]{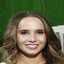}};
\node (3-9) at (9*\figurewidth,3*\figurewidth) {\includegraphics[width=\figurewidth]{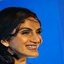}};
\node (4-1) at (1*\figurewidth,4*\figurewidth) {\includegraphics[width=\figurewidth]{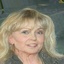}};
\node[font=\strut,minimum width=\figurewidth,align=center,anchor=south] at (4-1.north) {\#1};
\node[rotate=90,font=\strut,minimum width=\figurewidth,align=center,anchor=south] at (4-1.west) {$\eta=0.8$};
\node (4-2) at (2*\figurewidth,4*\figurewidth) {\includegraphics[width=\figurewidth]{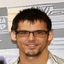}};
\node[font=\strut,minimum width=\figurewidth,align=center,anchor=south] at (4-2.north) {\#2};
\node (4-3) at (3*\figurewidth,4*\figurewidth) {\includegraphics[width=\figurewidth]{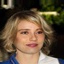}};
\node[font=\strut,minimum width=\figurewidth,align=center,anchor=south] at (4-3.north) {\#3};
\node (4-4) at (4*\figurewidth,4*\figurewidth) {\includegraphics[width=\figurewidth]{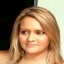}};
\node[font=\strut,minimum width=\figurewidth,align=center,anchor=south] at (4-4.north) {\#4};
\node (4-5) at (5*\figurewidth,4*\figurewidth) {\includegraphics[width=\figurewidth]{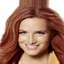}};
\node[font=\strut,minimum width=\figurewidth,align=center,anchor=south] at (4-5.north) {\#5};
\node (4-6) at (6*\figurewidth,4*\figurewidth) {\includegraphics[width=\figurewidth]{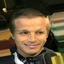}};
\node[font=\strut,minimum width=\figurewidth,align=center,anchor=south] at (4-6.north) {\#6};
\node (4-7) at (7*\figurewidth,4*\figurewidth) {\includegraphics[width=\figurewidth]{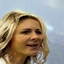}};
\node[font=\strut,minimum width=\figurewidth,align=center,anchor=south] at (4-7.north) {\#7};
\node (4-8) at (8*\figurewidth,4*\figurewidth) {\includegraphics[width=\figurewidth]{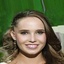}};
\node[font=\strut,minimum width=\figurewidth,align=center,anchor=south] at (4-8.north) {\#8};
\node (4-9) at (9*\figurewidth,4*\figurewidth) {\includegraphics[width=\figurewidth]{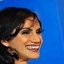}};
\node[font=\strut,minimum width=\figurewidth,align=center,anchor=south] at (4-9.north) {\#9};
\end{tikzpicture}%
  \label{fig:sample_compare_celeba}
  \caption{\textbf{Masking leads to more realistic generations.} Non-curated samples from models trained on CelebA. Sample \#1 shows finer neck wrinkles at $\eta=0.8$. In sample \#6, an unrealistic bulge on the head at $\eta=0$ gradually disappears as $\eta$ increases. A similar artifact in sample \#9 of the baseline is also removed by masking.}
  \vspace{-1em}
  \label{fig:full_sample_compare}
\end{figure*}

\subsection{Masking achieves competitive FID scores (Q1)}

\begin{wraptable}{r}{0.5\textwidth}
    \centering
    \footnotesize
    \vspace{-\baselineskip} %
    \caption{FID ($\sigma$) ($\downarrow$) on different datasets.}
    \begin{tabular}{lcc}
        \toprule
        Dataset & Baseline (FM/DiT) & SSD (ours) \\
        \midrule
        CIFAR10    & 2.28 (0.04) & \textbf{2.24 (0.05)} \\
        LSUN       & 1.40 (0.04) & \textbf{1.39 (0.01)} \\
        CelebA-50K & 2.23 (0.23) & \textbf{1.85 (0.14)} \\
        CelebA-10K & 13.72 (--)   & \textbf{5.10 (--)} \\
        ImageNet   & 4.98 (--)    & \textbf{4.83 (--)} \\
        \bottomrule
    \end{tabular}
    \vspace{-1em}
    \label{tab:fid_smd}
\end{wraptable}

We plot the evaluation set FID scores across training epochs to visualize the learning dynamics on three datasets, with curves averaged over four runs.~\cref{fig:full_plot} shows that models with an 80\% masking ratio generally achieve lower comparable scores to the unmasked baselines. Notably, in \cref{fig:celeba_50k}, the baseline FM model diverges after 1500 epochs, whereas our method with $\eta = 0.8$ remains stable. This phenomenon is further validated in our CelebA-10K experiments as shown in~\cref{fig:teaser}, where the baseline exhibits even more severe divergence. We also report the final evaluation FID scores on all datasets in~\cref{tab:fid_smd}. For ImageNet, we set SSD with $\eta=0.98$ and train the models for 180k steps. At CFG=1.5, SSD slightly improves full-ImageNet FID from 4.98 to 4.83. On a 100K-image subset, the improvement is larger: after 200K steps, FID decreases from 25.99 to 15.58 with $\eta=0.98$ (\cref{tab:imagenet_data_budget}). These results suggest that the benefit depends on both dataset size and training budget.

\begin{wraptable}{r}{0.5\textwidth}
    \centering
    \footnotesize
    \vspace{-\baselineskip} %
    \caption{\textbf{Masking mitigates memorization.} FID ($\downarrow$) and average $\ell_2$ distance ($\uparrow$) to the nearest training samples over 10k generated images on CelebA-10K. Larger $\ell_2$ indicates less memorization.}
    \begin{tabular}{lcc}
        \toprule
        Method & Baseline ($\eta=0.0$) & Ours ($\eta=0.98$) \\ 
        \midrule
        FID ($\downarrow$) & 13.72 & \textbf{5.10} \\
        $\ell_2$ distance ($\uparrow$) & 42.32 & \textbf{46.02} \\
        \bottomrule
    \end{tabular}
    \vspace{-1em}
    \label{tab:celeba_10k_memorization}
\end{wraptable}
We visualize non-curated generated samples from different models in \cref{fig:full_sample_compare}. The same noises are fed into each model to observe how the generated images look like over different mask ratios. While the generated images from the same noise generally appear similar, we observe improved spatial consistency as $\eta$ increases. For instance, the unrealistic bulge on the head in sample $\#6$ gradually diminishes as $\eta$ increases. Moreover, sample $\#1$ shows more detailed neck wrinkles with $\eta=0.8$ compared to the image with $\eta=0$.

\subsection{Masking mitigates training divergence (Q2)}
We further compare SSD with the baseline on CelebA-10K, which contains 10k images. As shown in~\cref{fig:teaser}, SSD performs well even with masking ratios as high as 98\%, and it effectively mitigates the severe divergence observed in baseline models trained for a long time without masking or with lower masking ratios such as $\eta = 0.5$.

\subsection{Masking mitigates memorization (Q3)}

\begin{wrapfigure}{r}{0.5\textwidth}
    \centering
    \vspace{-\baselineskip} %
    \includegraphics[width=0.95\linewidth]{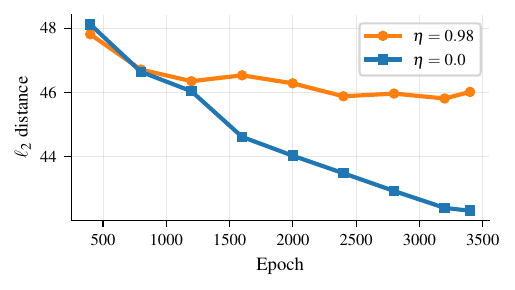}
    \vspace{-1em}
    \caption{Average $\ell_2$ distance to the nearest training samples over training iterations on CelebA-10K. Larger values indicate less memorization.}
    \label{fig:celeba10k_l2}
    \vspace{-1em}
\end{wrapfigure}
We also observe that the proposed masking strategy can mitigate the memorization problem in diffusion model training, which is particularly critical for small datasets. We generate 10k images using models trained on CelebA-10K and compute the average of $\ell_2$ distances to their nearest neighbors in the training set. The results in~\cref{tab:celeba_10k_memorization} show that, with $\eta = 0.98$, SSD achieves larger $\ell_2$ distances compared to the baseline without masking, indicating that SSD generates images that are more distinct from the training samples, while maintaining comparable or even lower FID, as illustrated in~\cref{fig:teaser}.

~\cref{fig:celeba10k_l2} shows the $\ell_2$ distance evolution during training, demonstrating that SSD effectively mitigates memorization. Following~\citet{bonnaire2025diffusion}, we compute the memorization distance as $d_{\textrm{mem}}=|x_{\tau} - a^{\mu}|_2$, where $x_{\tau}$ denotes a generated sample and $a^{\mu}$ its nearest neighbor in the training set. Additional comparisons with dropout and MDT~\citep{gao2023masked} are provided in~\cref{tab:mdt_dropout_comparison} in the appendix.

\subsection{Masking promotes essential contextual usage (Q4)}\label{sec:gradient_sensitivity}
The gradient sensitivity~\citep{niedoba2025towards} reflects how pixel positions are correlated when diffusion models generate pixels. For each timestep $t$, the gradient sensitivity heatmap is computed as
\begin{equation}
S(q, t) = \mathbb{E}_{\vz \sim p_t(\vx^{(i)}, \vz)} \bigg[ \sum_{c=1}^{3} \left| \nabla_{\vz_c} \vv_\theta(\vz, t)_{q,c} \right| \bigg],
\label{equ:sensitivity_define}
\end{equation}
where $\vv_\theta(\vz, t)_{q,c}$ denotes the output of the vector field network at pixel position $q$ and image channel $c$. In our experiments, we compute gradients at the central point of each image. A higher gradient usually means a stronger influence from the location when generating a pixel.

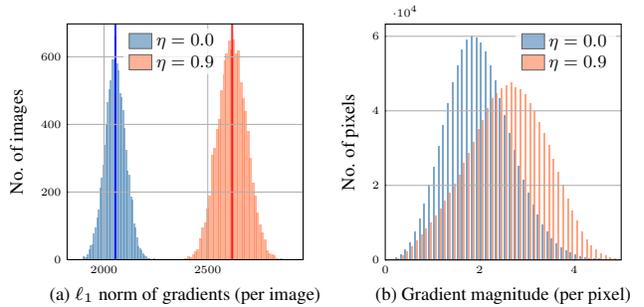
\begin{figure}[t]
  \centering\footnotesize
  \setlength{\figurewidth}{.38\columnwidth}
  \setlength{\figureheight}{.5\figurewidth}

  \pgfplotsset{
      axis on top,
      scale only axis,
      xticklabel style={font=\tiny,scale=1.},
      yticklabel style={font=\tiny,scale=1.},
      ylabel near ticks,
      xlabel near ticks,
      tick align=outside,
      tick pos=both,
      tick style={major tick length=0pt, minor tick length=0pt},
      title style={font=\strut\scriptsize,yshift=-6pt},
    }
\hspace{-10pt}
  \begin{subfigure}[t]{.47\columnwidth}
    \centering
\begin{tikzpicture}

\definecolor{coral}{RGB}{255,127,80}
\definecolor{darkgray176}{RGB}{176,176,176}
\definecolor{steelblue}{RGB}{70,130,180}

\begin{axis}[
height=\figureheight,
tick align=outside,
tick pos=left,
width=\figurewidth,
x grid style={darkgray176},
xlabel={(a) $\ell_1$ norm of gradients (per image)},
xmajorgrids,
xmin=1822.10256347656, xmax=2959.58151855469,
xtick style={color=black},
y grid style={darkgray176},
ylabel={No. of images},
ymajorgrids,
ymin=0, ymax=697.2,
ytick style={color=black},
xlabel style={font=\scriptsize},
ylabel style={font=\scriptsize},
legend style={at={(0.45,0.98)},
    anchor=north,draw=none,text opacity=1,inner xsep=2pt, inner ysep=0.5pt, nodes={inner sep=2pt, text depth=0.1em},fill=white,fill opacity=0.8},legend style={nodes={scale=0.75, transform shape}},
legend image post style={xscale=0.5}
]
\draw[draw=none,fill=steelblue,opacity=0.6] (axis cs:1873.80615234375,0) rectangle (axis cs:1881.22426757812,1);
\draw[draw=none,fill=steelblue,opacity=0.6] (axis cs:1881.22426757812,0) rectangle (axis cs:1888.6423828125,0);
\draw[draw=none,fill=steelblue,opacity=0.6] (axis cs:1888.6423828125,0) rectangle (axis cs:1896.06049804687,2);
\draw[draw=none,fill=steelblue,opacity=0.6] (axis cs:1896.06049804687,0) rectangle (axis cs:1903.47861328125,9);
\draw[draw=none,fill=steelblue,opacity=0.6] (axis cs:1903.47861328125,0) rectangle (axis cs:1910.89672851562,8);
\draw[draw=none,fill=steelblue,opacity=0.6] (axis cs:1910.89672851562,0) rectangle (axis cs:1918.31484375,8);
\draw[draw=none,fill=steelblue,opacity=0.6] (axis cs:1918.31484375,0) rectangle (axis cs:1925.73295898438,15);
\draw[draw=none,fill=steelblue,opacity=0.6] (axis cs:1925.73295898438,0) rectangle (axis cs:1933.15107421875,32);
\draw[draw=none,fill=steelblue,opacity=0.6] (axis cs:1933.15107421875,0) rectangle (axis cs:1940.56918945312,42);
\draw[draw=none,fill=steelblue,opacity=0.6] (axis cs:1940.56918945313,0) rectangle (axis cs:1947.9873046875,46);
\draw[draw=none,fill=steelblue,opacity=0.6] (axis cs:1947.9873046875,0) rectangle (axis cs:1955.40541992187,78);
\draw[draw=none,fill=steelblue,opacity=0.6] (axis cs:1955.40541992187,0) rectangle (axis cs:1962.82353515625,95);
\draw[draw=none,fill=steelblue,opacity=0.6] (axis cs:1962.82353515625,0) rectangle (axis cs:1970.24165039062,125);
\draw[draw=none,fill=steelblue,opacity=0.6] (axis cs:1970.24165039062,0) rectangle (axis cs:1977.659765625,147);
\draw[draw=none,fill=steelblue,opacity=0.6] (axis cs:1977.659765625,0) rectangle (axis cs:1985.07788085938,214);
\draw[draw=none,fill=steelblue,opacity=0.6] (axis cs:1985.07788085938,0) rectangle (axis cs:1992.49599609375,243);
\draw[draw=none,fill=steelblue,opacity=0.6] (axis cs:1992.49599609375,0) rectangle (axis cs:1999.91411132813,281);
\draw[draw=none,fill=steelblue,opacity=0.6] (axis cs:1999.91411132813,0) rectangle (axis cs:2007.3322265625,339);
\draw[draw=none,fill=steelblue,opacity=0.6] (axis cs:2007.3322265625,0) rectangle (axis cs:2014.75034179687,400);
\draw[draw=none,fill=steelblue,opacity=0.6] (axis cs:2014.75034179688,0) rectangle (axis cs:2022.16845703125,467);
\draw[draw=none,fill=steelblue,opacity=0.6] (axis cs:2022.16845703125,0) rectangle (axis cs:2029.58657226562,503);
\draw[draw=none,fill=steelblue,opacity=0.6] (axis cs:2029.58657226562,0) rectangle (axis cs:2037.0046875,531);
\draw[draw=none,fill=steelblue,opacity=0.6] (axis cs:2037.0046875,0) rectangle (axis cs:2044.42280273437,591);
\draw[draw=none,fill=steelblue,opacity=0.6] (axis cs:2044.42280273437,0) rectangle (axis cs:2051.84091796875,591);
\draw[draw=none,fill=steelblue,opacity=0.6] (axis cs:2051.84091796875,0) rectangle (axis cs:2059.25903320312,664);
\draw[draw=none,fill=steelblue,opacity=0.6] (axis cs:2059.25903320312,0) rectangle (axis cs:2066.6771484375,579);
\draw[draw=none,fill=steelblue,opacity=0.6] (axis cs:2066.6771484375,0) rectangle (axis cs:2074.09526367187,559);
\draw[draw=none,fill=steelblue,opacity=0.6] (axis cs:2074.09526367187,0) rectangle (axis cs:2081.51337890625,493);
\draw[draw=none,fill=steelblue,opacity=0.6] (axis cs:2081.51337890625,0) rectangle (axis cs:2088.93149414063,480);
\draw[draw=none,fill=steelblue,opacity=0.6] (axis cs:2088.93149414062,0) rectangle (axis cs:2096.349609375,443);
\draw[draw=none,fill=steelblue,opacity=0.6] (axis cs:2096.349609375,0) rectangle (axis cs:2103.76772460937,390);
\draw[draw=none,fill=steelblue,opacity=0.6] (axis cs:2103.76772460937,0) rectangle (axis cs:2111.18583984375,325);
\draw[draw=none,fill=steelblue,opacity=0.6] (axis cs:2111.18583984375,0) rectangle (axis cs:2118.60395507813,264);
\draw[draw=none,fill=steelblue,opacity=0.6] (axis cs:2118.60395507813,0) rectangle (axis cs:2126.0220703125,250);
\draw[draw=none,fill=steelblue,opacity=0.6] (axis cs:2126.0220703125,0) rectangle (axis cs:2133.44018554687,185);
\draw[draw=none,fill=steelblue,opacity=0.6] (axis cs:2133.44018554688,0) rectangle (axis cs:2140.85830078125,178);
\draw[draw=none,fill=steelblue,opacity=0.6] (axis cs:2140.85830078125,0) rectangle (axis cs:2148.27641601562,106);
\draw[draw=none,fill=steelblue,opacity=0.6] (axis cs:2148.27641601562,0) rectangle (axis cs:2155.69453125,94);
\draw[draw=none,fill=steelblue,opacity=0.6] (axis cs:2155.69453125,0) rectangle (axis cs:2163.11264648438,66);
\draw[draw=none,fill=steelblue,opacity=0.6] (axis cs:2163.11264648437,0) rectangle (axis cs:2170.53076171875,39);
\draw[draw=none,fill=steelblue,opacity=0.6] (axis cs:2170.53076171875,0) rectangle (axis cs:2177.94887695312,33);
\draw[draw=none,fill=steelblue,opacity=0.6] (axis cs:2177.94887695312,0) rectangle (axis cs:2185.3669921875,29);
\draw[draw=none,fill=steelblue,opacity=0.6] (axis cs:2185.3669921875,0) rectangle (axis cs:2192.78510742188,20);
\draw[draw=none,fill=steelblue,opacity=0.6] (axis cs:2192.78510742188,0) rectangle (axis cs:2200.20322265625,13);
\draw[draw=none,fill=steelblue,opacity=0.6] (axis cs:2200.20322265625,0) rectangle (axis cs:2207.62133789062,9);
\draw[draw=none,fill=steelblue,opacity=0.6] (axis cs:2207.62133789062,0) rectangle (axis cs:2215.039453125,4);
\draw[draw=none,fill=steelblue,opacity=0.6] (axis cs:2215.039453125,0) rectangle (axis cs:2222.45756835937,4);
\draw[draw=none,fill=steelblue,opacity=0.6] (axis cs:2222.45756835937,0) rectangle (axis cs:2229.87568359375,1);
\draw[draw=none,fill=steelblue,opacity=0.6] (axis cs:2229.87568359375,0) rectangle (axis cs:2237.29379882813,1);
\draw[draw=none,fill=steelblue,opacity=0.6] (axis cs:2237.29379882812,0) rectangle (axis cs:2244.7119140625,3);
\draw[draw=none,fill=coral,opacity=0.6] (axis cs:2331.79541015625,0) rectangle (axis cs:2343.31706054687,1);
\draw[draw=none,fill=coral,opacity=0.6] (axis cs:2343.31706054687,0) rectangle (axis cs:2354.8387109375,1);
\draw[draw=none,fill=coral,opacity=0.6] (axis cs:2354.8387109375,0) rectangle (axis cs:2366.36036132812,0);
\draw[draw=none,fill=coral,opacity=0.6] (axis cs:2366.36036132812,0) rectangle (axis cs:2377.88201171875,0);
\draw[draw=none,fill=coral,opacity=0.6] (axis cs:2377.88201171875,0) rectangle (axis cs:2389.40366210938,1);
\draw[draw=none,fill=coral,opacity=0.6] (axis cs:2389.40366210938,0) rectangle (axis cs:2400.9253125,5);
\draw[draw=none,fill=coral,opacity=0.6] (axis cs:2400.9253125,0) rectangle (axis cs:2412.44696289062,9);
\draw[draw=none,fill=coral,opacity=0.6] (axis cs:2412.44696289062,0) rectangle (axis cs:2423.96861328125,7);
\draw[draw=none,fill=coral,opacity=0.6] (axis cs:2423.96861328125,0) rectangle (axis cs:2435.49026367187,15);
\draw[draw=none,fill=coral,opacity=0.6] (axis cs:2435.49026367188,0) rectangle (axis cs:2447.0119140625,26);
\draw[draw=none,fill=coral,opacity=0.6] (axis cs:2447.0119140625,0) rectangle (axis cs:2458.53356445313,31);
\draw[draw=none,fill=coral,opacity=0.6] (axis cs:2458.53356445313,0) rectangle (axis cs:2470.05521484375,64);
\draw[draw=none,fill=coral,opacity=0.6] (axis cs:2470.05521484375,0) rectangle (axis cs:2481.57686523437,94);
\draw[draw=none,fill=coral,opacity=0.6] (axis cs:2481.57686523437,0) rectangle (axis cs:2493.098515625,109);
\draw[draw=none,fill=coral,opacity=0.6] (axis cs:2493.098515625,0) rectangle (axis cs:2504.62016601562,151);
\draw[draw=none,fill=coral,opacity=0.6] (axis cs:2504.62016601562,0) rectangle (axis cs:2516.14181640625,189);
\draw[draw=none,fill=coral,opacity=0.6] (axis cs:2516.14181640625,0) rectangle (axis cs:2527.66346679688,310);
\draw[draw=none,fill=coral,opacity=0.6] (axis cs:2527.66346679688,0) rectangle (axis cs:2539.1851171875,330);
\draw[draw=none,fill=coral,opacity=0.6] (axis cs:2539.1851171875,0) rectangle (axis cs:2550.70676757812,386);
\draw[draw=none,fill=coral,opacity=0.6] (axis cs:2550.70676757812,0) rectangle (axis cs:2562.22841796875,415);
\draw[draw=none,fill=coral,opacity=0.6] (axis cs:2562.22841796875,0) rectangle (axis cs:2573.75006835937,501);
\draw[draw=none,fill=coral,opacity=0.6] (axis cs:2573.75006835938,0) rectangle (axis cs:2585.27171875,588);
\draw[draw=none,fill=coral,opacity=0.6] (axis cs:2585.27171875,0) rectangle (axis cs:2596.79336914063,611);
\draw[draw=none,fill=coral,opacity=0.6] (axis cs:2596.79336914063,0) rectangle (axis cs:2608.31501953125,647);
\draw[draw=none,fill=coral,opacity=0.6] (axis cs:2608.31501953125,0) rectangle (axis cs:2619.83666992188,622);
\draw[draw=none,fill=coral,opacity=0.6] (axis cs:2619.83666992188,0) rectangle (axis cs:2631.3583203125,651);
\draw[draw=none,fill=coral,opacity=0.6] (axis cs:2631.3583203125,0) rectangle (axis cs:2642.87997070312,610);
\draw[draw=none,fill=coral,opacity=0.6] (axis cs:2642.87997070312,0) rectangle (axis cs:2654.40162109375,619);
\draw[draw=none,fill=coral,opacity=0.6] (axis cs:2654.40162109375,0) rectangle (axis cs:2665.92327148438,524);
\draw[draw=none,fill=coral,opacity=0.6] (axis cs:2665.92327148438,0) rectangle (axis cs:2677.444921875,478);
\draw[draw=none,fill=coral,opacity=0.6] (axis cs:2677.444921875,0) rectangle (axis cs:2688.96657226563,439);
\draw[draw=none,fill=coral,opacity=0.6] (axis cs:2688.96657226563,0) rectangle (axis cs:2700.48822265625,380);
\draw[draw=none,fill=coral,opacity=0.6] (axis cs:2700.48822265625,0) rectangle (axis cs:2712.00987304687,300);
\draw[draw=none,fill=coral,opacity=0.6] (axis cs:2712.00987304688,0) rectangle (axis cs:2723.5315234375,227);
\draw[draw=none,fill=coral,opacity=0.6] (axis cs:2723.5315234375,0) rectangle (axis cs:2735.05317382813,188);
\draw[draw=none,fill=coral,opacity=0.6] (axis cs:2735.05317382813,0) rectangle (axis cs:2746.57482421875,139);
\draw[draw=none,fill=coral,opacity=0.6] (axis cs:2746.57482421875,0) rectangle (axis cs:2758.09647460938,105);
\draw[draw=none,fill=coral,opacity=0.6] (axis cs:2758.09647460938,0) rectangle (axis cs:2769.618125,81);
\draw[draw=none,fill=coral,opacity=0.6] (axis cs:2769.618125,0) rectangle (axis cs:2781.13977539062,59);
\draw[draw=none,fill=coral,opacity=0.6] (axis cs:2781.13977539062,0) rectangle (axis cs:2792.66142578125,28);
\draw[draw=none,fill=coral,opacity=0.6] (axis cs:2792.66142578125,0) rectangle (axis cs:2804.18307617188,19);
\draw[draw=none,fill=coral,opacity=0.6] (axis cs:2804.18307617188,0) rectangle (axis cs:2815.7047265625,17);
\draw[draw=none,fill=coral,opacity=0.6] (axis cs:2815.7047265625,0) rectangle (axis cs:2827.22637695313,12);
\draw[draw=none,fill=coral,opacity=0.6] (axis cs:2827.22637695313,0) rectangle (axis cs:2838.74802734375,4);
\draw[draw=none,fill=coral,opacity=0.6] (axis cs:2838.74802734375,0) rectangle (axis cs:2850.26967773437,2);
\draw[draw=none,fill=coral,opacity=0.6] (axis cs:2850.26967773437,0) rectangle (axis cs:2861.791328125,3);
\draw[draw=none,fill=coral,opacity=0.6] (axis cs:2861.791328125,0) rectangle (axis cs:2873.31297851563,0);
\draw[draw=none,fill=coral,opacity=0.6] (axis cs:2873.31297851563,0) rectangle (axis cs:2884.83462890625,0);
\draw[draw=none,fill=coral,opacity=0.6] (axis cs:2884.83462890625,0) rectangle (axis cs:2896.35627929688,1);
\draw[draw=none,fill=coral,opacity=0.6] (axis cs:2896.35627929688,0) rectangle (axis cs:2907.8779296875,1);
\addlegendimage{area legend,
  draw=none,
  fill=steelblue,
  opacity=0.6
}
\addlegendentry{$\eta=0.0$}

\addlegendimage{area legend,
  draw=none,
  fill=coral,
  opacity=0.6
}
\addlegendentry{$\eta=0.9$}
\addplot [thick, blue, opacity=0.8]
table[row sep=\\]{
2054.78466046143 0 \\
2054.78466046143 697.2 \\
};
\addplot [thick, red, opacity=0.8]
table[row sep=\\]{
2617.83427099609 0 \\
2617.83427099609 697.2 \\
};
\end{axis}

\end{tikzpicture}

     \label{fig:grads_l1}
  \end{subfigure}
  \hspace{0.005\columnwidth}
  \begin{subfigure}[t]{.47\columnwidth}
    \centering
\begin{tikzpicture}

\definecolor{coral}{RGB}{255,127,80}
\definecolor{darkgray176}{RGB}{176,176,176}
\definecolor{steelblue}{RGB}{70,130,180}

\begin{axis}[
height=\figureheight,
tick align=outside,
tick pos=left,
width=\figurewidth,
x grid style={darkgray176},
xlabel={(b) Gradient magnitude (per pixel)},
xmin=0, xmax=5,
xtick style={color=black},
y grid style={darkgray176},
ylabel={No. of pixels},
ymajorgrids,
ymin=0, ymax=63261.45,
ytick style={color=black},
xlabel style={font=\scriptsize},
ylabel style={font=\scriptsize},
legend style={draw=none,text opacity=1,inner xsep=2pt, inner ysep=0.5pt, nodes={inner sep=2pt, text depth=0.1em},fill=white,fill opacity=0.8},legend style={nodes={scale=0.75, transform shape}},
legend image post style={xscale=0.5}
]
\draw[draw=none,fill=steelblue,opacity=0.7,very thin] (axis cs:0.01,0) rectangle (axis cs:0.05,28);
\draw[draw=none,fill=steelblue,opacity=0.7,very thin] (axis cs:0.11,0) rectangle (axis cs:0.15,192);
\draw[draw=none,fill=steelblue,opacity=0.7,very thin] (axis cs:0.21,0) rectangle (axis cs:0.25,583);
\draw[draw=none,fill=steelblue,opacity=0.7,very thin] (axis cs:0.31,0) rectangle (axis cs:0.35,1393);
\draw[draw=none,fill=steelblue,opacity=0.7,very thin] (axis cs:0.41,0) rectangle (axis cs:0.45,2691);
\draw[draw=none,fill=steelblue,opacity=0.7,very thin] (axis cs:0.51,0) rectangle (axis cs:0.55,4789);
\draw[draw=none,fill=steelblue,opacity=0.7,very thin] (axis cs:0.61,0) rectangle (axis cs:0.65,7508);
\draw[draw=none,fill=steelblue,opacity=0.7,very thin] (axis cs:0.71,0) rectangle (axis cs:0.75,11153);
\draw[draw=none,fill=steelblue,opacity=0.7,very thin] (axis cs:0.81,0) rectangle (axis cs:0.85,15140);
\draw[draw=none,fill=steelblue,opacity=0.7,very thin] (axis cs:0.91,0) rectangle (axis cs:0.95,20010);
\draw[draw=none,fill=steelblue,opacity=0.7,very thin] (axis cs:1.01,0) rectangle (axis cs:1.05,24968);
\draw[draw=none,fill=steelblue,opacity=0.7,very thin] (axis cs:1.11,0) rectangle (axis cs:1.15,30503);
\draw[draw=none,fill=steelblue,opacity=0.7,very thin] (axis cs:1.21,0) rectangle (axis cs:1.25,36116);
\draw[draw=none,fill=steelblue,opacity=0.7,very thin] (axis cs:1.31,0) rectangle (axis cs:1.35,42009);
\draw[draw=none,fill=steelblue,opacity=0.7,very thin] (axis cs:1.41,0) rectangle (axis cs:1.45,47574);
\draw[draw=none,fill=steelblue,opacity=0.7,very thin] (axis cs:1.51,0) rectangle (axis cs:1.55,52151);
\draw[draw=none,fill=steelblue,opacity=0.7,very thin] (axis cs:1.61,0) rectangle (axis cs:1.65,56354);
\draw[draw=none,fill=steelblue,opacity=0.7,very thin] (axis cs:1.71,0) rectangle (axis cs:1.75,59020);
\draw[draw=none,fill=steelblue,opacity=0.7,very thin] (axis cs:1.81,0) rectangle (axis cs:1.85,60249);
\draw[draw=none,fill=steelblue,opacity=0.7,very thin] (axis cs:1.91,0) rectangle (axis cs:1.95,59532);
\draw[draw=none,fill=steelblue,opacity=0.7,very thin] (axis cs:2.01,0) rectangle (axis cs:2.05,58178);
\draw[draw=none,fill=steelblue,opacity=0.7,very thin] (axis cs:2.11,0) rectangle (axis cs:2.15,55832);
\draw[draw=none,fill=steelblue,opacity=0.7,very thin] (axis cs:2.21,0) rectangle (axis cs:2.25,52267);
\draw[draw=none,fill=steelblue,opacity=0.7,very thin] (axis cs:2.31,0) rectangle (axis cs:2.35,48223);
\draw[draw=none,fill=steelblue,opacity=0.7,very thin] (axis cs:2.41,0) rectangle (axis cs:2.45,43373);
\draw[draw=none,fill=steelblue,opacity=0.7,very thin] (axis cs:2.51,0) rectangle (axis cs:2.55,38865);
\draw[draw=none,fill=steelblue,opacity=0.7,very thin] (axis cs:2.61,0) rectangle (axis cs:2.65,33956);
\draw[draw=none,fill=steelblue,opacity=0.7,very thin] (axis cs:2.71,0) rectangle (axis cs:2.75,29577);
\draw[draw=none,fill=steelblue,opacity=0.7,very thin] (axis cs:2.81,0) rectangle (axis cs:2.85,25210);
\draw[draw=none,fill=steelblue,opacity=0.7,very thin] (axis cs:2.91,0) rectangle (axis cs:2.95,21592);
\draw[draw=none,fill=steelblue,opacity=0.7,very thin] (axis cs:3.01,0) rectangle (axis cs:3.05,17926);
\draw[draw=none,fill=steelblue,opacity=0.7,very thin] (axis cs:3.11,0) rectangle (axis cs:3.15,14676);
\draw[draw=none,fill=steelblue,opacity=0.7,very thin] (axis cs:3.21,0) rectangle (axis cs:3.25,12106);
\draw[draw=none,fill=steelblue,opacity=0.7,very thin] (axis cs:3.31,0) rectangle (axis cs:3.35,9448);
\draw[draw=none,fill=steelblue,opacity=0.7,very thin] (axis cs:3.41,0) rectangle (axis cs:3.45,7574);
\draw[draw=none,fill=steelblue,opacity=0.7,very thin] (axis cs:3.51,0) rectangle (axis cs:3.55,5791);
\draw[draw=none,fill=steelblue,opacity=0.7,very thin] (axis cs:3.61,0) rectangle (axis cs:3.65,4513);
\draw[draw=none,fill=steelblue,opacity=0.7,very thin] (axis cs:3.71,0) rectangle (axis cs:3.75,3348);
\draw[draw=none,fill=steelblue,opacity=0.7,very thin] (axis cs:3.81,0) rectangle (axis cs:3.85,2564);
\draw[draw=none,fill=steelblue,opacity=0.7,very thin] (axis cs:3.91,0) rectangle (axis cs:3.95,1967);
\draw[draw=none,fill=steelblue,opacity=0.7,very thin] (axis cs:4.01,0) rectangle (axis cs:4.05,1459);
\draw[draw=none,fill=steelblue,opacity=0.7,very thin] (axis cs:4.11,0) rectangle (axis cs:4.15,1034);
\draw[draw=none,fill=steelblue,opacity=0.7,very thin] (axis cs:4.21,0) rectangle (axis cs:4.25,724);
\draw[draw=none,fill=steelblue,opacity=0.7,very thin] (axis cs:4.31,0) rectangle (axis cs:4.35,492);
\draw[draw=none,fill=steelblue,opacity=0.7,very thin] (axis cs:4.41,0) rectangle (axis cs:4.45,364);
\draw[draw=none,fill=steelblue,opacity=0.7,very thin] (axis cs:4.51,0) rectangle (axis cs:4.55,307);
\draw[draw=none,fill=steelblue,opacity=0.7,very thin] (axis cs:4.61,0) rectangle (axis cs:4.65,190);
\draw[draw=none,fill=steelblue,opacity=0.7,very thin] (axis cs:4.71,0) rectangle (axis cs:4.75,146);
\draw[draw=none,fill=steelblue,opacity=0.7,very thin] (axis cs:4.81,0) rectangle (axis cs:4.85,113);
\draw[draw=none,fill=steelblue,opacity=0.7,very thin] (axis cs:4.91,0) rectangle (axis cs:4.95,64);
\draw[draw=none,fill=coral,opacity=0.7,very thin] (axis cs:0.05,0) rectangle (axis cs:0.09,26);
\draw[draw=none,fill=coral,opacity=0.7,very thin] (axis cs:0.15,0) rectangle (axis cs:0.19,178);
\draw[draw=none,fill=coral,opacity=0.7,very thin] (axis cs:0.25,0) rectangle (axis cs:0.29,503);
\draw[draw=none,fill=coral,opacity=0.7,very thin] (axis cs:0.35,0) rectangle (axis cs:0.39,1184);
\draw[draw=none,fill=coral,opacity=0.7,very thin] (axis cs:0.45,0) rectangle (axis cs:0.49,2255);
\draw[draw=none,fill=coral,opacity=0.7,very thin] (axis cs:0.55,0) rectangle (axis cs:0.59,3626);
\draw[draw=none,fill=coral,opacity=0.7,very thin] (axis cs:0.65,0) rectangle (axis cs:0.69,5089);
\draw[draw=none,fill=coral,opacity=0.7,very thin] (axis cs:0.75,0) rectangle (axis cs:0.79,6684);
\draw[draw=none,fill=coral,opacity=0.7,very thin] (axis cs:0.85,0) rectangle (axis cs:0.89,8375);
\draw[draw=none,fill=coral,opacity=0.7,very thin] (axis cs:0.95,0) rectangle (axis cs:0.99,10030);
\draw[draw=none,fill=coral,opacity=0.7,very thin] (axis cs:1.05,0) rectangle (axis cs:1.09,11761);
\draw[draw=none,fill=coral,opacity=0.7,very thin] (axis cs:1.15,0) rectangle (axis cs:1.19,13678);
\draw[draw=none,fill=coral,opacity=0.7,very thin] (axis cs:1.25,0) rectangle (axis cs:1.29,15956);
\draw[draw=none,fill=coral,opacity=0.7,very thin] (axis cs:1.35,0) rectangle (axis cs:1.39,17982);
\draw[draw=none,fill=coral,opacity=0.7,very thin] (axis cs:1.45,0) rectangle (axis cs:1.49,20379);
\draw[draw=none,fill=coral,opacity=0.7,very thin] (axis cs:1.55,0) rectangle (axis cs:1.59,22918);
\draw[draw=none,fill=coral,opacity=0.7,very thin] (axis cs:1.65,0) rectangle (axis cs:1.69,26025);
\draw[draw=none,fill=coral,opacity=0.7,very thin] (axis cs:1.75,0) rectangle (axis cs:1.79,28773);
\draw[draw=none,fill=coral,opacity=0.7,very thin] (axis cs:1.85,0) rectangle (axis cs:1.89,31586);
\draw[draw=none,fill=coral,opacity=0.7,very thin] (axis cs:1.95,0) rectangle (axis cs:1.99,35162);
\draw[draw=none,fill=coral,opacity=0.7,very thin] (axis cs:2.05,0) rectangle (axis cs:2.09,37548);
\draw[draw=none,fill=coral,opacity=0.7,very thin] (axis cs:2.15,0) rectangle (axis cs:2.19,40751);
\draw[draw=none,fill=coral,opacity=0.7,very thin] (axis cs:2.25,0) rectangle (axis cs:2.29,43323);
\draw[draw=none,fill=coral,opacity=0.7,very thin] (axis cs:2.35,0) rectangle (axis cs:2.39,44771);
\draw[draw=none,fill=coral,opacity=0.7,very thin] (axis cs:2.45,0) rectangle (axis cs:2.49,45948);
\draw[draw=none,fill=coral,opacity=0.7,very thin] (axis cs:2.55,0) rectangle (axis cs:2.59,46714);
\draw[draw=none,fill=coral,opacity=0.7,very thin] (axis cs:2.65,0) rectangle (axis cs:2.69,47543);
\draw[draw=none,fill=coral,opacity=0.7,very thin] (axis cs:2.75,0) rectangle (axis cs:2.79,46384);
\draw[draw=none,fill=coral,opacity=0.7,very thin] (axis cs:2.85,0) rectangle (axis cs:2.89,45683);
\draw[draw=none,fill=coral,opacity=0.7,very thin] (axis cs:2.95,0) rectangle (axis cs:2.99,44393);
\draw[draw=none,fill=coral,opacity=0.7,very thin] (axis cs:3.05,0) rectangle (axis cs:3.09,41999);
\draw[draw=none,fill=coral,opacity=0.7,very thin] (axis cs:3.15,0) rectangle (axis cs:3.19,39801);
\draw[draw=none,fill=coral,opacity=0.7,very thin] (axis cs:3.25,0) rectangle (axis cs:3.29,36932);
\draw[draw=none,fill=coral,opacity=0.7,very thin] (axis cs:3.35,0) rectangle (axis cs:3.39,33945);
\draw[draw=none,fill=coral,opacity=0.7,very thin] (axis cs:3.45,0) rectangle (axis cs:3.49,30685);
\draw[draw=none,fill=coral,opacity=0.7,very thin] (axis cs:3.55,0) rectangle (axis cs:3.59,27185);
\draw[draw=none,fill=coral,opacity=0.7,very thin] (axis cs:3.65,0) rectangle (axis cs:3.69,23613);
\draw[draw=none,fill=coral,opacity=0.7,very thin] (axis cs:3.75,0) rectangle (axis cs:3.79,20243);
\draw[draw=none,fill=coral,opacity=0.7,very thin] (axis cs:3.85,0) rectangle (axis cs:3.89,16440);
\draw[draw=none,fill=coral,opacity=0.7,very thin] (axis cs:3.95,0) rectangle (axis cs:3.99,13224);
\draw[draw=none,fill=coral,opacity=0.7,very thin] (axis cs:4.05,0) rectangle (axis cs:4.09,10457);
\draw[draw=none,fill=coral,opacity=0.7,very thin] (axis cs:4.15,0) rectangle (axis cs:4.19,7524);
\draw[draw=none,fill=coral,opacity=0.7,very thin] (axis cs:4.25,0) rectangle (axis cs:4.29,5516);
\draw[draw=none,fill=coral,opacity=0.7,very thin] (axis cs:4.35,0) rectangle (axis cs:4.39,3972);
\draw[draw=none,fill=coral,opacity=0.7,very thin] (axis cs:4.45,0) rectangle (axis cs:4.49,2710);
\draw[draw=none,fill=coral,opacity=0.7,very thin] (axis cs:4.55,0) rectangle (axis cs:4.59,1767);
\draw[draw=none,fill=coral,opacity=0.7,very thin] (axis cs:4.65,0) rectangle (axis cs:4.69,1087);
\draw[draw=none,fill=coral,opacity=0.7,very thin] (axis cs:4.75,0) rectangle (axis cs:4.79,672);
\draw[draw=none,fill=coral,opacity=0.7,very thin] (axis cs:4.85,0) rectangle (axis cs:4.89,415);
\draw[draw=none,fill=coral,opacity=0.7,very thin] (axis cs:4.95,0) rectangle (axis cs:4.99,242);

\addlegendimage{area legend,
  draw=none,
  fill=steelblue,
  opacity=0.6
}
\addlegendentry{$\eta=0.0$}

\addlegendimage{area legend,
  draw=none,
  fill=coral,
  opacity=0.6
}
\addlegendentry{$\eta=0.9$}
\end{axis}

\end{tikzpicture}
     \label{fig:grads_magnitude}
  \end{subfigure}
  \caption{\textbf{Masking helps capturing essential contextual information.} (a) Distribution of $\ell_1$ norm of gradients computed per image and, (b) distribution of gradient magnitude per pixel over 10k images for CIFAR10 dataset.}
  \vspace{-1em}
  \label{fig:gradients_l1_and_magnitude}
\end{figure}

To assess how different pixel locations contribute to the generation of a target pixel, we compute the average gradient sensitivity over 10k CIFAR10 images. Specifically, we evaluate the $\ell_1$ norm of the gradients, $\lVert S(q, t) \rVert_{1}$, which measures the overall strength of correlations from all other pixel positions to a central pixel in an image. We set $q$ as the central pixel location and $t = 0.789$ as a low noise level. To further examine how individual pixel positions contribute differently, we analyze the distribution of per-pixel gradient magnitudes. 

As shown in \cref{fig:gradients_l1_and_magnitude} (b), the baseline method exhibits a narrow gradient-magnitude distribution, indicating that most pixels exert a similar level of influence. In contrast, our method produces a wider distribution, reflecting more diverse and differentiated contributions across pixel locations. This suggests that masking enables the model to distinguish between pixels and focus on essential correlations, whereas the baseline relies on broadly distributed correlations across most pixels, which can lead to unrealistic artifacts. Although our method captures correlations more selectively, \cref{fig:gradients_l1_and_magnitude} (a) shows that it exhibits substantially stronger correlations (higher $\ell_1$ norm) with specific pixels. Overall, these results demonstrate that our approach effectively captures essential correlations while suppressing unnecessary ones, consistent with prior observations on locality~\cite{lukoianovlocality}.

\subsection{Masking improves spatial consistency (Q5)}
\begin{figure*}[t]
    \centering\footnotesize
    \begin{tikzpicture}[inner sep=0]
      \def\sep{.11\textwidth}
      \node[rotate=90,font=\strut,minimum width=\sep,align=center] at (0,0) {Data};
      \node[rotate=90,font=\strut,minimum width=\sep,align=center] at (0,-\sep) {Baseline \\ $\eta=0.0$};
      \node[rotate=90,font=\strut,minimum width=\sep,align=center] at (0,-2*\sep) {Ours \\ $\eta=0.5$};      
    \end{tikzpicture}%
    \includegraphics[width=0.968\textwidth,trim=35 0 0 0,clip]{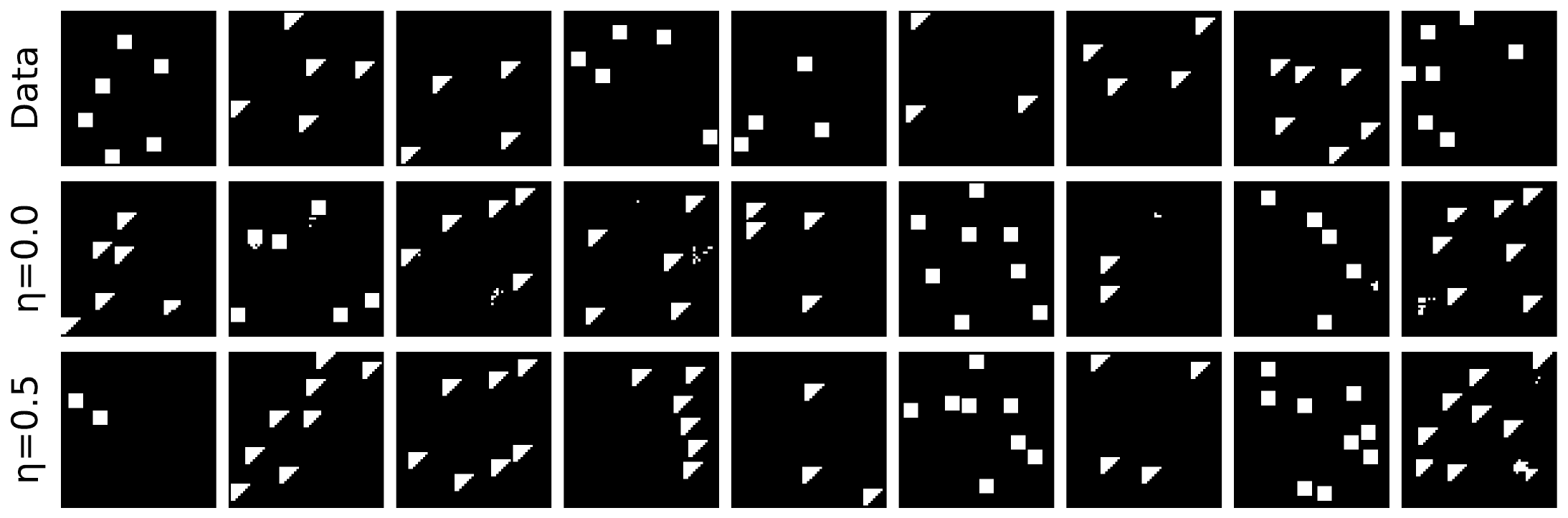}
    \caption{
    \textbf{Demonstration of improved spatial consistency.} After training for 5k epochs on a binary image dataset of triangles 
    (\,\protect\tikz[baseline=-.5ex]\protect\draw[anchor=base,fill=black] (0,-3pt) -- ++(0,6pt) -- ++(6pt,0) -- cycle;\,)
    and squares
    (\,\protect\tikz[baseline=-.5ex]\protect\node[anchor=base,inner sep=3pt,draw,fill=black]{};\,) (\textit{1st row}),
    our model generates images with less scattered dots (\textit{3rd row}),~\ie enhanced structural integrity, compared to the standard diffusion model without masking (\textit{2nd row}). Quantitative results are reported in~\cref{tab:spatial_consistency}.}
    \vspace{-1em}
    \label{fig:comparison_shape}
\end{figure*}

\begin{wraptable}{r}{0.5\textwidth}
    \centering
    \footnotesize
    \vspace{-\baselineskip} %
    \caption{\textbf{Masking improves spatial consistency}. The table shows the number of scattered pixel clusters and number of images containing such clusters over 5k generated images for the illustrative dataset shown in \cref{fig:comparison_shape}.}
    \setlength{\tabcolsep}{4pt}
    \begin{tabular}{lcc}
        \toprule
        Method & No.\ of scatters ($\downarrow$) & No.\ of inconsistent images ($\downarrow$) \\
        \midrule
        Baseline & 11,208 & 3,899 \\
        Ours     & \textbf{1,014} & \textbf{792} \\
        \bottomrule
    \end{tabular}
    \vspace{-1em}
    \label{tab:spatial_consistency}
\end{wraptable}
We construct a toy dataset comprising 500 binary images containing randomly positioned squares and triangles. \cref{fig:comparison_shape} presents generated samples from SSD ($\eta=0.5$) and baseline FM ($\eta=0$), alongside samples from the training dataset. The results show that, with masking, SSD produces fewer unstructured shapes, whereas the baseline FM model occasionally generates scattered dots in some images.

To quantify spatial consistency, we count the scattered pixel clusters, defined as pixel clusters smaller than both the triangle and square sizes, across 5k generated images for each method. \cref{tab:spatial_consistency} reports both the total number of scattered pixel clusters and the number of images containing such clusters. The results show that, by masking, SSD substantially improves spatial consistency, reducing the number of spatially inconsistent images from 3,899 to 792. Furthermore, the total number of scattered clusters decreases from 11,208 to 1,014, corresponding to a 11$\times$ reduction.

\begin{wraptable}{r}{0.49\textwidth}
    \centering\footnotesize
    \vspace{-\baselineskip}
    \setlength{\tabcolsep}{4pt}
    \caption{Comparison of finger-count errors among 1,000 generated 11kHands samples.}
    \label{tab:hands_consistency}
    \begin{tabular}{lcc}
        \toprule
        Error type ($\downarrow$) & Baseline ($\eta=0$) & SSD ($\eta=0.8$) \\
        \midrule
        Six fingers & 26 & \textbf{8} \\
        Four fingers & \textbf{9} & 14 \\
        Total & 35 & \textbf{22} \\
        \bottomrule
    \end{tabular}
\end{wraptable}
\paragraph{Spatial consistency on natural images} We additionally evaluate on 11kHands~\citep{afifi2019hands} at 64$\times$64 resolution. We train the baseline ($\eta=0$) and SSD ($\eta=0.8$) for 400 epochs and generate 1,000 samples per model using the same random seed. Counting four- and six-finger cases, SSD reduces the total from 35 to 22 as shown in \cref{tab:hands_consistency}. We also observe that the baseline model frequently generates hands with six fingers, while SSD mitigates this issue but occasionally produces hands with only four fingers. Qualitative samples are provided in~\cref{fig:hands_samples}. We hypothesize that by tuning the masking ratio, we can further improve the spatial consistency.

\begin{figure*}[t]
\captionsetup[subfigure]{labelformat=empty}
    \centering
    \setlength{\figurewidth}{.165\textwidth}
    \setlength{\figureheight}{\figurewidth}  
    \pgfplotsset{
      axis on top,
      scale only axis,
      xticklabel style={font=\tiny,scale=.7},
      yticklabel style={font=\tiny,scale=.7},
      ylabel near ticks,
      xlabel near ticks,
      tick align=outside,
      tick pos=both,
      xlabel={$x_1$},
      tick style={major tick length=0pt, minor tick length=0pt},
      title style={font=\strut\scriptsize,yshift=-6pt}
    }
    \tiny
    \def\mytext{}
    \begin{subfigure}[t]{.21\textwidth}
      \raggedleft
      \newcommand{\ssdscoreimagea}[2][]{\includegraphics[#1]{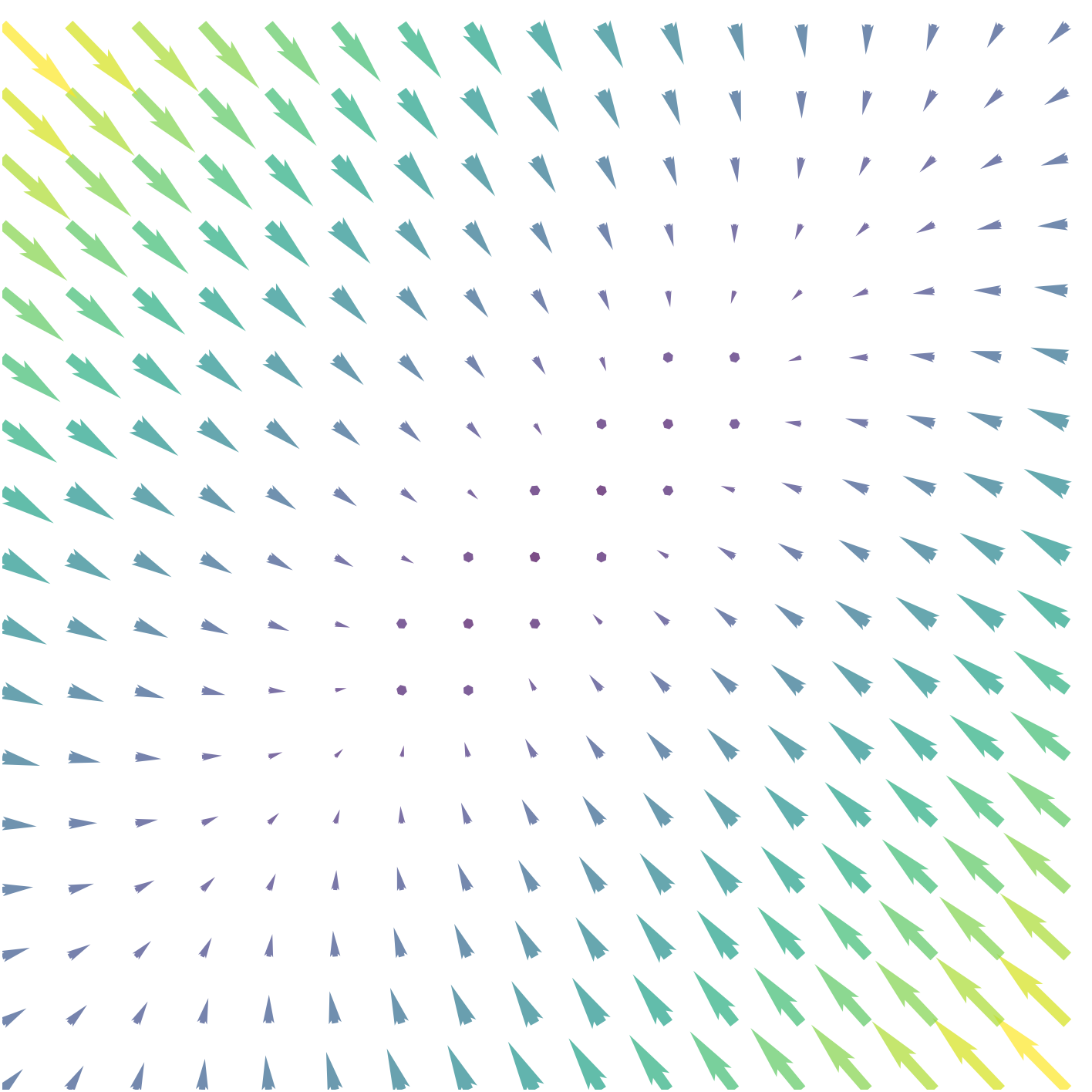}}
      \pgfplotsset{ytick={-2,0,2},ylabel={$x_2$},title={Population score}}
\begin{tikzpicture}

\definecolor{color0}{rgb}{0.12156862745098,0.466666666666667,0.705882352941177}

\begin{axis}[
axis on top,
height=\figureheight,
width=\figurewidth,
xmin=-3, xmax=3,
ymin=-3, ymax=3,
]
\addplot graphics [includegraphics cmd=\ssdscoreimagea,xmin=-3, xmax=3, ymin=-3, ymax=3] {figs/score_figures/population_score.png};
\addplot [draw=color0, fill=color0, mark=x, only marks, opacity=1, scatter]
table[row sep=\\]{
x  y \\
4 4 \\
};

\draw[black,line width=1pt] (axis cs:-2.98,1) rectangle (axis cs:-1,2.98);

\node[font=\scriptsize\strut] at (axis cs: 0,-2) {\mytext};
\end{axis}

\end{tikzpicture}
     \end{subfigure}
    \hfill
    \begin{subfigure}[t]{.19\textwidth}
      \raggedleft
      \newcommand{\ssdscoreimageb}[2][]{\includegraphics[#1]{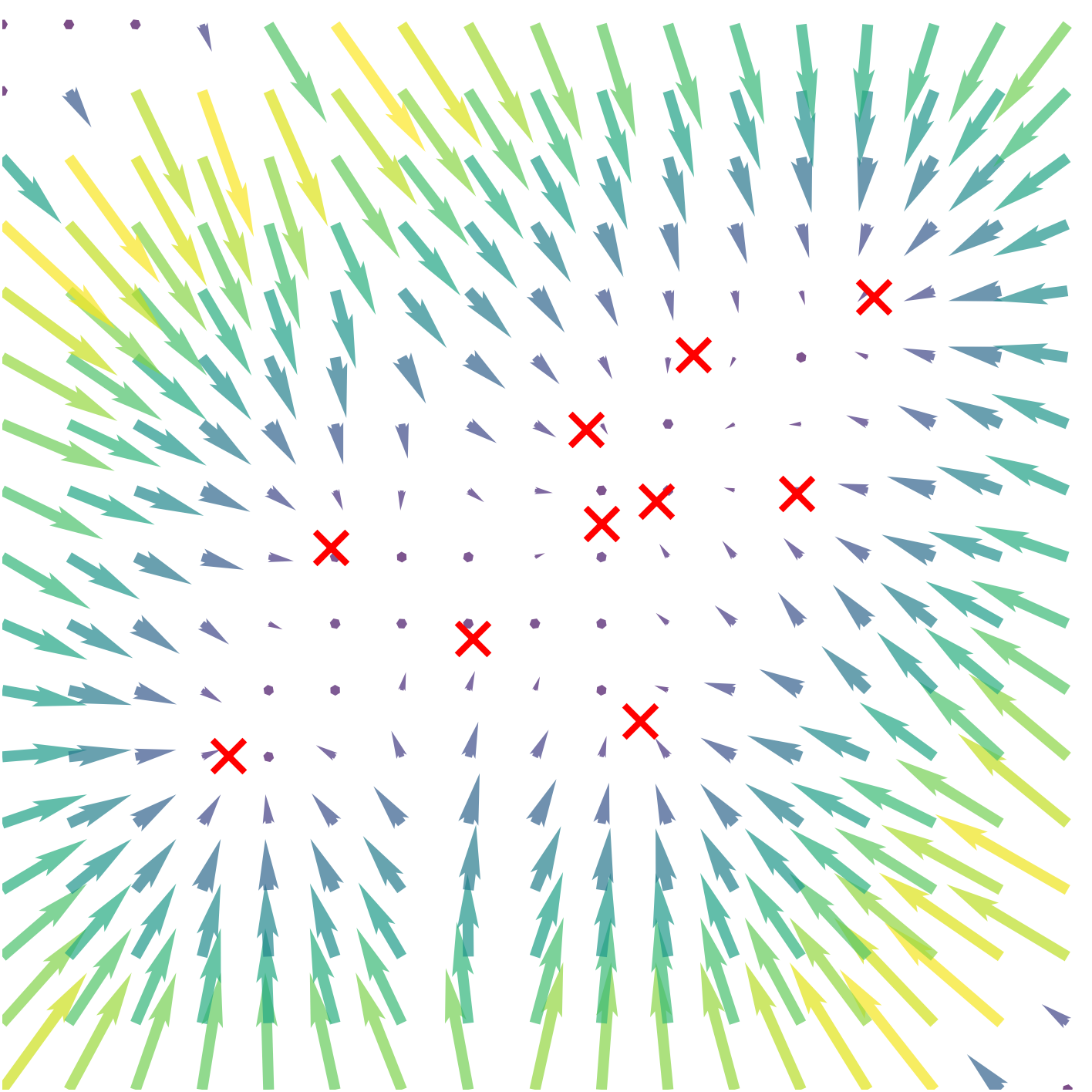}}
      \pgfplotsset{ytick=\empty,ylabel={},title={Emp.\ score ($\eta=0$)}}
\begin{tikzpicture}

\definecolor{color0}{rgb}{0.12156862745098,0.466666666666667,0.705882352941177}

\begin{axis}[
axis on top,
height=\figureheight,
width=\figurewidth,
xmin=-3, xmax=3,
ymin=-3, ymax=3,
]
\addplot graphics [includegraphics cmd=\ssdscoreimageb,xmin=-3, xmax=3, ymin=-3, ymax=3] {figs/score_figures/empirical_score_eta_0.png};
\addplot [draw=color0, fill=color0, mark=x, only marks, opacity=1, scatter]
table[row sep=\\]{
x  y \\
4 4 \\
};

\draw[black,line width=1pt] (axis cs:-2.98,1) rectangle (axis cs:-1,2.98);

\node[font=\scriptsize\strut] at (axis cs: 0,-2) {\mytext};
\end{axis}

\end{tikzpicture}
     \end{subfigure}
    \hfill
    \begin{subfigure}[t]{.19\textwidth}
      \raggedleft
      \newcommand{\ssdscoreimagec}[2][]{\includegraphics[#1]{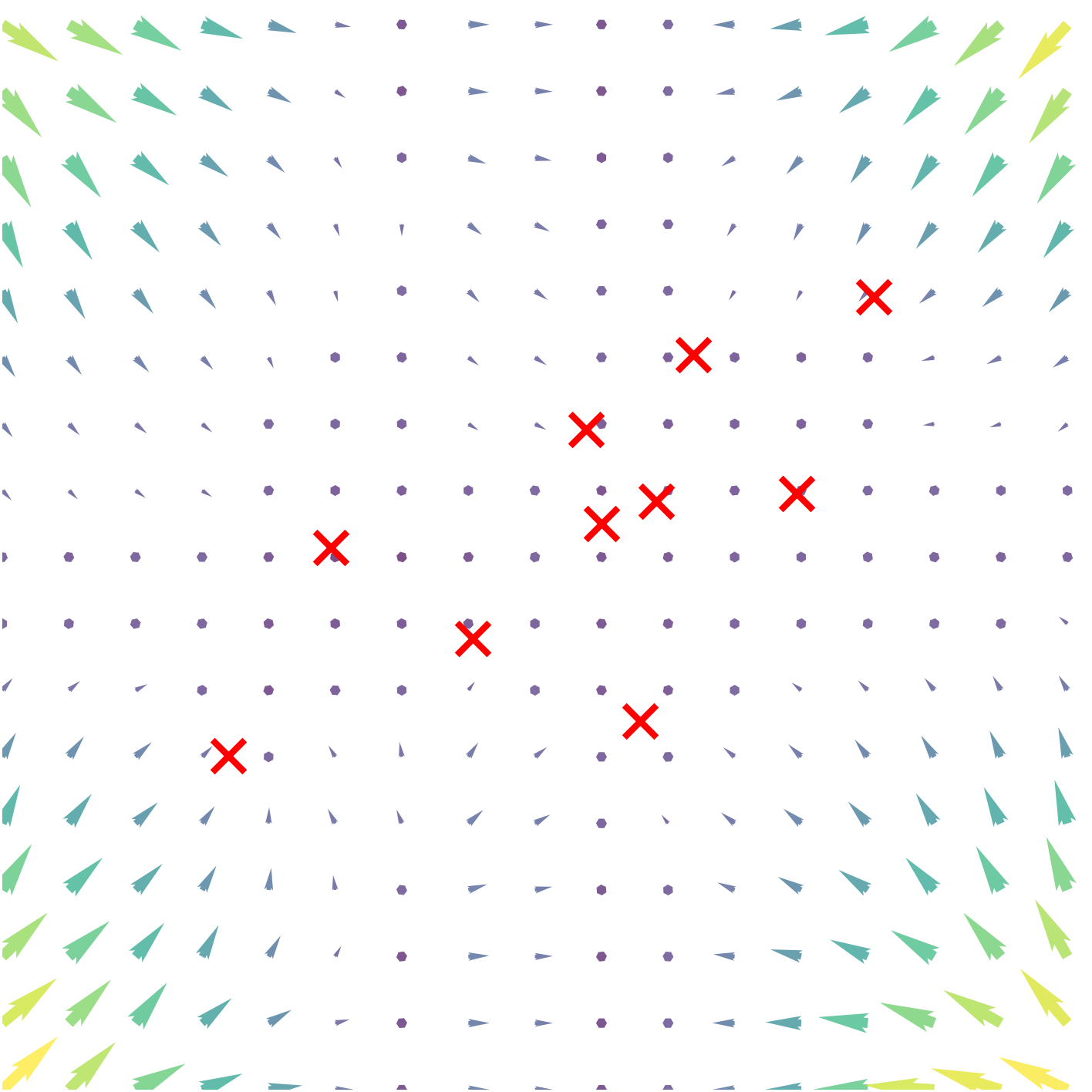}}
      \pgfplotsset{ytick=\empty,ylabel={},title={Emp.\ score ($\eta=0.5$)}}
\begin{tikzpicture}

\definecolor{color0}{rgb}{0.12156862745098,0.466666666666667,0.705882352941177}

\begin{axis}[
axis on top,
height=\figureheight,
width=\figurewidth,
xmin=-3, xmax=3,
ymin=-3, ymax=3,
]
\addplot graphics [includegraphics cmd=\ssdscoreimagec,xmin=-3, xmax=3, ymin=-3, ymax=3] {figs/score_figures/empirical_score_eta_0.5.png};
\addplot [draw=color0, fill=color0, mark=x, only marks, opacity=1, scatter]
table[row sep=\\]{
x  y \\
4 4 \\
};

\draw[black,line width=1pt] (axis cs:-2.98,1) rectangle (axis cs:-1,2.98);

\node[font=\scriptsize\strut] at (axis cs: 0,-2) {\mytext};
\end{axis}

\end{tikzpicture}
     \end{subfigure}
    \hfill
    \begin{subfigure}[t]{.19\textwidth}
      \raggedleft
      \newcommand{\ssdscoreimaged}[2][]{\includegraphics[#1]{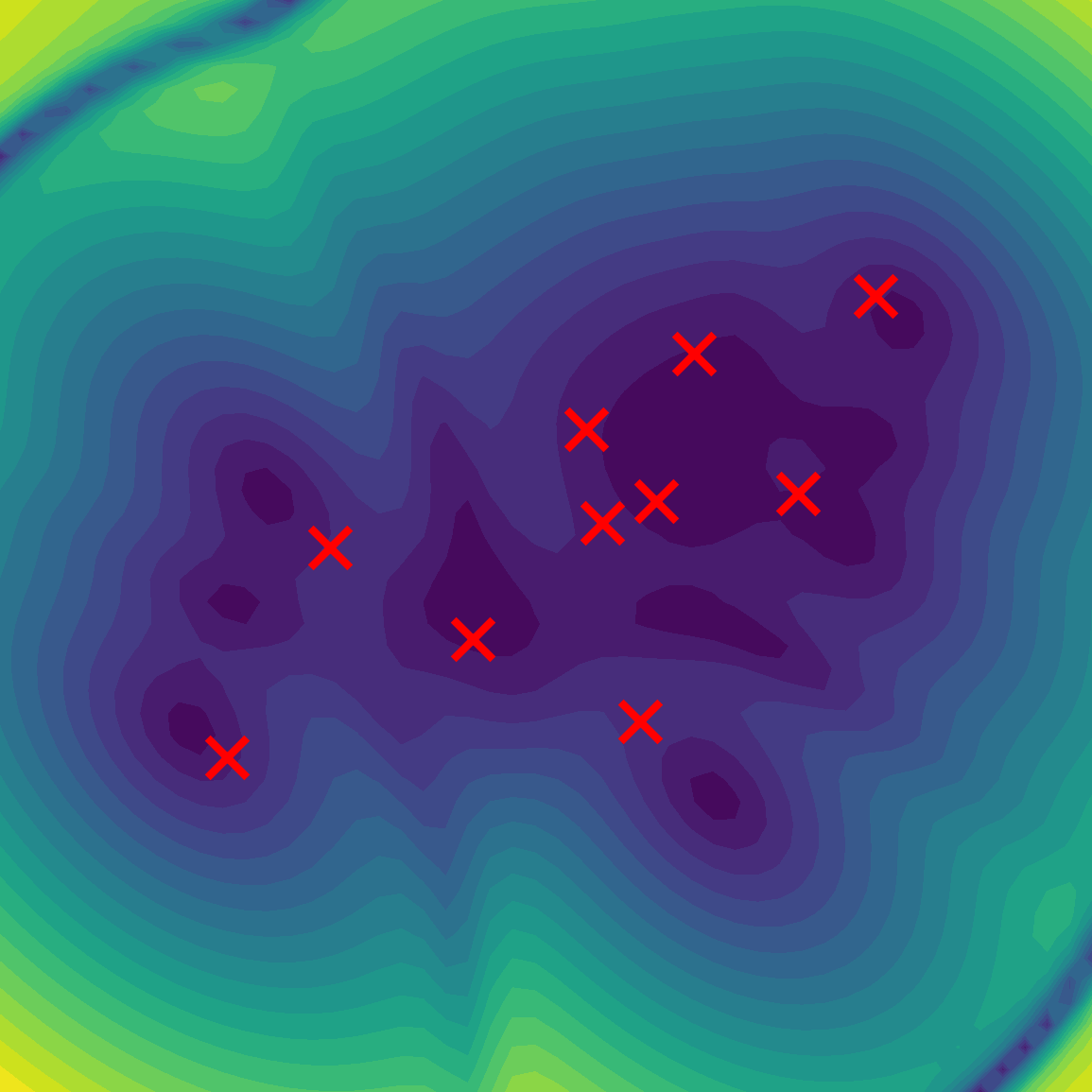}}
      \pgfplotsset{ytick=\empty,ylabel={},title={$\eta=0$ (Avg. error = 3.65)}}
\begin{tikzpicture}

\definecolor{color0}{rgb}{0.12156862745098,0.466666666666667,0.705882352941177}

\begin{axis}[
axis on top,
height=\figureheight,
width=\figurewidth,
xmin=-3, xmax=3,
ymin=-3, ymax=3,
]
\addplot graphics [includegraphics cmd=\ssdscoreimaged,xmin=-3, xmax=3, ymin=-3, ymax=3] {figs/score_figures/error_eta_0.png};
\addplot [draw=color0, fill=color0, mark=x, only marks, opacity=1, scatter]
table[row sep=\\]{
x  y \\
4 4 \\
};

\node[font=\scriptsize\strut] at (axis cs: 0,-2) {\mytext};
\end{axis}

\end{tikzpicture}
     \end{subfigure}
    \hfill
    \begin{subfigure}[t]{.19\textwidth}
      \raggedleft
      \newcommand{\ssdscoreimagee}[2][]{\includegraphics[#1]{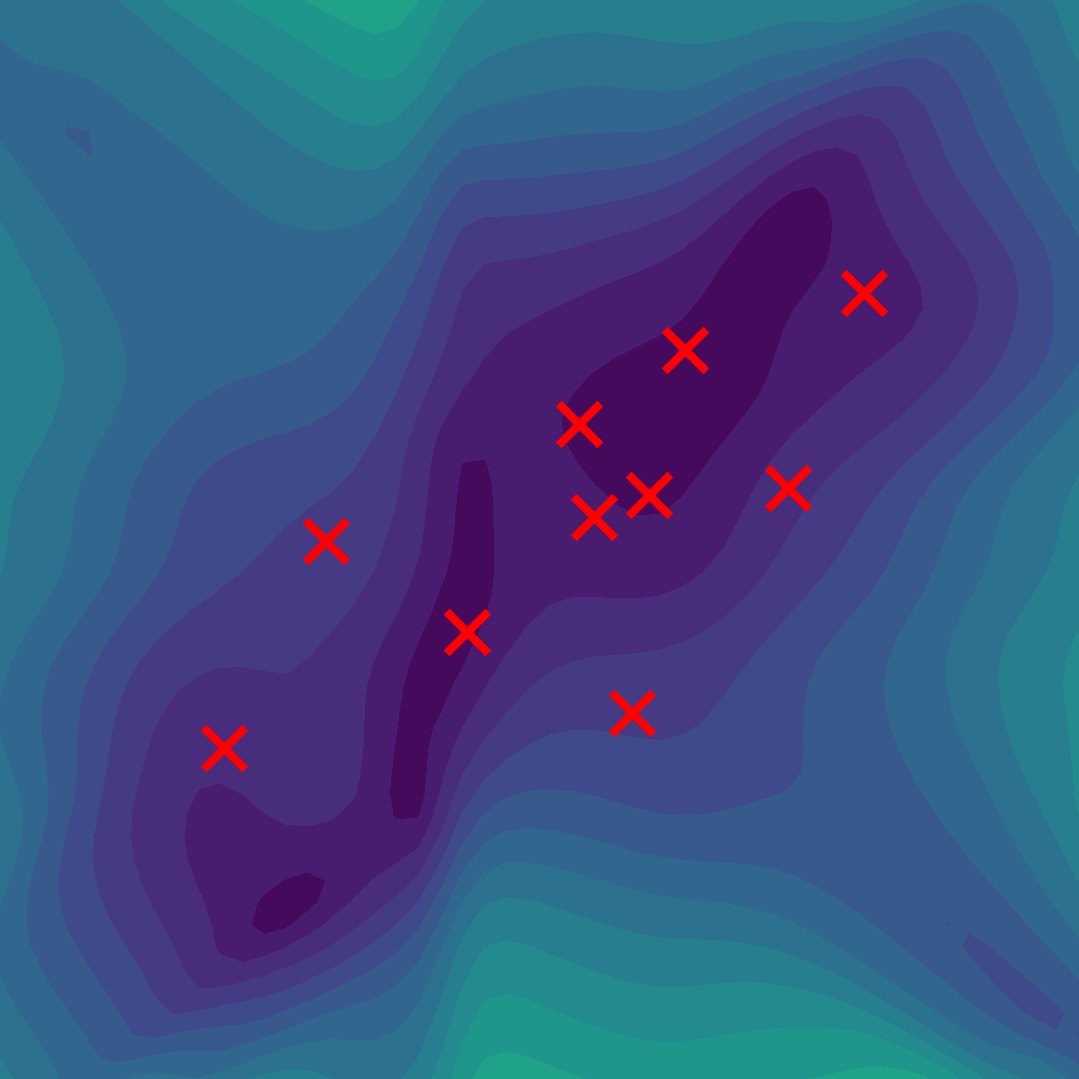}}
      \pgfplotsset{ytick=\empty,ylabel={},title={$\eta=0.5$ (Avg. error = 2.78)}}
\begin{tikzpicture}

\definecolor{color0}{rgb}{0.12156862745098,0.466666666666667,0.705882352941177}

\begin{axis}[
axis on top,
height=\figureheight,
width=\figurewidth,
xmin=-3, xmax=3,
ymin=-3, ymax=3,
]
\addplot graphics [includegraphics cmd=\ssdscoreimagee,xmin=-3, xmax=3, ymin=-3, ymax=3] {figs/score_figures/error_eta_0.5.png};
\addplot [draw=color0, fill=color0, mark=x, only marks, opacity=1, scatter]
table[row sep=\\]{
x  y \\
4 4 \\
};

\node[font=\scriptsize\strut] at (axis cs: 0,-2) {\mytext};
\end{axis}

\end{tikzpicture}
     \end{subfigure}
    \def\colorbar{\protect\includegraphics[width=2.5em,height=.7em]{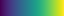}}
    \def\mycross{\protect\tikz[baseline=-0.6ex]\protect\draw[red,line width=0.9pt] (-0.08,0.08) -- (0.08,-0.08) (-0.08,-0.08) -- (0.08,0.08);}
    \vspace*{-4pt}
    \caption{\textbf{Masking improves estimating the population score function}. Visualization of population scores (ground truth) and empirical scores estimated with $\eta=0$ (baseline) and $\eta=0.5$ (ours), along with a comparison of score estimation errors (right two figures) such that the error is color coded by 0~\colorbar~10.63. The baseline model only accurately estimates scores near observed data points (shown as~\mycross), whereas the proposed masking enables more accurate estimation over broader regions (squared areas) and yields lower overall errors.}
    \vspace{-1em}
    \label{fig:score_combined}

\end{figure*}

\subsection{Masking improves score function estimation (Q6)}
\label{sec:score}

\begin{wraptable}{r}{0.45\textwidth}
    \centering
    \footnotesize
    \vspace{-\baselineskip}
    \caption{Comparison of empirical score estimation errors on a 2D Gaussian data distribution. Lower ($\downarrow$) is better.}
    \begin{tabular}{lcc}
        \toprule
        Metric & Baseline & W/ Masking (ours) \\
        \midrule
        Mean error ($\downarrow$) & 3.65  & \textbf{2.78} \\
        Max  error ($\downarrow$) & 10.63 & \textbf{6.18} \\
        \bottomrule
    \end{tabular}
    \label{tab:error_comparison}
    \vspace{-1em}
\end{wraptable}
FID is limited in its ability to assess how closely generated samples match the true data distribution. This limitation helps to explain why diffusion models with excellent FID scores can still produce unrealistic images, even when trained exclusively on natural images~\citep{bonnaire2025diffusion}. Fundamentally, this issue arises because the training objective estimates only the empirical score function rather than the true underlying score function, thereby leading to both memorization and unrealistic generations.

Since we generally do not know the ground truth score function of real datasets~\citep{ho2020denoising}, we design experiments on synthesized 2D Gaussian data, where we can analytically compute the score function. We compare the approximated empirical score functions obtained by the baseline and our method under different masking ratios. \cref{fig:score_combined} visualizes the estimated scores on a grid over the square, based on the given ten data points. By applying masking ($\eta = 0.5$), which generates multiple partial views of the data points, the estimation becomes more accurate over a broader region. The two figures on the right in~\cref{fig:score_combined} presents the estimation errors, defined as the absolute difference from the population score, with and without masking. As reported in~\cref{tab:error_comparison}, the masked estimation achieves an average error of 2.78, which is lower than the baseline’s average error of 3.65. The improved accuracy arises from the mechanism whereby masking introduces an independence prior across data dimensions, thereby reducing the model’s overreliance on misleading correlations induced by limited training samples. The detailed description of the numerical experiment is in~\cref{sec:score_supplement}. 

\section{Discussion and conclusion}

Although diffusion models can generate high-quality images, spatial inconsistency remains a significant challenge. We hypothesize that such unrealistic artifacts arise because excessive correlations present in training data are indiscriminately encoded by diffusion models. Motivated by this observation, we propose \emph{sparsely supervised diffusion} that encourages the model to capture critical, but selective, contextual information when predicting unseen pixel locations. The analytical results show that masking fundamentally alters diffusion model training by modifying the spectrum of the data covariance and suppressing correlations in the data covariance matrix. Extensive experiments show that the proposed SSD offers a \emph{free lunch} for diffusion model training, with several advantages: {\em (1)}~it achieves comparable FID scores across datasets while avoiding long-training instability; {\em (2)}~it mitigates memorization; {\em (3)}~it promotes essential contextual awareness during sample generation; and {\em (4)}~it reduces spatially inconsistent artifacts and yields a more accurate approximation of the population score function. We leave the investigation of the optimal masking ratio for future work.

\renewcommand{\acksection}{\section*{Acknowledgments}}
\begin{ack}
We acknowledge support through Research Council of Finland grants 352788, 362407, 373999, 373780, 362408, and 339730. This research also received support through the Council's Flagship programme, the Finnish Center for Artificial Intelligence (FCAI). Computing resources were supplied by the Aalto Science-IT project. We acknowledge CSC -- IT Center for Science, Finland, for awarding this project access to the LUMI supercomputer, owned by the EuroHPC Joint Undertaking, hosted by CSC (Finland) and the LUMI consortium through CSC. We thank the anonymous reviewers for their constructive feedback and helpful suggestions.
This work was partially supported by the Wallenberg AI, Autonomous Systems and
Software Program (WASP) funded by the Knut and Alice Wallenberg Foundation.
\end{ack}

{\small %

}

\newpage
\appendix

\section{Random masking changes spectrum of covariance}\label{sec:app_spectrum}
\subsection{Setup}

Consider data $\vx  \sim \mathcal{N}(\mathbf{0}, \boldsymbol{\Sigma})$ with covariance matrix $\boldsymbol{\Sigma}$. We apply random binary mask $\vm \in \{0,1\}^d$ where each $\vm_j \sim \text{Bernoulli}(1-\eta)$ independently, with $\eta$ being the probability of being masked (\ie, $\mathbb{P}(\vm_j = 0) = \eta$ and $\mathbb{P}(\vm_j = 1) = 1-\eta$). The masked data is:
\begin{equation}
\tilde{\vx } = \vm \odot \vx ,
\end{equation}
where $\odot$ denotes element-wise multiplication.

\subsection{Covariance of masked data}

The covariance matrix is:
\[
\tilde{\bm\Sigma} = \mathbb{E}[\tilde{\vx }\tilde{\vx }^\top] - \mathbb{E}[\tilde{\vx }]\mathbb{E}[\tilde{\vx }]^\top = \mathbb{E}[\tilde{\vx }\tilde{\vx }^\top] .
\]
The $(j,k)$-th element is:
\[
[\tilde{\bm\Sigma}]_{jk} = \mathbb{E}[\tilde{\vx}_j \tilde{\vx}_k] = \mathbb{E}[\vm_j \vm_k \vx_j \vx_k] .
\]

Since $\vm$ and $\vx $ are independent:
\[
[\tilde{\bm\Sigma}]_{jk} = \mathbb{E}[\vm_j \vm_k] \cdot \mathbb{E}[\vx_j \vx_k] = \mathbb{E}[\vm_j \vm_k] \cdot \boldsymbol{\Sigma}_{jk} .
\]

\textbf{For diagonal elements ($j = k$)}
\[
\mathbb{E}[\vm_j^2] = \mathbb{E}[\vm_j] = 1-\eta .
\]

Therefore:
\begin{equation}
[\tilde{\bm\Sigma}]_{jj} = (1-\eta) \boldsymbol{\Sigma}_{jj}.
\end{equation}

\textbf{For off-diagonal elements ($j \neq k$)}

Since $\vm_j$ and $\vm_k$ are independent:
\[
\mathbb{E}[\vm_j \vm_k] = \mathbb{E}[\vm_j]\mathbb{E}[\vm_k] = (1-\eta)^2 .
\]

Therefore:
\begin{equation*}
[\tilde{\bm\Sigma}]_{jk} = (1-\eta)^2 \boldsymbol{\Sigma}_{jk}.
\end{equation*}

Combining both cases:
\[
\tilde{\bm\Sigma} = (1-\eta)\,\text{diag}(\boldsymbol{\Sigma}) + (1-\eta)^2(\boldsymbol{\Sigma} - \text{diag}(\boldsymbol{\Sigma})).
\]
where $\text{diag}(\boldsymbol{\Sigma})$ means the diagonal matrix formed from the diagonal entries of $\bm\Sigma$.

Or equivalently:
\begin{equation}
\boxed{\tilde{\bm\Sigma} = (1-\eta)^2\boldsymbol{\Sigma} + \eta(1-\eta)\,\text{diag}(\boldsymbol{\Sigma})}
\end{equation}

\subsection{Masking affects spectrum of covariance}
Let eigendecomposition of $\bm\Sigma$ be
\[
\boldsymbol{\Sigma} = \mathbf U \boldsymbol{\Lambda} \mathbf U^\top,
\qquad
\boldsymbol{\Lambda} = \mathrm{diag}(\lambda_1,\ldots,\lambda_d).
\]

Project $\tilde{\bm\Sigma}$ onto one eigenvector $\vu_i$ of $\boldsymbol{\Sigma}$,
\begin{equation}
\tilde{\lambda}_i
:= \vu_i^\top \tilde{\boldsymbol{\Sigma}} \vu_i
= (1-\eta)^2 \lambda_i
+ \eta(1-\eta)\sum_{k} \vu_{ik}^2 \boldsymbol{\Sigma}_{kk}.
\end{equation}

Therefore, we can compute the eigenvalue shrinkage ratio:
\begin{equation}
\frac{\tilde{\lambda}_i}{\lambda_i}
= (1-\eta)^2
+ \eta(1-\eta)\frac{\sum_k \vu_{ik}^2 \boldsymbol{\Sigma}_{kk}}{\lambda_i}.
\end{equation}

If $\boldsymbol{\Sigma}$ is \emph{diagonal}, then $\vu_i$ is a standard basis vector, so
\[
\sum_{k} \vu_{ik}^2 \boldsymbol{\Sigma}_{kk} = \lambda_i,
\]
for the $i$-th component, giving
\[
\frac{\tilde{\lambda}_i}{\lambda_i} = 1-\eta,
\]
which means that all eigenvalues are equally scaled down.

If $\boldsymbol{\Sigma}$ is \emph{non-diagonal}, the term
\[
\sum_{k} \vu_{ik}^2 \boldsymbol{\Sigma}_{kk}=u_i^\top \mathbf D \vu_i,
\]
where $\mathbf{D}$ is $\text{diag}(\bm\Sigma)$, is \textbf{not equal} to $\lambda_i$ in general. This term mixes the diagonal entries of $\boldsymbol{\Sigma}$ weighted by the components of the eigenvector $u_i$.

We can thus simplify the ratio as
\begin{equation}
\boxed{\frac{\tilde{\lambda}_i}{\lambda_i} \propto \frac{u_i^\top \mathbf D u_i}{\lambda_i}.}
\end{equation}
This expression indicates that the masking operation promotes the direction $u_i$ when $\frac{u_i^\top \mathbf D u_i}{\lambda_i}$ is large, and attenuates it when the ratio is small. Intuitively, a large value of $\frac{u_i^\top \mathbf D u_i}{\lambda_i}$ implies that the coordinate-wise variations along direction $u_i$ are strong (\ie, $u_i^\top \mathbf D u_i$ is large), while the overall data correlations captured by the full covariance matrix lead to a relatively small variance $\lambda_i$ along this direction.

\section{Description of 2D Gaussian experiment}
\label{sec:score_supplement}
This section describes the detailed setting of the score estimation experiment in~\cref{sec:score}, including the 2D Gaussian distribution we use, how we compute the population score, and how we compute the empirical scores with and without masking. 

\paragraph{2D Gaussian distribution} 
We consider a 2D Gaussian distribution as our data distribution:
\begin{equation}
p_0(x) = \mathcal{N}(x; 0, \boldsymbol{\Sigma}),
\end{equation}
where $\vx = (x_1, x_2)^T \in \mathbb{R}^2$ and the covariance matrix is
\begin{equation}
\boldsymbol{\Sigma} = \begin{bmatrix} 1 & \rho \\ \rho & 1 \end{bmatrix}.
\end{equation}
Here, $\rho \in [-1, 1]$ is the correlation coefficient between the two dimensions: $\rho = 0$ corresponds to independent dimensions, $\rho > 0$ to positive correlation, and $\rho < 0$ to negative correlation. In our experiment, we set $\rho=0.7$ to account for the fact that in many high-dimensional datasets, such as images, the dimensions exhibit strong correlations. The probability density function is
\begin{equation}
p_0(\vx) = \frac{1}{2\pi\sqrt{1-\rho^2}} \exp\Bigg(-\frac{1}{2(1-\rho^2)} \big(x_1^2 - 2\rho x_1 x_2 + x_2^2\big)\Bigg).
\end{equation}

\paragraph{Forward diffusion process} 
Given data $\vx_0 \sim p_0(\vx)$, the forward noising process is defined as
\begin{equation}
\vx_t = e^{-t} \vx_0 + \sqrt{1 - e^{-2t}} \, \epsilon, \quad \epsilon \sim \mathcal{N}(0, \mI).
\end{equation}
We set $t=0.1$ to reflect low-level noises and $\delta_t = 1 - e^{-2t}$ denotes the noise variance at time $t$.

\paragraph{Population score function} 
For a Gaussian distribution with covariance $\boldsymbol{\Sigma}$, the score function at time $t$ is
\begin{equation}
\nabla_x \log p_t(\vx) = -\boldsymbol{\Sigma}_t^{-1} x,
\end{equation}
where the time-dependent covariance is
\begin{equation}
\boldsymbol{\Sigma}_t = e^{-2t} \boldsymbol{\Sigma} + \delta_t \mI.
\end{equation}

\paragraph{Empirical score estimation} 
Given training data $\{\vx_0^{(i)}\}_{i=1}^n$, the empirical score at a query point $\vx$ is estimated using kernel density estimation:
\begin{equation}
\nabla_\vx \log \hat{p}_t(\vx) = \sum_{i=1}^n w_i(\vx) \cdot \frac{\vx_t^{(i)} - \vx}{\delta_t},
\end{equation}
with
\begin{equation*}
\vx_t^{(i)} = e^{-t} \vx_0^{(i)}, \quad 
d_i(\vx) = \frac{\|\vx_t^{(i)} - \vx\|^2}{\delta_t}, \quad 
w_i(\vx) = \frac{\exp(-\frac{1}{2} d_i(\vx))}{\sum_{j=1}^n \exp(-\frac{1}{2} d_j(\vx))}.
\end{equation*}

\paragraph{Masked score estimation} 
When a mask $\vm^{(i)} \in \{0,1\}^d$ indicates observed dimensions for each sample, the masked score is
\begin{equation}
\nabla_\vx \log \hat{p}_t^{\mathrm{mask}}(\vx) = \sum_{i=1}^n w_i^{\mathrm{mask}}(\vx) \cdot \frac{\vm^{(i)} \odot (\vx_t^{(i)} - \vx)}{\delta_t},
\end{equation}
where
\begin{equation}
    w_i^{\mathrm{mask}}(\vx) = \frac{\exp\Big(-\frac{1}{2} d_i^{\mathrm{mask}}(\vx)\Big)}{\sum_{j=1}^n \exp\Big(-\frac{1}{2} d_j^{\mathrm{mask}}(\vx)\Big)},
\end{equation}
and
\begin{equation*}
d_i^{\mathrm{mask}}(\vx) = \frac{\| \vm^{(i)} \odot (\vx_t^{(i)} - \vx) \|^2}{\delta_t}.
\end{equation*}
The final masked score is obtained by averaging over multiple random masks, where each dimension is observed independently with probability $1-\eta$.
\clearpage

\section{Implementation details}\label{sec:supp_implementation}
We report the implementation details for experiments on CelebA-50K, CIFAR10, and LSUN Bedroom datasets in~\cref{tab:implementation_details}. For ImageNet, we use DiT-XL/2 at 256$\times$256 resolution with the Stable Diffusion VAE (ft-MSE)~\citep{peebles2023scalable}. Baseline and SSD use the same training budget: 180K steps with global batch size 4,096 on full ImageNet, and 200K steps with global batch size 2,048 on the 100K-image subset. Sampling uses 250 DDPM steps with CFG=1.5. We compute FID on 50K generated images against the ImageNet validation set using the ADM evaluation suite. For the 11kHands experiment, we evaluate the 400-epoch checkpoints at 64$\times$64 resolution, sampling 1,000 images per model with seed 0, EMA weights, and Heun's second-order solver with NFE=50.
\begin{table}[ht]
  \centering
  \caption{Key hyperparameters for experiments on three datasets}
  \label{tab:hyperparams}
  \begin{tabular}{llll}
  \toprule
  \textbf{Parameter} & \textbf{CelebA-50K} & \textbf{CIFAR10} & \textbf{LSUN Bedroom}\\
  \midrule
  \multicolumn{2}{l}{\textit{Training Configuration}} \\
  Epochs & 2000 & 3000 & 2000\\
  Effective Batch Size & 128 & 256 & 1024\\
  Learning Rate & 0.0001 & 0.0001 & 0.0001\\
  Optimizer Betas & [0.9, 0.95] & [0.9, 0.95] & [0.9, 0.95]\\
  Use EMA & True & True & True\\
  Resolution & 64\texttimes64 & 32\texttimes32 & 32\texttimes32\\
  \midrule
  \multicolumn{2}{l}{\textit{Flow Matching Configuration}} \\
  Skewed Timesteps & True & True & True\\
  EDM Schedule & True & True & True\\
  \midrule
  \multicolumn{2}{l}{\textit{Sampling Configuration}} \\
  ODE Method & Heun2 & Heun2 & Heun2\\
  Number of Function Evaluations & 50 & 50 & 50\\
  ODE Tolerance (atol/rtol) & -- & -- & --\\
  \midrule
  \multicolumn{2}{l}{\textit{Dataset \& Evaluation}} \\
  Number of Images & 50,000 & 50,000 & 303,125\\
  Number of Classes & -- & 10 & --\\
  FID Samples & 50,000 & 50,000 & 50,000\\
  Evaluation Frequency & 100 & 100 & 200\\
  CFG Scale & -- & -- & --\\
  Class Drop Prob & -- & -- & --\\
  \midrule
  \multicolumn{2}{l}{\textit{Model Architecture}} \\
  Input/Output Channels & 3 & 3 & 3\\
  Model Channels & 128 & 128 & 128\\
  Number of ResBlocks & 4 & 4 & 4\\
  Attention Resolutions & [2, 4] & [2] & [2]\\
  Dropout & 0.2 & 0.3 & 0.3\\
  Channel Multipliers & [1, 2, 2, 4] & [2, 2, 2] & [2, 2, 2]\\
  Convolution Resample & True & False & False\\
  Number of Heads & 2 & 1 & 1\\
  Head Channels & 64 & -1 & -1\\
  Scale Shift Norm & True & True & True\\
  ResBlock Up/Down & True & False & False\\
  New Attention Order & True & True & True\\
  \midrule
  \multicolumn{2}{l}{\textit{System Configuration}} \\
  Number of GPUs & 8 & 8 & 8\\
  \bottomrule
  \end{tabular}
  \label{tab:implementation_details}
\end{table}

\clearpage

\section{More experiments}
\subsection{Comparison with other regularization methods}
We further compare our method against the commonly used regularization technique of dropout and the MDT~\citep{gao2023masked} baseline in~\cref{tab:mdt_dropout_comparison}. 
\label{sec:more_results}
\begin{table}[ht]
    \centering
    \footnotesize
    \caption{Comparison with dropout and MDT baselines. Both dropout and MDT fail to prevent divergence during training. All methods on CelebA-10K are trained for 2000 epochs.}
    \label{tab:mdt_dropout_comparison}
    \begin{tabular}{lcc}
        \toprule
        Method & CelebA-10K FID ($\downarrow$) & CIFAR10 FID ($\downarrow$) \\
        \midrule
        MDT                      & 24.72          & -- \\
        Dropout ($p=0.30$) & 13.72          & 2.28 \\
        Dropout ($p=0.98$)       & 13.38          & 9.37 \\
        SSD (ours, $\eta=0.98$)  & \textbf{5.10}  & \textbf{2.20} \\
        \bottomrule
    \end{tabular}
\end{table}

\subsection{ImageNet under data constraints}
On ImageNet-100K, SSD achieves a lower FID than the baseline after 200K training steps, whereas the improvement is less pronounced on the full ImageNet dataset. This is expected, as prolonged training on smaller datasets can lead to overfitting and memorization, degrading generation quality.

\begin{table}[ht]
    \centering\footnotesize
    \caption{ImageNet FID at CFG=1.5 using DiT-XL/2 at 256$\times$256.}
    \label{tab:imagenet_data_budget}
    \begin{tabular}{llccc}
        \toprule
        Training data & Training steps & Baseline & SSD ($\eta=0.8$) & SSD ($\eta=0.98$) \\
        \midrule
        ImageNet-100K & 200K & 25.99 & 18.45 & 15.58 \\
        Full ImageNet & 180K & 4.98 & -- & 4.83 \\
        \bottomrule
    \end{tabular}
\end{table}

\subsection{Patch-wise masking}
We compare element-wise loss masking with 8$\times$8 patch-wise loss masking on CelebA-10K at 64$\times$64 resolution in \cref{tab:patch_masking}. The results show that the proposed sparse supervision also works well with different masking strategies.

\begin{table}[h]
    \centering\footnotesize
    \caption{Masking-granularity ablation on CelebA-10K. Results are reported at 3,500 epochs.}
    \label{tab:patch_masking}
    \begin{tabular}{lcc}
        \toprule
        Masking strategy & Mask ratio $\eta$ & Final FID ($\downarrow$) \\
        \midrule
        None (baseline) & 0 & 13.72 \\
        Element-wise & 0.5 & 13.29 \\
        Element-wise & 0.95 & 6.36 \\
        Element-wise & 0.98 & 5.10 \\
        Patch-wise (8$\times$8) & 0.6 & 7.61 \\
        Patch-wise (8$\times$8) & 0.9 & 5.14 \\
        \bottomrule
    \end{tabular}
\end{table}

\clearpage
\section{Generated hand samples}
\begin{figure}[ht]
    \centering
    \begin{subfigure}{0.98\linewidth}
        \centering
        \includegraphics[width=\linewidth]{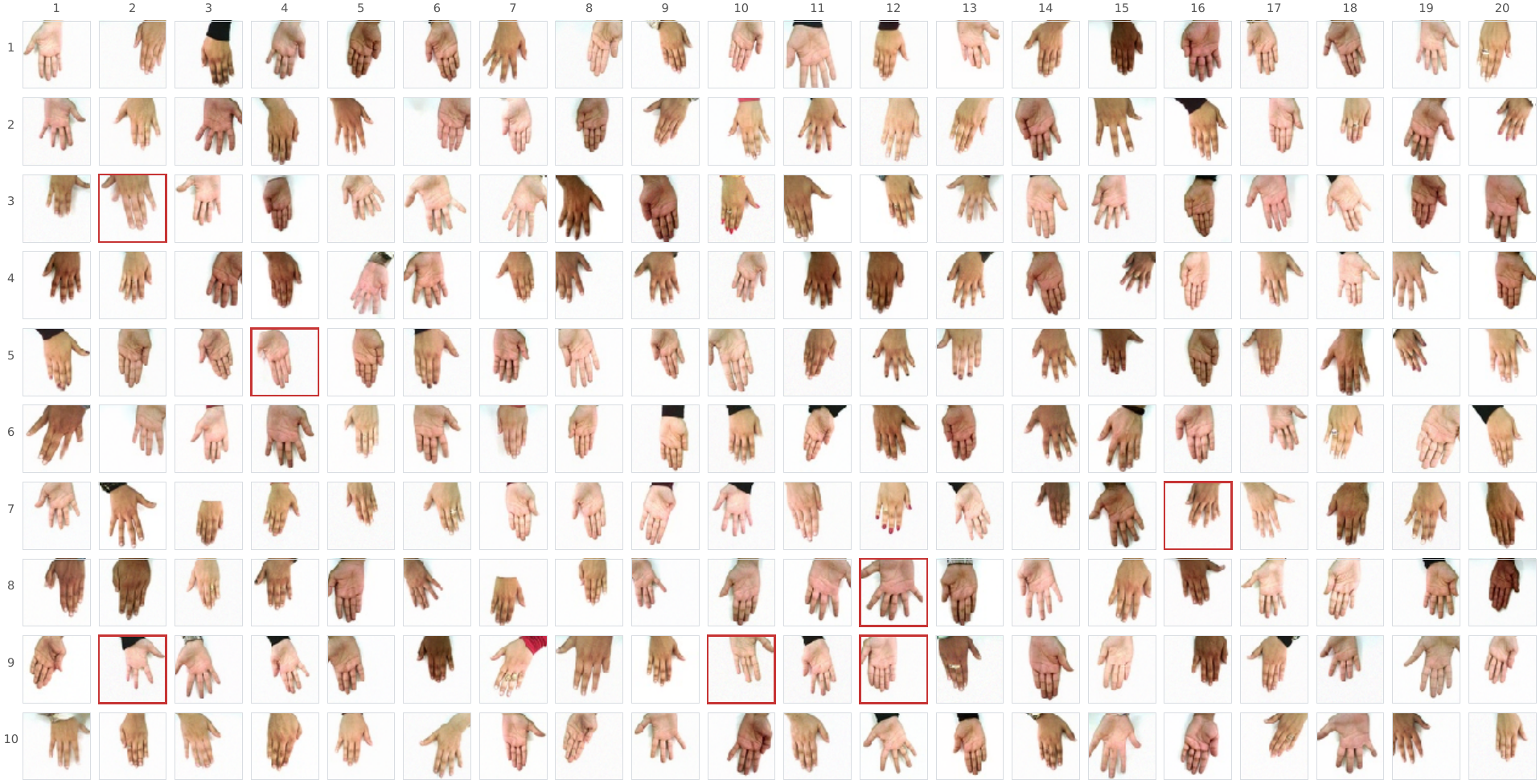}
        \caption{Baseline ($\eta=0$)}
    \end{subfigure}
    \par\medskip
    \begin{subfigure}{0.98\linewidth}
        \centering
        \includegraphics[width=\linewidth]{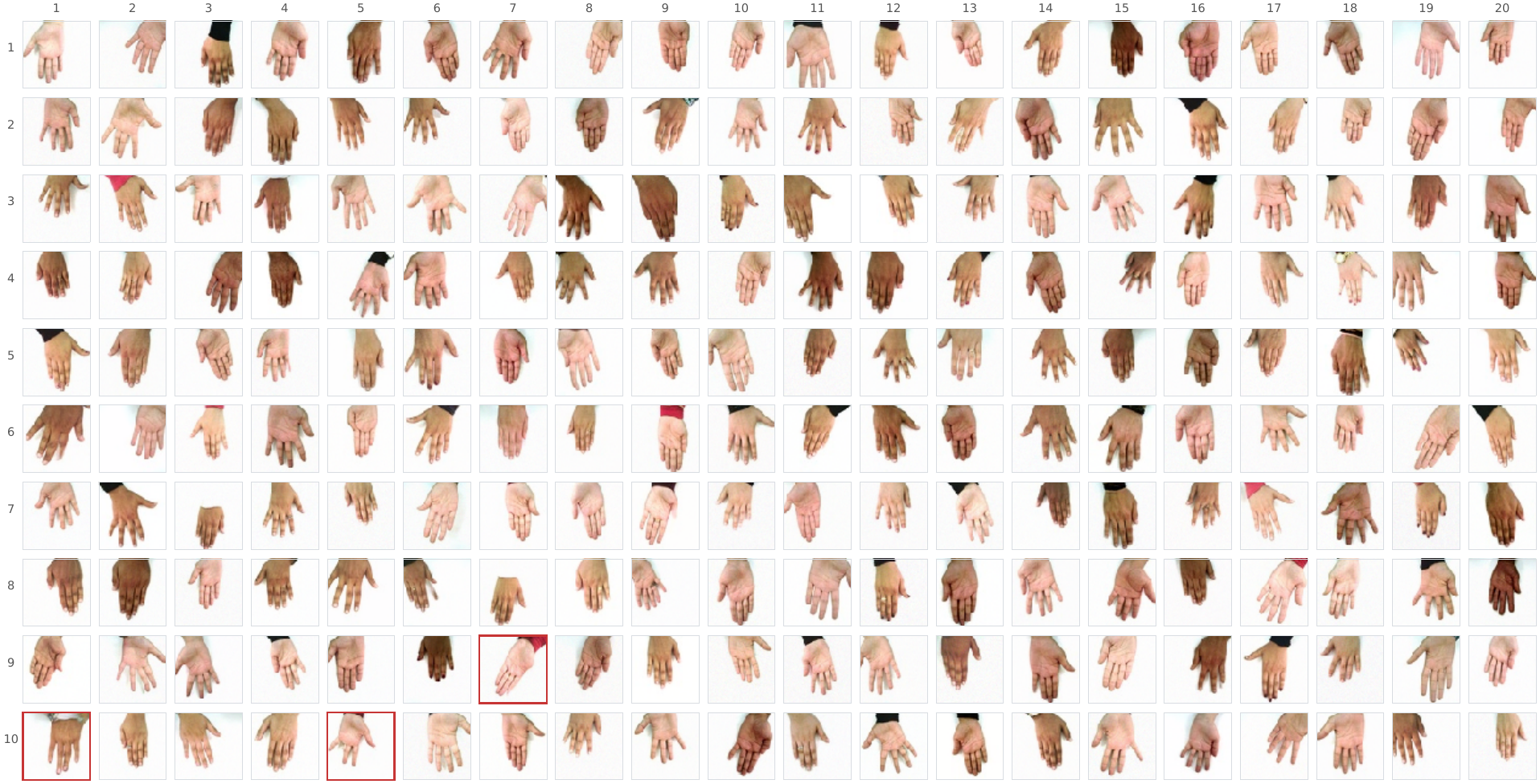}
        \caption{SSD ($\eta=0.8$)}
    \end{subfigure}
    \caption{Comparison of generated samples on the 11kHands dataset. Each panel presents 200 samples generated using the same random seed. Thin red outlines highlight samples exhibiting spatial inconsistencies, specifically an incorrect number of fingers. (a) The baseline model frequently generates hands with six fingers, (b) whereas SSD mitigates this issue but occasionally produces hands with only four fingers. The full evaluation results are reported in~\cref{tab:hands_consistency}.}
    \label{fig:hands_samples}
\end{figure}
\clearpage

\section{Generation visualization}
\begin{figure}[ht]
    \centering
    \begin{subfigure}[b]{0.85\linewidth}
        \centering
        \includegraphics[width=0.9\linewidth]{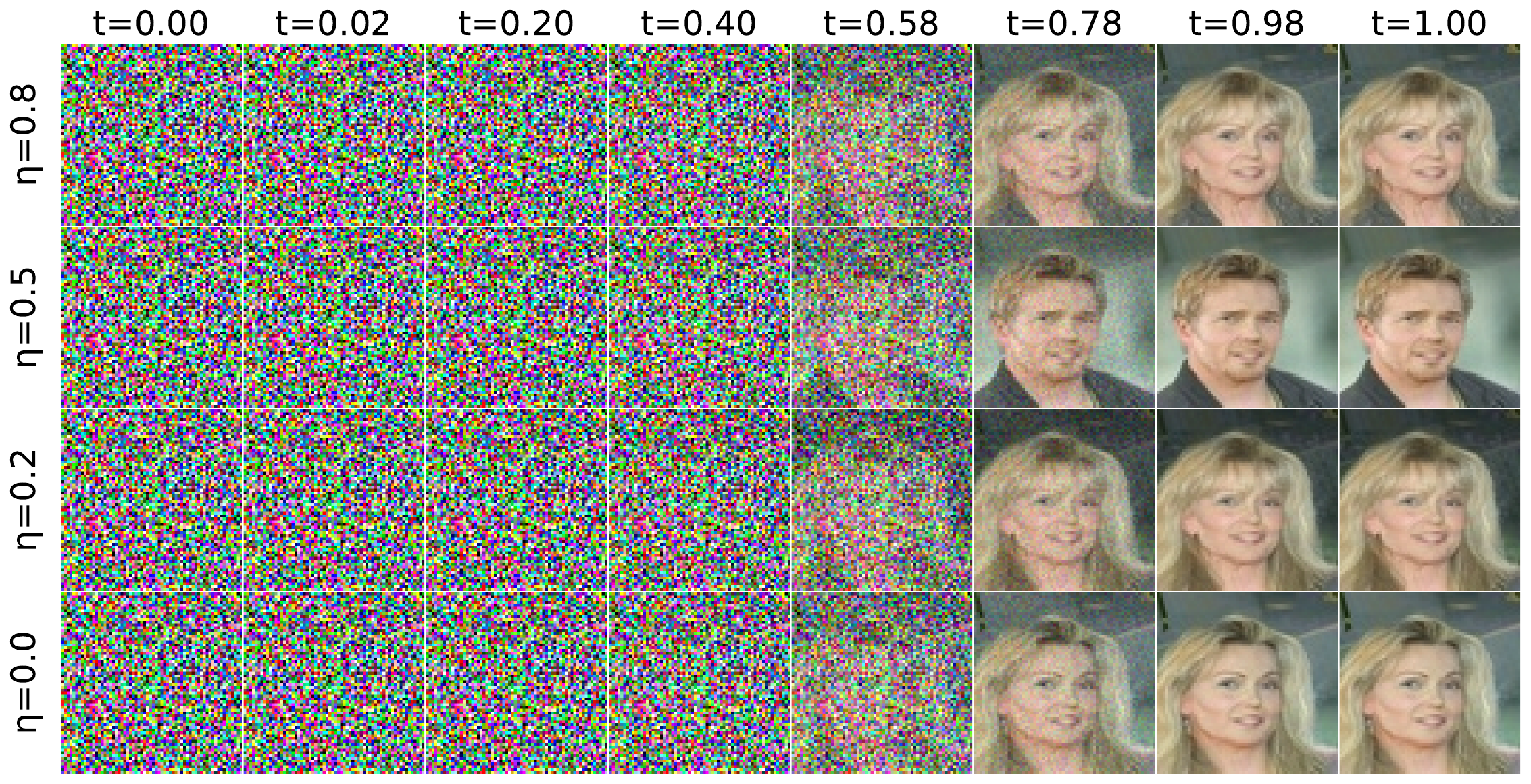}
        \caption{CelebA-50K 64\texttimes64}
        \label{fig:sample_paths_celeba}
    \end{subfigure}
    
    \begin{subfigure}[b]{0.85\linewidth}
        \centering
        \includegraphics[width=0.9\linewidth]{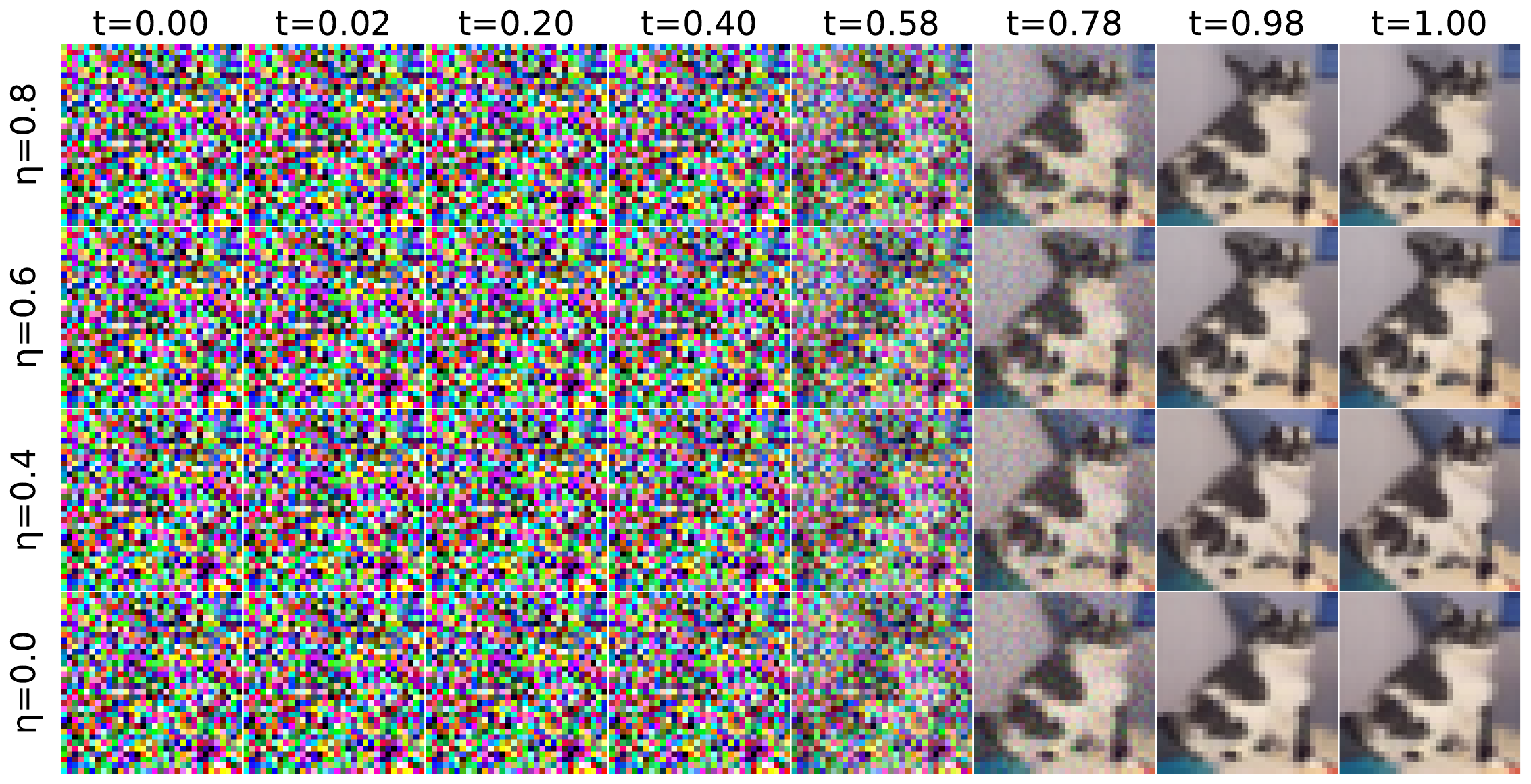}
        \caption{CIFAR10 32\texttimes32}
        \label{fig:sample_paths_cifar10}
    \end{subfigure}
    
    \begin{subfigure}[b]{0.85\linewidth}
        \centering
        \includegraphics[width=0.9\linewidth]{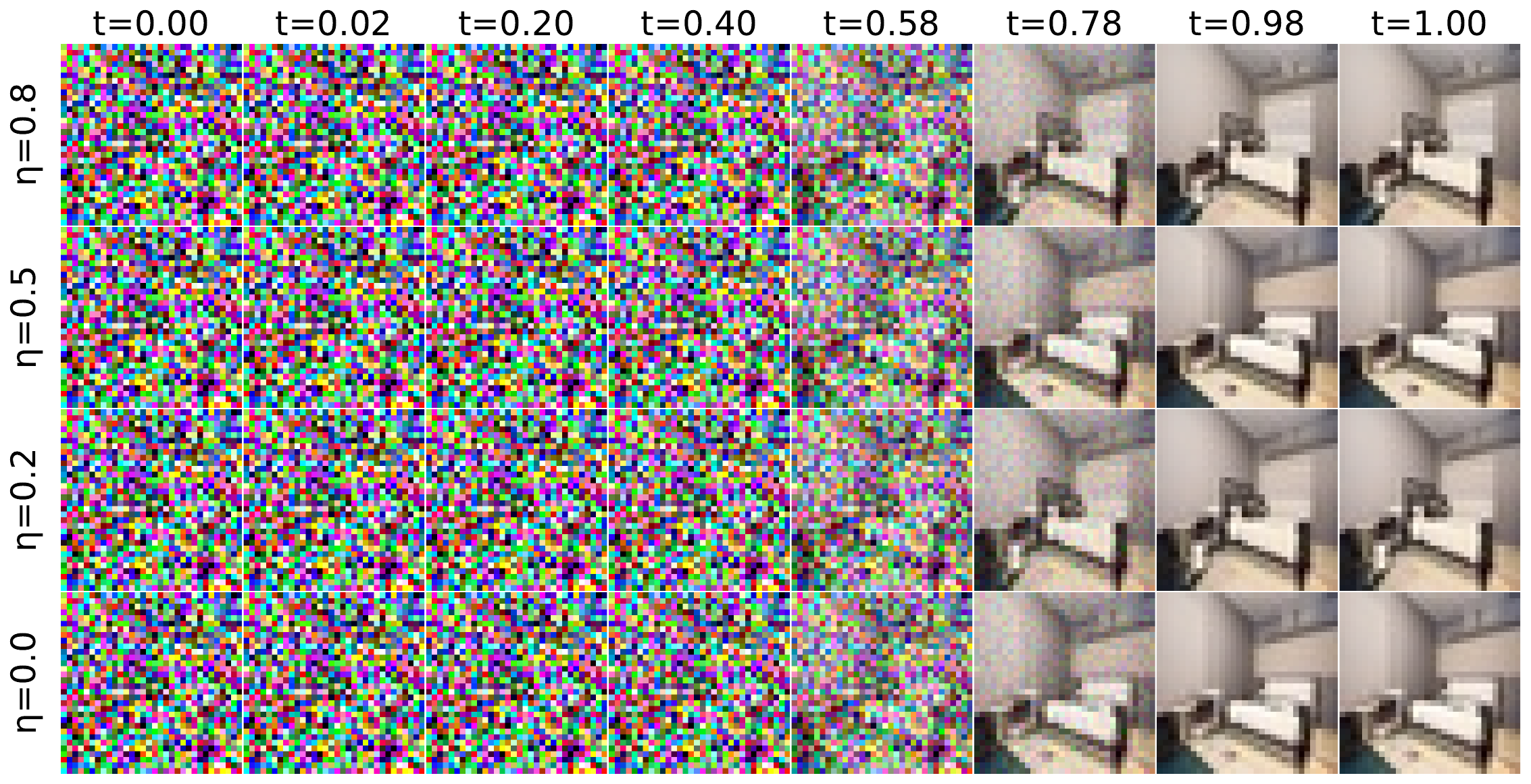}
        \caption{LSUN Bedroom 32\texttimes32}
        \label{fig:sample_paths_lsun}
    \end{subfigure}
    
    \caption{Sample paths from same initial noises with models trained with different $\eta$ across three datasets.}
    \label{fig:sample_paths_comparison}
\end{figure}

\begin{figure}[ht]
    \centering
    \includegraphics[width=0.8\linewidth]{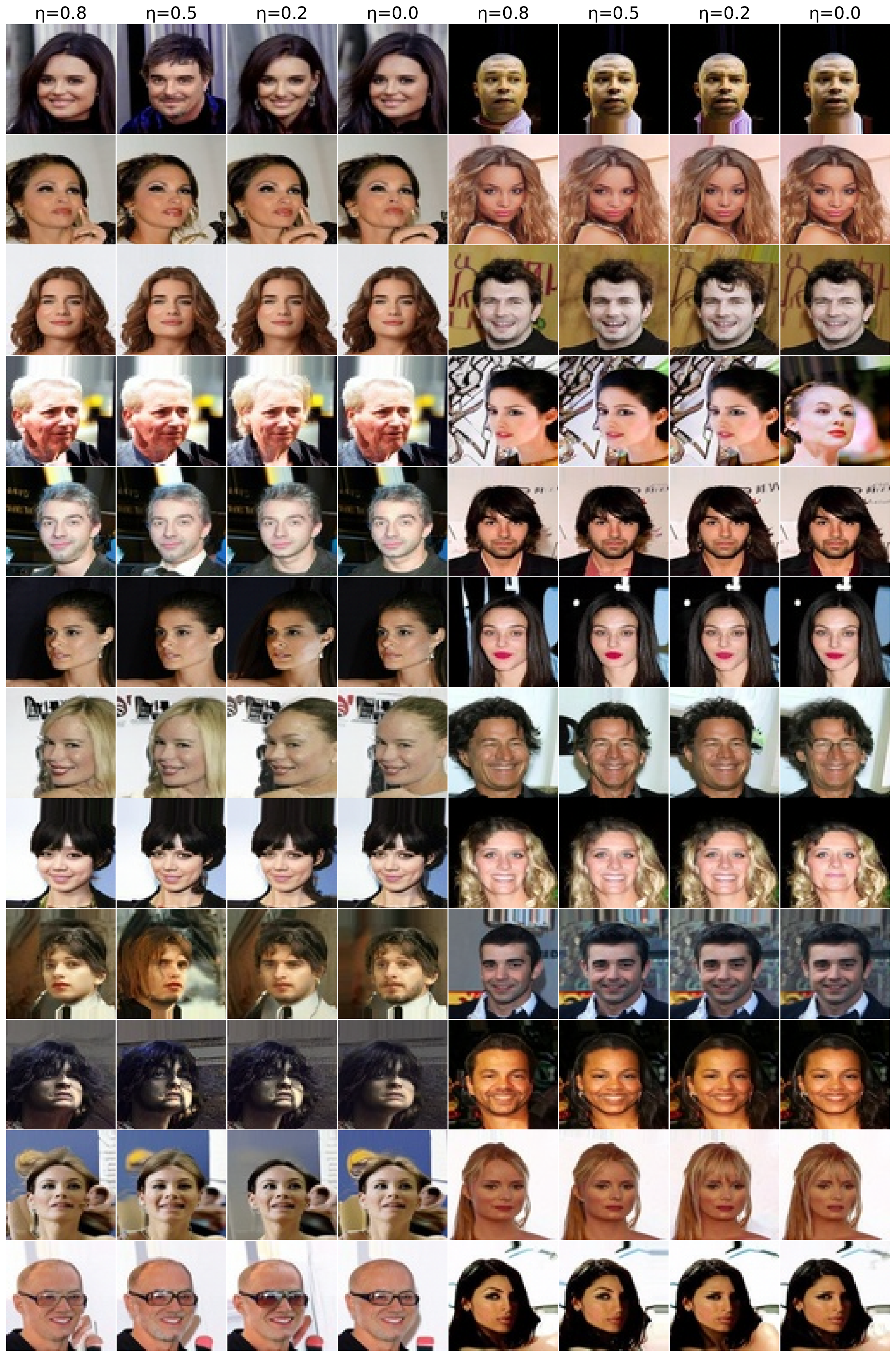}
    \caption{More samples generated by models trained on CelebA-50K with different mask ratios. }
    \label{fig:celeba_more_compare}
\end{figure}

\begin{figure}[ht]
    \centering
    \includegraphics[width=0.8\linewidth]{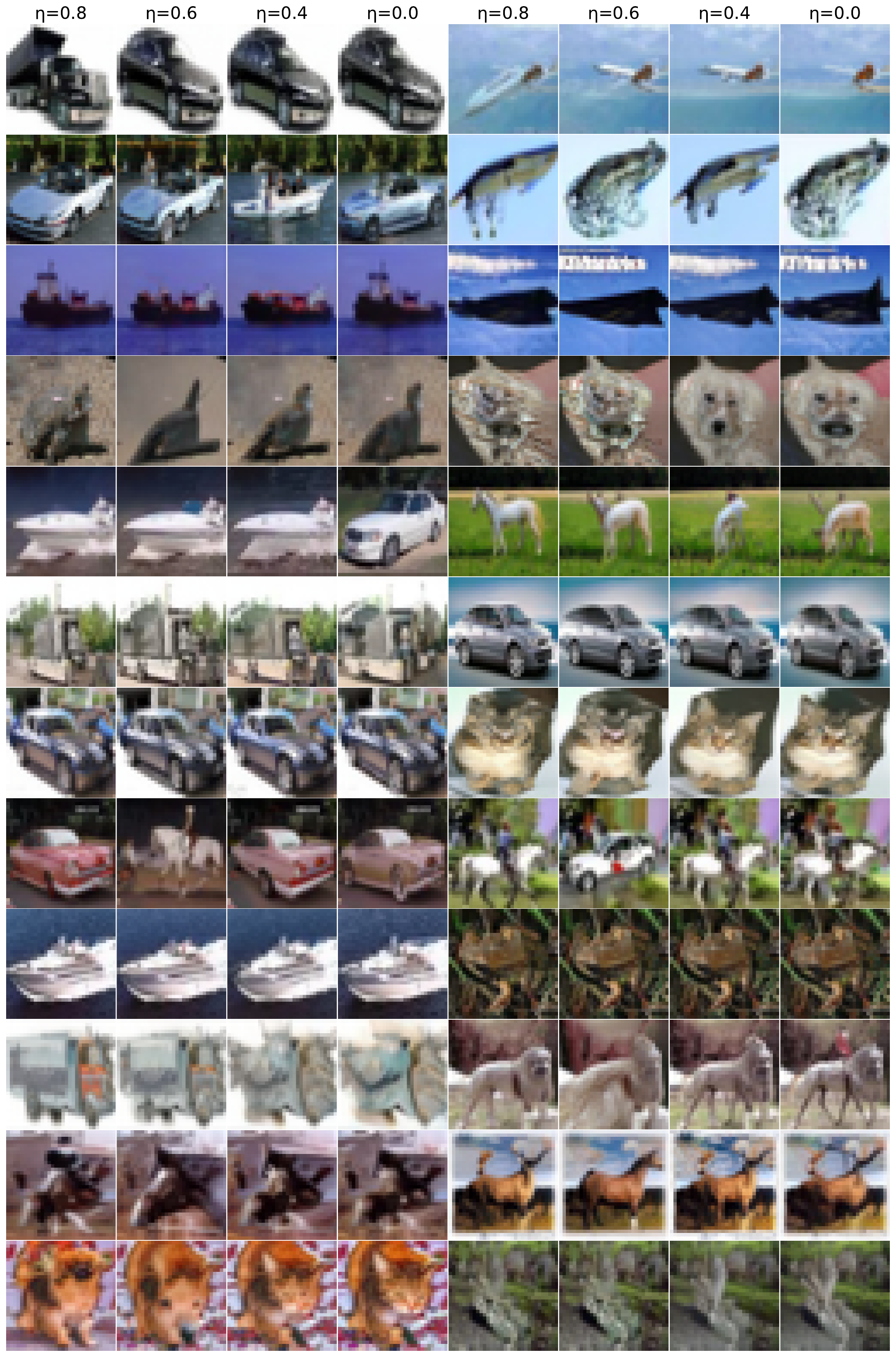}
    \caption{More samples generated by models trained on CIFAR10 with different mask ratios. }
    \label{fig:cifar10_more_compare}
\end{figure}

\begin{figure}[ht]
    \centering
    \includegraphics[width=0.8\linewidth]{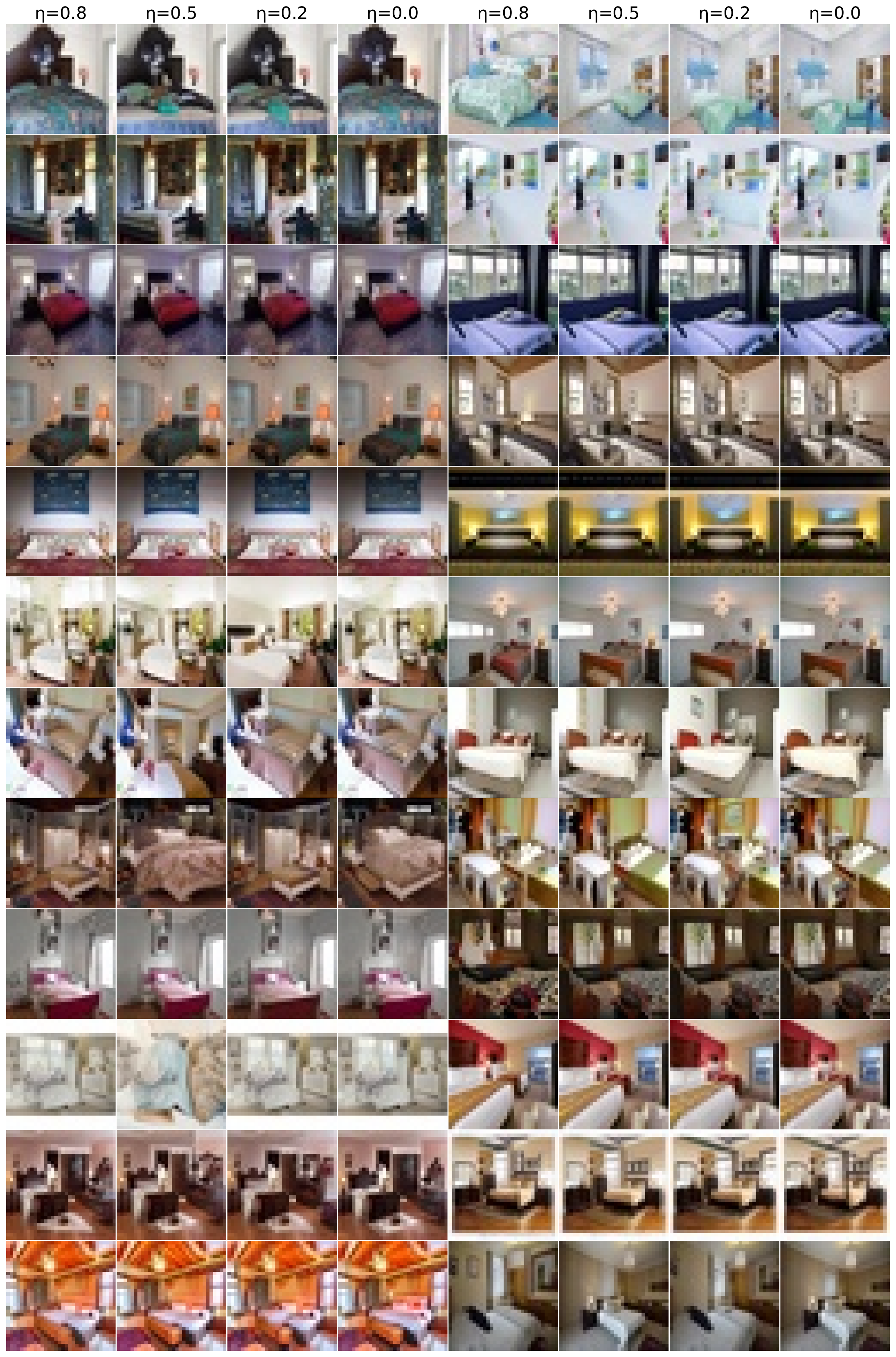}
    \caption{More samples generated by models trained on LSUN Bedroom with different mask ratios. }
    \label{fig:lsun_more_compare}
\end{figure}

We visualize the sample paths from the same initial noise using models trained with different mask ratios. The sampling is based on the method proposed in~\citet{karras2022elucidating} and we use Heun's second-order method as the ODE solver~\citep{ascher1998computer}. We set NFE as 50. The sample paths in~\cref{fig:sample_paths_comparison} show that the models may diverge from around $t=0.58$. More generated samples are shown in~\ref{fig:celeba_more_compare},~\ref{fig:cifar10_more_compare} and~\ref{fig:lsun_more_compare}.

\clearpage
\section*{NeurIPS Paper Checklist}

\begin{enumerate}

\item {\bf Claims}
    \item[] Question: Do the main claims made in the abstract and introduction accurately reflect the paper's contributions and scope?
    \item[] Answer: \answerYes{} %
    \item[] Justification: The abstract and Sec. 1 (Introduction) clearly state our three main contributions: (i) the analytical result that SSD modifies the spectrum of the data covariance (developed in Sec. 4.2 and Appendix A); (ii) targeted experiments showing improved spatial consistency, more accurate population score estimation, and better contextual usage (Secs. 5.4–5.6); and (iii) competitive FID scores at high masking ratios together with mitigation of memorization and training instability on small datasets (Secs. 5.1–5.3, Table 1).

\item {\bf Limitations}
    \item[] Question: Does the paper discuss the limitations of the work performed by the authors?
    \item[] Answer: \answerYes{} %
    \item[] Justification: The limitation regarding the choice of masking ratio is acknowledged in Sec. 5 ("In all experiments, we do not tune the masking ratio... Determining an optimal masking ratio is left for future work") and reiterated in Sec. 6 (Discussion and conclusion).

\item {\bf Theory assumptions and proofs}
    \item[] Question: For each theoretical result, does the paper provide the full set of assumptions and a complete (and correct) proof?
    \item[] Answer: \answerYes{} %
    \item[] Justification: The theoretical results on how masking changes the covariance spectrum are stated in Sec. 4.2 (Eqs. 9–13) and Sec. 4.3 (locality analysis). The full derivation, including all assumptions (zero-mean Gaussian data, independent Bernoulli masking) and the complete proof, is provided in Appendix A (Secs. A.1–A.3, Eqs. 15–20).

    \item {\bf Experimental result reproducibility}
    \item[] Question: Does the paper fully disclose all the information needed to reproduce the main experimental results of the paper to the extent that it affects the main claims and/or conclusions of the paper (regardless of whether the code and data are provided or not)?
    \item[] Answer: \answerYes{} %
    \item[] Justification: The training objective and method are fully specified in Sec. 4.1 (Eq. 8). Datasets, model architectures (U-Net, DiT), ODE solver (Heun's second-order), NFE, and FID evaluation protocol are described in Sec. 5 (Data \& experimental setup). Complete hyperparameters for CelebA-50K, CIFAR-10, and LSUN Bedroom are reported in Appendix C, Table 6, and ImageNet settings follow DiT defaults as noted in Appendix~C. The 2D Gaussian score-estimation experiment is fully specified in Appendix~B (Eqs.~21--29).

\item {\bf Open access to data and code}
    \item[] Question: Does the paper provide open access to the data and code, with sufficient instructions to faithfully reproduce the main experimental results, as described in supplemental material?
    \item[] Answer: \answerYes{} %
    \item[] Justification: Our method is model-agnostic and can be implemented with only a few lines of code. We have released the code in the project website. We are only using public datasets. Please refer to their downloading instructions.

\item {\bf Experimental setting/details}
    \item[] Question: Does the paper specify all the training and test details (e.g., data splits, hyperparameters, how they were chosen, type of optimizer) necessary to understand the results?
    \item[] Answer: \answerYes{} %
    \item[] Justification: Training and evaluation details (epochs, batch sizes, optimizer, learning rate, EMA, resolution, FM configuration, sampling configuration, and model architecture) are reported in Sec. 5 and comprehensively in Appendix C, Table 6. The 2D Gaussian experiment setup is detailed in Appendix B.

\item {\bf Experiment statistical significance}
    \item[] Question: Does the paper report error bars suitably and correctly defined or other appropriate information about the statistical significance of the experiments?
    \item[] Answer: \answerYes{} %
    \item[] Justification: FID curves in Fig. 3 (Sec. 5.1) are averaged over four runs, with shaded regions denoting 95\% confidence intervals. Standard deviations across four runs are also reported in Table 1 (Sec. 5.1) for CIFAR-10, LSUN, and CelebA-50K.

\item {\bf Experiments compute resources}
    \item[] Question: For each experiment, does the paper provide sufficient information on the computer resources (type of compute workers, memory, time of execution) needed to reproduce the experiments?
    \item[] Answer: \answerYes{} %
    \item[] Justification: Compute resources are reported in Appendix C, Table 6 (8 GPUs for CelebA-50K, CIFAR-10, and LSUN Bedroom). For ImageNet, the global batch size (4096) and number of training steps (180k) are specified in Appendix C.
    
\item {\bf Code of ethics}
    \item[] Question: Does the research conducted in the paper conform, in every respect, with the NeurIPS Code of Ethics \url{https://neurips.cc/public/EthicsGuidelines}?
    \item[] Answer: \answerYes{} %
    \item[] Justification: The research uses standard public benchmark datasets (CIFAR-10, CelebA, LSUN Bedroom, ImageNet) cited in Sec. 5 and conforms to the NeurIPS Code of Ethics. No human subjects, crowdsourcing, or sensitive data are involved.

\item {\bf Broader impacts}
    \item[] Question: Does the paper discuss both potential positive societal impacts and negative societal impacts of the work performed?
    \item[] Answer: \answerNA{} %
    \item[] Justification: This paper presents work whose goal is to advance the field of Machine Learning. There are many potential societal consequences of our work, none of which we feel must be specifically highlighted here.
    
\item {\bf Safeguards}
    \item[] Question: Does the paper describe safeguards that have been put in place for responsible release of data or models that have a high risk for misuse (e.g., pre-trained language models, image generators, or scraped datasets)?
    \item[] Answer: \answerNA{} %
    \item[] Justification: \answerNA{}

\item {\bf Licenses for existing assets}
    \item[] Question: Are the creators or original owners of assets (e.g., code, data, models), used in the paper, properly credited and are the license and terms of use explicitly mentioned and properly respected?
    \item[] Answer: \answerYes{} %
    \item[] Justification: All existing assets used — datasets (CIFAR-10 [46], CelebA [47], LSUN [48], ImageNet [49]) and code/methods (DiT [7], Flow Matching [50], EDM sampler [53], U-Net [51], Heun's method [52], FID [54]) — are properly cited in Sec. 5 and the References. Their use is consistent with their respective licenses and intended research use.

\item {\bf New assets}
    \item[] Question: Are new assets introduced in the paper well documented and is the documentation provided alongside the assets?
    \item[] Answer: \answerNA{} %
    \item[] Justification: \answerNA{}

\item {\bf Crowdsourcing and research with human subjects}
    \item[] Question: For crowdsourcing experiments and research with human subjects, does the paper include the full text of instructions given to participants and screenshots, if applicable, as well as details about compensation (if any)? 
    \item[] Answer: \answerNA{} %
    \item[] Justification: \answerNA{}

\item {\bf Institutional review board (IRB) approvals or equivalent for research with human subjects}
    \item[] Question: Does the paper describe potential risks incurred by study participants, whether such risks were disclosed to the subjects, and whether Institutional Review Board (IRB) approvals (or an equivalent approval/review based on the requirements of your country or institution) were obtained?
    \item[] Answer: \answerNA{} %
    \item[] Justification: \answerNA{}

\item {\bf Declaration of LLM usage}
    \item[] Question: Does the paper describe the usage of LLMs if it is an important, original, or non-standard component of the core methods in this research? Note that if the LLM is used only for writing, editing, or formatting purposes and does \emph{not} impact the core methodology, scientific rigor, or originality of the research, declaration is not required.
    \item[] Answer: \answerNA{} %
    \item[] Justification: \answerNA{}

\end{enumerate} 
\end{document}